\documentclass[11pt]{article}

\PassOptionsToPackage{table}{xcolor}
\usepackage[preprint]{acl}

\usepackage{times}
\usepackage{latexsym}
\usepackage[T1]{fontenc}
\usepackage{iftex}
\ifluatex\else\ifxetex\else\usepackage[utf8]{inputenc}\fi\fi
\ifluatex
\DeclareFontFamily{TU}{georgia}{}
\DeclareFontShape{TU}{georgia}{m}{n}{<->"[fonts/georgia.ttf]:mode=node;+kern;+liga;+tlig"}{}
\DeclareFontShape{TU}{georgia}{b}{n}{<->"[fonts/georgiab.ttf]:mode=node;+kern;+liga;+tlig"}{}
\DeclareFontShape{TU}{georgia}{m}{it}{<->"[fonts/georgiai.ttf]:mode=node;+kern;+liga;+tlig"}{}
\DeclareFontShape{TU}{georgia}{b}{it}{<->"[fonts/georgiaz.ttf]:mode=node;+kern;+liga;+tlig"}{}
\DeclareFontShape{TU}{georgia}{bx}{n}{<->ssub*georgia/b/n}{}
\DeclareFontShape{TU}{georgia}{bx}{it}{<->ssub*georgia/b/it}{}
\newcommand{\figfont}{\fontencoding{TU}\fontfamily{georgia}\fontsize{9}{11}\selectfont}
\else\ifxetex   %
\DeclareFontFamily{TU}{georgia}{}
\DeclareFontShape{TU}{georgia}{m}{n}{<->"[fonts/georgia.ttf]:+kern;+liga;mapping=tex-text"}{}
\DeclareFontShape{TU}{georgia}{b}{n}{<->"[fonts/georgiab.ttf]:+kern;+liga;mapping=tex-text"}{}
\DeclareFontShape{TU}{georgia}{m}{it}{<->"[fonts/georgiai.ttf]:+kern;+liga;mapping=tex-text"}{}
\DeclareFontShape{TU}{georgia}{b}{it}{<->"[fonts/georgiaz.ttf]:+kern;+liga;mapping=tex-text"}{}
\DeclareFontShape{TU}{georgia}{bx}{n}{<->ssub*georgia/b/n}{}
\DeclareFontShape{TU}{georgia}{bx}{it}{<->ssub*georgia/b/it}{}
\newcommand{\figfont}{\fontencoding{TU}\fontfamily{georgia}\fontsize{9}{11}\selectfont}
\else
\newcommand{\figfont}{\fontsize{9}{11}\selectfont}
\fi\fi
\usepackage{microtype}
\usepackage{inconsolata}
\usepackage{graphicx}
\usepackage{booktabs}
\usepackage{array}
\usepackage{amsmath}
\usepackage{xcolor}
\usepackage{tikz}
\usepackage{enumitem}
\usepackage{amsthm}
\usepackage{algorithm}
\usepackage{algorithmic}
\usepackage{wrapfig}
\makeatletter\newcommand{\captionof}[1]{\def\@captype{#1}\caption}\makeatother
\newsavebox{\rowboxA}\newsavebox{\rowboxB}\newlength{\rowh}\newlength{\radunit}
\theoremstyle{plain}\newtheorem{proposition}{Proposition}
\setlist[itemize]{noitemsep, topsep=3pt, leftmargin=*}
\setlist[enumerate]{noitemsep, topsep=2pt, leftmargin=*}
\usetikzlibrary{arrows.meta,positioning,fit,backgrounds,patterns}

\newsavebox{\fitboxbox}
\let\origresizebox\resizebox
\newcommand{\fitbox}[1]{\sbox{\fitboxbox}{#1}\ifdim\wd\fitboxbox>\linewidth\typeout{FITBOX shrink \the\wd\fitboxbox\space -> \the\linewidth\space (page \thepage, line \the\inputlineno)}\origresizebox{\linewidth}{!}{\usebox{\fitboxbox}}\else\usebox{\fitboxbox}\fi}
\newcommand{\floattext}{\footnotesize\let\small\footnotesize\let\scriptsize\footnotesize\def\resizebox##1##2##3{\fitbox{##3}}}
\makeatletter   %
\def\floattext@alg{algorithm}
\g@addto@macro\@floatboxreset{\ifx\@captype\floattext@alg\else\floattext\fi}
\makeatother
\AtBeginEnvironment{wrapfigure}{\floattext}
\definecolor{boxbar}{HTML}{2F3E4C}\definecolor{boxbg}{HTML}{F4F6F8}\definecolor{boxrule}{HTML}{C5CCD3}
\pgfdeclarelayer{boxbg}\pgfsetlayers{boxbg,main}
\newcommand{\ph}[1]{\texttt{\textcolor{cOurs}{\{#1\}}}}
\newcommand{\bhskip}{}  %
\newcommand{\bh}[1]{\par\bhskip\gdef\bhskip{\smallskip}\noindent{\bfseries #1}\par\nobreak\noindent\ignorespaces}
\newcommand{\codebox}[1]{\par\smallskip\noindent\fcolorbox{boxrule}{white}{\parbox{\dimexpr\linewidth-2\fboxsep-2\fboxrule\relax}{\raggedright\ttfamily #1}}\par}
\newcommand{\boxicon}{\tikz[baseline=-0.6ex]{\draw[white, line width=0.45pt, rounded corners=0.7pt] (0,-2.6pt) rectangle (8pt,3.4pt);
  \draw[white, line width=0.5pt] (1.6pt,1.7pt) -- (3.2pt,0.4pt) -- (1.6pt,-0.9pt); \draw[white, line width=0.5pt] (4pt,-1pt) -- (6.4pt,-1pt);}}
\newcounter{pbox}\renewcommand{\thepbox}{\alph{pbox}}
\newcommand{\boxref}[1]{Box~(\ref{#1})}
\newcommand{\promptbox}[5]{\par\medskip\noindent\begin{minipage}{\linewidth}\begin{tikzpicture}[font=\footnotesize]
  \node[anchor=north west, inner sep=5pt, text width=\dimexpr\linewidth-10.8pt\relax, align=left] (pb) at (0,0)
    {\raggedright\setlength{\parskip}{0pt}\gdef\bhskip{}#3};
  \begin{pgfonlayer}{boxbg}
    \draw[boxrule, fill=boxbg, line width=0.4pt, rounded corners=3pt] (pb.south west) rectangle ([yshift=12pt]pb.north east);
    \fill[boxbar] (pb.north west) [rounded corners=3pt] -- ([yshift=12pt]pb.north west) -- ([yshift=12pt]pb.north east)
      [sharp corners] -- (pb.north east) -- cycle;
  \end{pgfonlayer}
  \node[anchor=west, text=white, font=\footnotesize\bfseries, inner sep=0pt] at ([xshift=5pt, yshift=6pt]pb.north west) {\boxicon\enspace #1};
  \node[anchor=east, fill=white, text=boxbar, font=\footnotesize, rounded corners=1.5pt, inner xsep=2.5pt, inner ysep=1pt]
    at ([xshift=-4pt, yshift=6pt]pb.north east) {#2};
\end{tikzpicture}\par\nobreak\smallskip
{\centering\refstepcounter{pbox}\label{#5}{\fontsize{10}{12}\selectfont(\thepbox)~#4}\par}\end{minipage}\par\medskip}

\definecolor{cOurs}{HTML}{0050D0}   %
\definecolor{cOursL}{HTML}{4DA3FF}  %
\definecolor{cProc}{HTML}{F28C28}   %
\definecolor{cProcL}{HTML}{FFC58A}  %
\definecolor{cVote}{HTML}{6FA8FF}   %
\definecolor{cAnn}{HTML}{8FD3F5}    %
\definecolor{cPlain}{HTML}{FFB8D2}  %
\definecolor{cSCnine}{HTML}{FF7F7F} %
\definecolor{cSCsix}{HTML}{FFBDB5}  %
\definecolor{cPercep}{HTML}{7FD48F} %
\definecolor{cSimToM}{HTML}{66D0C4} %
\definecolor{cCoT}{HTML}{F59AC3}
\definecolor{cZS}{HTML}{C7B3F0}
\definecolor{cRetr}{HTML}{F5D45E}
\definecolor{cGEPA}{HTML}{A99BF7}
\definecolor{cNative}{HTML}{9DBAD6}
\definecolor{cPos}{HTML}{00A86B}
\definecolor{cNeg}{HTML}{E8590C}
\definecolor{cGray}{HTML}{B8C4CE}
\definecolor{cVC}{HTML}{00A08A}    %
\definecolor{cVCs}{HTML}{4CC4B4}
\definecolor{cAO}{HTML}{9ADBD2}
\definecolor{cSearch}{HTML}{F2B33D}
\colorlet{cRSI}{cProc}\colorlet{cRSIL}{cProcL}\colorlet{cSC}{cSCnine}\colorlet{cSCL}{cSCsix}
\colorlet{pos}{cPos}\colorlet{neg}{cNeg}
\definecolor{gray}{HTML}{7F8C8D}
\definecolor{ours}{HTML}{E3EDFF}    %
\definecolor{card}{HTML}{F4F6F7}    %
\newcommand{\up}[1]{\textsubscript{\textcolor{pos}{+#1}}}
\newcommand{\dn}[1]{\textsubscript{\textcolor{neg}{$-$#1}}}
\newcommand{\eq}{\textsubscript{\textcolor{gray}{0.0}}}

\newcommand{\AFMainPArb}{78.1}

\newcommand{\AFMainVsix}{78.9}

\newcommand{\ALibRounds}{4}
\newcommand{\ARoundsTotal}{24}

\newcommand{\AblEmotionAnn}{73.1}
\newcommand{\AblEmotionPlain}{77.2}

\newcommand{\AblNonlitAnn}{72.9}
\newcommand{\AblNonlitPlain}{80.5}
\newcommand{\AdmitA}{0--1}
\newcommand{\AdmitC}{2--5}
\newcommand{\ArbDevAnn}{254}
\newcommand{\ArbDevConf}{561}
\newcommand{\ArbDevEvid}{314}
\newcommand{\ArbDevNeutral}{218}
\newcommand{\ArbDevPlain}{267}
\newcommand{\ArbDevSimpleConf}{140}
\newcommand{\ArbDevSimpleEvid}{114}
\newcommand{\ArbDevSimpleNeutral}{23}

\newcommand{\CallsSC}{6.0}

\newcommand{\CallsVsix}{7.3}

\newcommand{\ConBeliefVsix}{+3.9}
\newcommand{\ConBeliefVsixCI}{[+2.5, +5.4]}

\newcommand{\ConKnowledgeVsix}{+1.1}
\newcommand{\ConKnowledgeVsixCI}{[+0.2, +2.0]}

\newcommand{\ConWBelief}{+1.2}
\newcommand{\ConWBeliefCI}{[+0.3, +2.1]}

\newcommand{\ConWMixed}{$-$4.1}
\newcommand{\ConWMixedCI}{[$-$8.4, $-$0.6]}

\newcommand{\ConWNonlit}{$-$1.7}
\newcommand{\ConWNonlitCI}{[$-$3.5, $-$0.1]}

\newcommand{\EOneCI}{[$-$7.3, +8.6]}
\newcommand{\EOneDiff}{+0.6}
\newcommand{\ETwoCI}{[$-$6.8, +3.2]}
\newcommand{\ETwoDiff}{$-$1.8}

\newcommand{\EnsSCPooled}{76.4}

\newcommand{\EnsVsixSingle}{78.7}

\newcommand{\FanRsip}{25.3}

\newcommand{\FanVsix}{28.0}

\newcommand{\FlipHelped}{76}
\newcommand{\FlipHurt}{39}

\newcommand{\InheritCovMax}{10}

\newcommand{\MacroEmptySCnine}{+1.5}
\newcommand{\MacroEmptySCnineCI}{[+0.6, +2.4]}
\newcommand{\MacroHandEmpty}{+3.0}
\newcommand{\MacroHandEmptyCI}{[+2.2, +3.7]}

\newcommand{\MacroLoopEmpty}{+1.7}
\newcommand{\MacroLoopEmptyCI}{[+1.1, +2.3]}

\newcommand{\OvRSIA}{59.3}
\newcommand{\OvRsip}{71.8}
\newcommand{\OvSC}{68.3}
\newcommand{\OvSCnine}{68.7}

\newcommand{\OvVsix}{73.1}

\newcommand{\PairCols}{8}

\newcommand{\PairFanVsix}{+4.3}
\newcommand{\PairFanVsixCI}{[+0.2, +8.6]}

\newcommand{\PairHiToMVsix}{+4.4}
\newcommand{\PairHiToMVsixCI}{[+1.2, +7.6]}

\newcommand{\PairHolmLossesVsix}{0}

\newcommand{\PairHolmWinsVsix}{0}

\newcommand{\PairLossesVsix}{1}

\newcommand{\PairSimpleVsixSC}{+19.9}

\newcommand{\PairTiesVsix}{5}

\newcommand{\PairWinsVsix}{2}
\newcommand{\PercFanP}{98.7}
\newcommand{\PercFanR}{73.8}
\newcommand{\PercToMiP}{98.2}
\newcommand{\PercToMiR}{38.3}

\newcommand{\RExploreNative}{67.2}

\newcommand{\RExploreRSIA}{73.9}

\newcommand{\RExploreRsip}{74.0}

\newcommand{\RExploreVsix}{72.5}

\newcommand{\RFanRSIA}{2.3}

\newcommand{\RHiToMRsip}{70.0}

\newcommand{\RHiToMVsix}{69.3}

\newcommand{\ROpenToMRsip}{76.5}

\newcommand{\ROpenToMVsix}{75.8}

\newcommand{\RSIRunHours}{4.3}

\newcommand{\RSimpleRsip}{81.7}

\newcommand{\RSimpleVsix}{90.0}

\newcommand{\RouteBeliefAdj}{353}

\newcommand{\RouteBeliefConf}{602}

\newcommand{\RouteNonlitAdj}{16}

\newcommand{\RouteNonlitConf}{27}
\newcommand{\RouteNonlitPlain}{25}

\newcommand{\RsipFinalN}{300}
\newcommand{\RsipNumSeeds}{6}

\newcommand{\SensMixBest}{79.9}

\newcommand{\SensMixOurs}{79.9}

\newcommand{\TokRatioVsix}{1.51}

\newcommand{\ToolOmitP}{99.7}
\newcommand{\ToolOmitR}{42.2}
\newcommand{\ToolOverP}{74.8}
\newcommand{\ToolOverR}{88.5}

\newcommand{\HOvSCnine}{68.5}

\newcommand{\HOvVsix}{73.5}
\newcommand{\HOvVthree}{72.0}

\newcommand{\AblRSIAMain}{59.3}

\newcommand{\TBLlamaSC}{69.6}

\newcommand{\TBLlamaVsix}{72.3}

\newcommand{\TBQwenLSC}{72.1}

\newcommand{\TBQwenLVsix}{75.6}

\newcommand{\SuiteHthreeQ}{2{,}470}

\newcommand{\SuiteMainQ}{3{,}349}

\newcommand{\BBMarginMax}{+8.6}
\newcommand{\BBMarginMin}{+5.1}
\newcommand{\BBRatioMax}{1.15}
\newcommand{\BBRatioMin}{0.98}
\newcommand{\CurveFinalEmpty}{59.3}
\newcommand{\CurveFinalHand}{59.7}
\newcommand{\CurveFinalLoop}{63.2}
\newcommand{\CurveLaterHi}{62.9}
\newcommand{\CurveLaterLo}{60.2}
\newcommand{\CurveOne}{62.3}
\newcommand{\CurveZero}{56.0}
\newcommand{\EmptyMaxMain}{70.3}

\newcommand{\FanPercepToMSCsix}{3.7}
\newcommand{\FanSimToMSC}{1.8}

\newcommand{\FsixtyTwoAllVsixGold}{25.2}
\newcommand{\FsixtyTwoAllVsixModel}{26.0}

\newcommand{\FsixtyTwoBelVsixGold}{91.2}
\newcommand{\FsixtyTwoBelVsixModel}{79.3}

\newcommand{\HHiSCnine}{69.1}
\newcommand{\HHiVsix}{74.2}

\newcommand{\HLoSCnine}{68.1}
\newcommand{\HLoVsix}{73.1}

\newcommand{\HNewAllSCnine}{67.7}
\newcommand{\HNewAllSimToM}{71.2}
\newcommand{\HNewAllVsix}{67.1}
\newcommand{\HOvAllRsip}{72.7}

\newcommand{\NewMeanPercepToMSCsix}{66.7}
\newcommand{\NewMeanRsip}{64.0}
\newcommand{\NewMeanSC}{65.3}
\newcommand{\NewMeanSCnine}{66.2}

\newcommand{\NewMeanVsix}{65.2}
\newcommand{\NewToMiSimToMSC}{85.6}
\newcommand{\NewToMiVsix}{66.0}
\newcommand{\OvEmpty}{70.1}
\newcommand{\OvPercepToMSCsix}{66.5}
\newcommand{\OvSimToMSC}{63.9}

\newcommand{\PTFBtwoPercepToMGold}{72.7}
\newcommand{\PTFBtwoPercepToMModel}{1.9}

\newcommand{\PTPercepToMGold}{59.3}
\newcommand{\PTPercepToMModel}{48.8}
\newcommand{\PTRsipGold}{81.0}
\newcommand{\PTRsipModel}{63.3}
\newcommand{\PTSCGold}{60.4}
\newcommand{\PTSCModel}{59.8}

\newcommand{\PTTBtwoPercepToMGold}{19.0}
\newcommand{\PTTBtwoPercepToMModel}{47.2}
\newcommand{\PTTBtwoRsipGold}{59.1}
\newcommand{\PTTBtwoRsipModel}{11.2}

\newcommand{\PTTBtwoVsixGold}{35.4}
\newcommand{\PTTBtwoVsixModel}{2.8}
\newcommand{\PTVsixGold}{70.3}
\newcommand{\PTVsixModel}{60.5}

\newcommand{\RestMainGain}{1.2}
\newcommand{\RsipMinMain}{70.9}

\newcommand{\ShareConfFS}{85}
\newcommand{\ShareMainFS}{81}

\newcommand{\RCFAO}{91.2}

\newcommand{\RCFVC}{96.4}

\newcommand{\RCohortSD}{2.7}

\newcommand{\RDevHand}{59.7}
\newcommand{\RDevLoop}{63.8}

\newcommand{\RExploreSlot}{84.7}

\newcommand{\RGEPAWordsHi}{2{,}383}
\newcommand{\RGEPAWordsLo}{1{,}838}
\newcommand{\RGapAO}{0.8}
\newcommand{\RGapLoop}{1.1}
\newcommand{\RGapVC}{1.8}
\newcommand{\RGapVCs}{1.0}

\newcommand{\RHarmAO}{16.7}
\newcommand{\RHarmLoop}{31.8}

\newcommand{\RInhAC}{+0.6}
\newcommand{\RInhMA}{$-$0.2}
\newcommand{\RInhMC}{+0.5}
\newcommand{\RInhciAC}{[$-$2.0, +4.1]}
\newcommand{\RInhciMA}{[$-$3.1, +2.2]}
\newcommand{\RInhciMC}{[$-$0.5, +1.5]}

\newcommand{\RMainGEPA}{67.0}

\newcommand{\RMainLoop}{71.8}

\newcommand{\RMainVCs}{71.9}

\newcommand{\RNAdmAO}{6}
\newcommand{\RNAdmLoop}{22}
\newcommand{\RNAdmVC}{12}
\newcommand{\RNAdmVCs}{6}
\newcommand{\RNCandAO}{94}
\newcommand{\RNCandLoop}{192}
\newcommand{\RNCandVC}{191}
\newcommand{\RNCandVCs}{96}

\newcommand{\RNRoundsLoop}{48}

\newcommand{\RNullAdm}{70}
\newcommand{\RPrecAO}{83.3}
\newcommand{\RPrecLoop}{63.6}

\newcommand{\RViewHiPercep}{+6.3}
\newcommand{\RViewHiSimToM}{+3.7}
\newcommand{\RViewPercep}{+1.3}
\newcommand{\RViewSimToM}{+0.9}
\newcommand{\RViewciPercep}{[+0.7, +2.0]}
\newcommand{\RViewciSimToM}{[+0.1, +1.7]}

\newcommand{\RciCFVCAO}{[+3.6, +6.8]}

\newcommand{\RciVCHand}{[$-$1.9, $-$0.6]}

\newcommand{\RdCFVCAO}{+5.1}

\newcommand{\RdVCHand}{$-$1.3}
\newcommand{\RdVCLoop}{0.0}

\newcommand{\QActFlipIntact}{94.9}
\newcommand{\QActFlipNoCond}{93.9}
\newcommand{\QActFlipSwap}{94.4}
\newcommand{\QActIntact}{96.6}
\newcommand{\QActNoCond}{96.0}
\newcommand{\QActSwap}{96.3}
\newcommand{\QActciNoCond}{[$-$1.6, +0.4]}
\newcommand{\QActciSwap}{[$-$1.2, +0.7]}
\newcommand{\QActdNoCond}{$-$0.5}
\newcommand{\QActdSwap}{$-$0.2}
\newcommand{\QAdmAO}{5/5}
\newcommand{\QAdmLoop}{14/14}
\newcommand{\QAdmVC}{7/9}
\newcommand{\QAdmVCs}{3/4}
\newcommand{\QAudAmb}{84}

\newcommand{\QAudInv}{286}
\newcommand{\QAudPreAmb}{270}
\newcommand{\QAudPreInv}{165}
\newcommand{\QAudPreVac}{109}
\newcommand{\QAudPreVal}{29}
\newcommand{\QAudPropCI}{[80.7, 92.2]}
\newcommand{\QAudPropShare}{86.7}
\newcommand{\QAudTexts}{573}
\newcommand{\QAudTrunc}{253}
\newcommand{\QAudTruncLeft}{43}
\newcommand{\QAudVac}{159}
\newcommand{\QAudVal}{44}

\newcommand{\QBUFlipAO}{95.5}
\newcommand{\QBUFlipB}{95.1}
\newcommand{\QBUFlipGEPA}{70.5}
\newcommand{\QBUFlipHand}{91.2}

\newcommand{\QBUFlipNone}{90.2}
\newcommand{\QBUFlipSCnine}{88.5}
\newcommand{\QBUFlipVC}{94.9}

\newcommand{\QBUGEPA}{84.4}
\newcommand{\QBUHand}{94.5}

\newcommand{\QBUNone}{94.0}
\newcommand{\QBUSCnine}{93.0}

\newcommand{\QBUSameNone}{97.8}

\newcommand{\QBUSameVC}{98.2}

\newcommand{\QBUVC}{96.6}

\newcommand{\QBUciVCAO}{[$-$1.4, +0.8]}
\newcommand{\QBUciVCB}{[$-$1.1, +0.9]}
\newcommand{\QBUciVCHand}{[+1.0, +3.5]}

\newcommand{\QBUciVCNone}{[+1.4, +3.9]}

\newcommand{\QBUdAONone}{+2.8}

\newcommand{\QBUdVCAO}{$-$0.2}
\newcommand{\QBUdVCB}{$-$0.1}
\newcommand{\QBUdVCHand}{+2.1}

\newcommand{\QBUdVCNone}{+2.6}

\newcommand{\QBWAO}{68.1}

\newcommand{\QBWFlipVC}{37.2}
\newcommand{\QBWHand}{68.1}
\newcommand{\QBWNone}{67.8}
\newcommand{\QBWVC}{68.0}

\newcommand{\QBWciVCNone}{[$-$2.5, +2.8]}

\newcommand{\QBWdVCNone}{+0.2}
\newcommand{\QBound}{0.84}
\newcommand{\QConfNCt}{176}
\newcommand{\QConfNCx}{1139}

\newcommand{\QDevCB}{+2.7}
\newcommand{\QDevCBci}{[$-$1.8, +6.6]}
\newcommand{\QGainCI}{[$-$0.82, +1.32]}
\newcommand{\QGainDiff}{+0.24}
\newcommand{\QGainInv}{$-$1.47}
\newcommand{\QGainVal}{$-$1.22}
\newcommand{\QGateA}{74.8}
\newcommand{\QGateAHi}{84}
\newcommand{\QGateALo}{67}
\newcommand{\QGateB}{79.2}
\newcommand{\QGateC}{78.7}
\newcommand{\QInducedCt}{2.8}
\newcommand{\QInducedCx}{8.6}

\newcommand{\QRepGate}{88.2}

\newcommand{\QRepOne}{84.8}

\newcommand{\QRepThree}{89.3}
\newcommand{\QRepThreeDec}{28}
\newcommand{\QRepThreeValid}{3}

\newcommand{\QSrcInv}{286}
\newcommand{\QSrcOut}{116}
\newcommand{\QTV}{0.16}
\newcommand{\QWrongWithoutCt}{35.2}
\newcommand{\QWrongWithoutCx}{29.4}

\newcommand{\cogrsi}{CogRSI}
\newcommand{\reader}{CogRSI}
\newcommand{\readerL}{CogRSI-L}
\newcommand{\libA}{Lib-A}
\newcommand{\libC}{Lib-C}
\newcommand{\sys}{\reader}

\title{CogRSI: Evolving and Auditing Cognitive Reasoning Strategies}

\author{Yandan Zheng\textsuperscript{1} \quad Haoran Luo\textsuperscript{1,$\dagger$} \quad Xinliang Zhou\textsuperscript{2} \quad Anh Tuan Luu\textsuperscript{1,3}\\
\textsuperscript{1}Nanyang Technological University, Singapore \quad \textsuperscript{2}Stanford University \quad \textsuperscript{3}VinUniversity, Vietnam\\
\texttt{\{yandan.zheng, haoran.luo, anhtuan.luu\}@ntu.edu.sg} \quad \texttt{xinliang.zhou@stanford.edu}}

\ifluatex\directlua{dofile('lastline.lua')}\fi
\begin{document}
\maketitle
{\renewcommand{\thefootnote}{$\dagger$}\footnotetext{Corresponding author.}}

\begin{abstract}
Can a language model learn better ways to reason about other people's minds by improving its own strategies, and are
the strategies it learns actually right? We study this with \cogrsi{}. A grounded reader marks who saw each event in a
story and uses a short written principle to settle disagreements between its readings. With a principle written by
experts, it beats self-consistency at the same token cost (\OvVsix{} vs.\ \OvSCnine{}). A self-improvement loop then writes
the principle itself. It keeps an update only if accuracy rises by a calibrated margin, and in one variant only if
it also passes counterfactual test pairs. The learned principles transfer: on 800 BigToM pairs that no system
had seen, each of them scores higher than the expert principle. We then check what the learned rules actually say.
Each rule is translated into code and run on thousands of small stories with known answers. Most of them
(\QAudPropShare{}\%) give a wrong answer in at least one situation they claim to cover, for example by treating a
character who is in the room as knowing what happened even while asleep. Accuracy does not reveal which rules are
wrong, and the counterfactual test does not fix this. A strategy can help in practice and still be wrong as a rule,
so both need checking. Our code and data are publicly available.\footnote{\url{https://github.com/koanatakiyo/cogRSI}}
\end{abstract}

\section{Introduction}
\begin{figure}[t]
\centering
\includegraphics[width=\columnwidth]{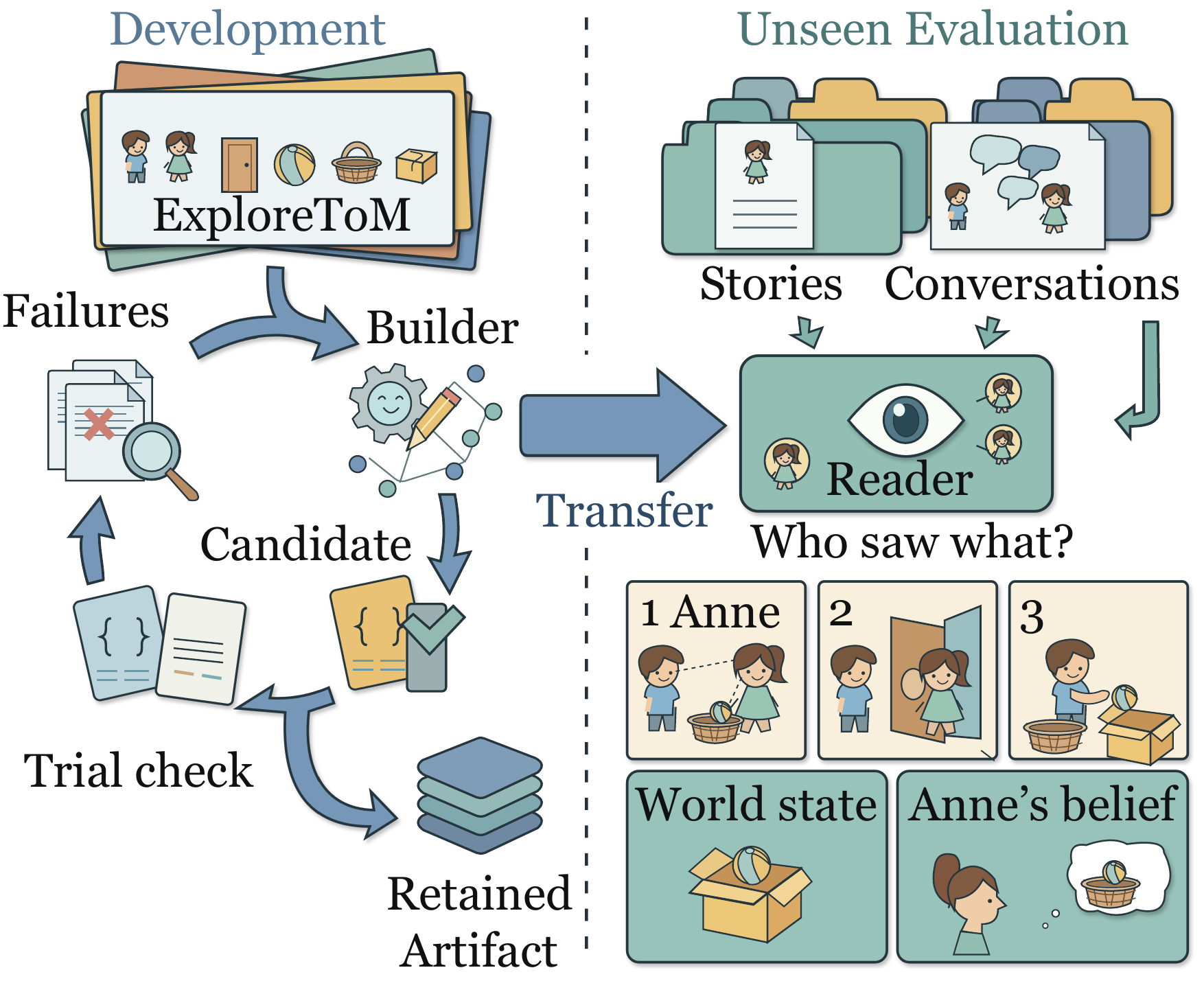}
\caption{\textbf{The task}: a strategy learned on one generator must track who saw what on benchmarks it never saw.}
\label{fig:teaser}
\end{figure}

Recursive self-improvement (RSI) lets a model change its own artifacts, such as prompts, tools or code, and keep the
changes that help \citep{wang2023voyager,yin2025godel,zhang2025dgm}. Reasoning about other minds is a hard test for it.
Theory of mind (ToM) is the clearest case: a system has to track what each person saw, was told and believes.
Benchmarks probe these states in stories and conversations (Figure~\ref{fig:teaser}), and a skill that looks general can
rest on surface cues \citep{ullman2023large,shapira2024clever}.

\begin{figure*}[t]
\centering
\includegraphics[width=\textwidth]{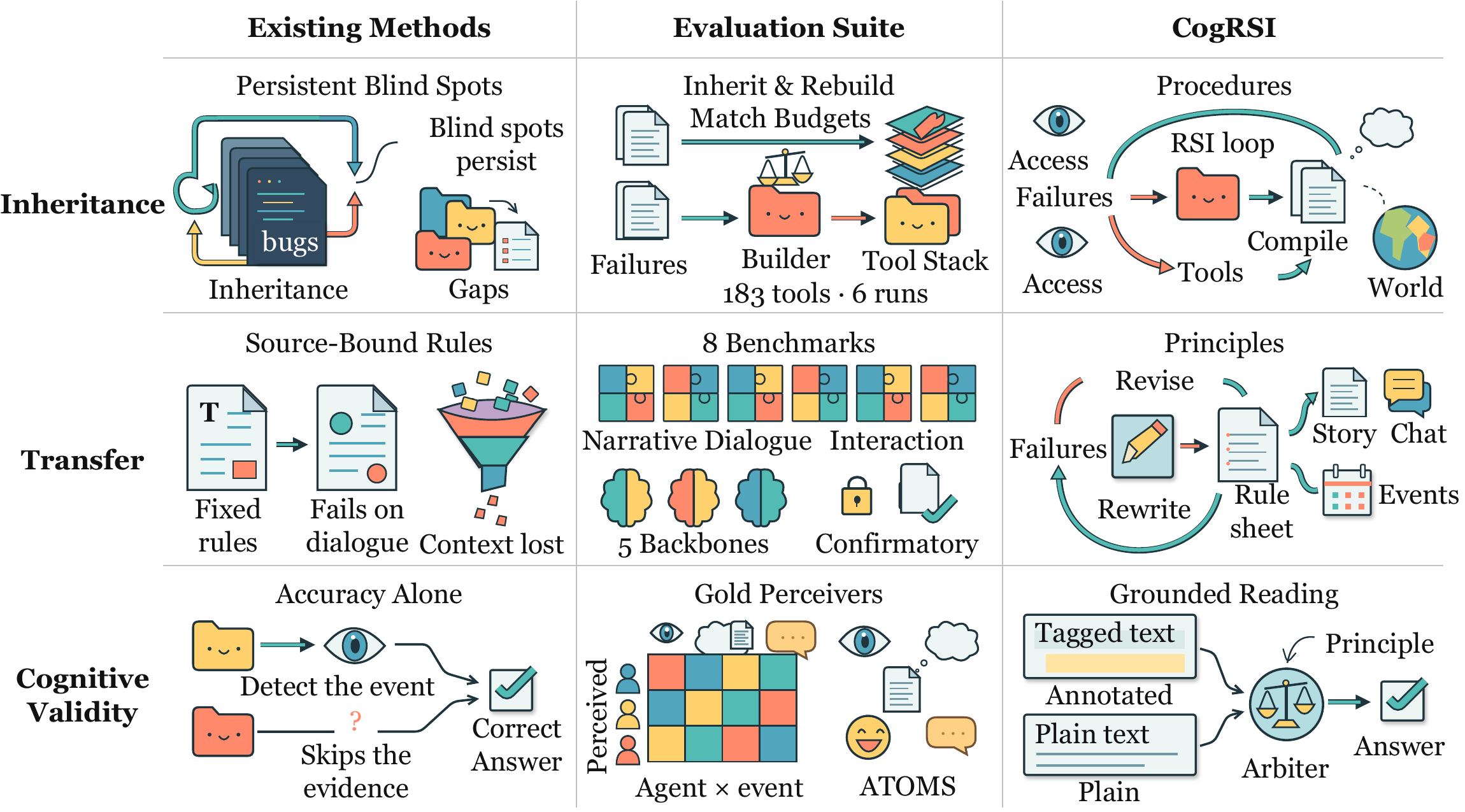}
\caption{\textbf{Overview.} Rows: what persists, transfer, cognitive validity. Existing self-improving and ToM
methods leave them open. Our suite measures them, and \cogrsi{} addresses them.}
\label{fig:overview}
\end{figure*}

Self-improving systems keep an artifact when it raises a score \citep{wang2023voyager,zhao2024expel,agrawal2025gepa}.
For reasoning about minds, two questions stay open. Does what the loop learns transfer beyond the data it learned from?
And is a strategy that helps also correct? Take the rule ``a character who is in the room knows what happens''. It
raises accuracy whenever being there means seeing. It gives the wrong answer when a character is in the room but asleep.

We study both questions with \cogrsi{} and make three contributions. \textbf{(C1) A grounded reader.} It marks who
perceived each sentence or utterance, reads the story with and without these marks, and uses a short written principle
to settle disagreements. \textbf{(C2) Counterfactual-guided strategy evolution.} A loop writes that principle as a
small library of rules, each with a condition for when it applies. It keeps an update only if accuracy rises by a
calibrated margin. In one variant, the update must also pass counterfactual pairs, two versions of a story that
differ in one fact about who could see what. \textbf{(C3) A semantic audit.} We check what each learned rule says. Two
independent models translate the rule into code, and we run the code on thousands of small stories whose answers
follow from who saw or was told what. A rule is \emph{invalid} if both translations give the same wrong answer on some
story. In other words, following the rule as written leads to a wrong belief in a situation it claims to cover.

The learned strategies transfer. On 800 BigToM pairs that no system had seen, every learned principle scores higher
than the expert principle (\S\ref{sec:rq2}). Yet \QAudPropShare{}\% of the learned rules are invalid. Accuracy does
not reveal which ones, and the counterfactual gate does not change this (\S\ref{sec:rq4}). A self-improving system for
reasoning about minds therefore needs two checks: how its strategies behave, and what they say.

\section{Related Work}
We review the closest lines of work. Appendix~\ref{app:related} gives more detail and a comparison table.

\paragraph{Self-improving systems.} Agents keep skills \citep{wang2023voyager}, insights \citep{zhao2024expel} or
workflows \citep{wang2025awm}. They improve their scaffolding \citep{zelikman2024stop,yin2025godel,robeyns2025sica},
design agents \citep{hu2025adas,zhang2025dgm} and evolve prompts and code
\citep{fernando2024promptbreeder,agrawal2025gepa,novikov2025alphaevolve}. Recent work regularises these edits
\citep{wu2026modularrsi,xia2026rrsi} and checks a skill before keeping it. The check can be a replay on held-out tasks
\citep{shang2026vag}, a counterexample to the edited assumption \citep{nie2026skillevoreg}, or a test of the rule as a
falsifiable hypothesis \citep{wang2026evobelief}. Audits find self-scored gains that do not hold
\citep{guo2026seal,xu2026phantom}, edits that only look like gains \citep{wang2026tampering}, selections that miss
held-out gains \citep{lin2026evopathbench}, and skill libraries that add little over plain context
\citep{guan2026continualskillbench}. All of these checks measure task success. We also check what an admitted rule
says, against a formal model of the domain.

\paragraph{Theory of mind and formal checks.} ToM methods take a character's perspective by filtering or prompting
\citep{wilf2024simtom,jung2024perceptom,sarangi2025decompose,shinoda2026shoes}. Others track beliefs symbolically or
over time \citep{sclar2023symbolictom,hou2024timetom}, verify belief traces \citep{wu2025deltom}, or use model-based
inference \citep{zhang2025autotom,kim2025thoughttracing}. Training with reinforcement learning \citep{lu2025tomrl} does
not transfer \citep{sarangi2025rltom}. None of these methods learns its strategy from failures. Our counterfactual pairs
follow contrast sets \citep{gardner2020contrast,kaushik2020learning}. Translating natural language into checkable logic
follows nl2spec \citep{cosler2023nl2spec}, and we apply the idea to strategies that an RSI loop has written.

\section{Task and Evaluation}
\label{sec:task}
Appendix~\ref{app:bench} details the evaluation suite (Figure~\ref{fig:suite}).

\paragraph{Task.} A context $c=(u_1,\dots,u_n)$ is a story of sentences or a conversation of utterances about agents
$\mathcal{A}$. A question $q$ asks about the mental state of one or more agents, and a system answers $\hat y=F(q,c)$ in
the benchmark's format. In artifact-level RSI, a builder model reads the current system's failures on development data
$\mathcal{D}$ and proposes an artifact $x$. Following MedRSI \citep{wu2026medrsi}, $x$ is registered (kept) if its mean
accuracy gain over $K$ fresh trial cohorts exceeds a margin $\delta$,
\begin{equation}
\textstyle \mathrm{reg}(x)\iff\frac{1}{K}\sum_{k=1}^{K}\Delta A_k(x)>\delta.
\label{eq:reg}
\end{equation}
All of $\mathcal{D}$ comes from one story generator, ExploreToM \citep{sclar2024exploretom}. The loop sees no other data.

\paragraph{Three kinds of evidence.} \textbf{Task accuracy} comes from eight benchmarks: FANToM \citep{kim2023fantom},
SimpleToM \citep{gu2024simpletom}, ExploreToM, Hi-ToM \citep{he2023hitom}, OpenToM \citep{xu2024opentom}, ToMBench
\citep{chen2024tombench}, CogToM \citep{tong2026cogtom} and BigToM \citep{gandhi2023bigtom}. The main test sample has
\SuiteMainQ{} questions. A confirmatory sample of \SuiteHthreeQ{} questions was drawn after every design decision, and
seven more benchmarks are used for testing only. \textbf{Behavioural validity} comes from counterfactual pairs. A pair
counts only if both versions are answered right. Our own generator gives a 96-pair gate split, the only one any loop
sees, and a 320-pair evaluation split. Its variants cover a character who watches, leaves, sleeps, is told, is lied to
or looks later. \textbf{BigToM-U} adds 800 independent pairs from BigToM conditions that no system had answered. In a
flip pair the agent does or does not see a change. In a same pair the agent does or does not see an irrelevant event.
\textbf{Semantic validity} asks whether a learned rule is correct (\S\ref{sec:audit}). Each benchmark scores answers
with its own rule, and \textbf{Main} is the mean over the eight benchmarks. Differences use a paired bootstrap over
stories, resampling seeds and scenarios for the loops.

\section{Methodology: \cogrsi{}}
\label{sec:method}
\begin{figure*}[t]
\centering
\includegraphics[width=\textwidth]{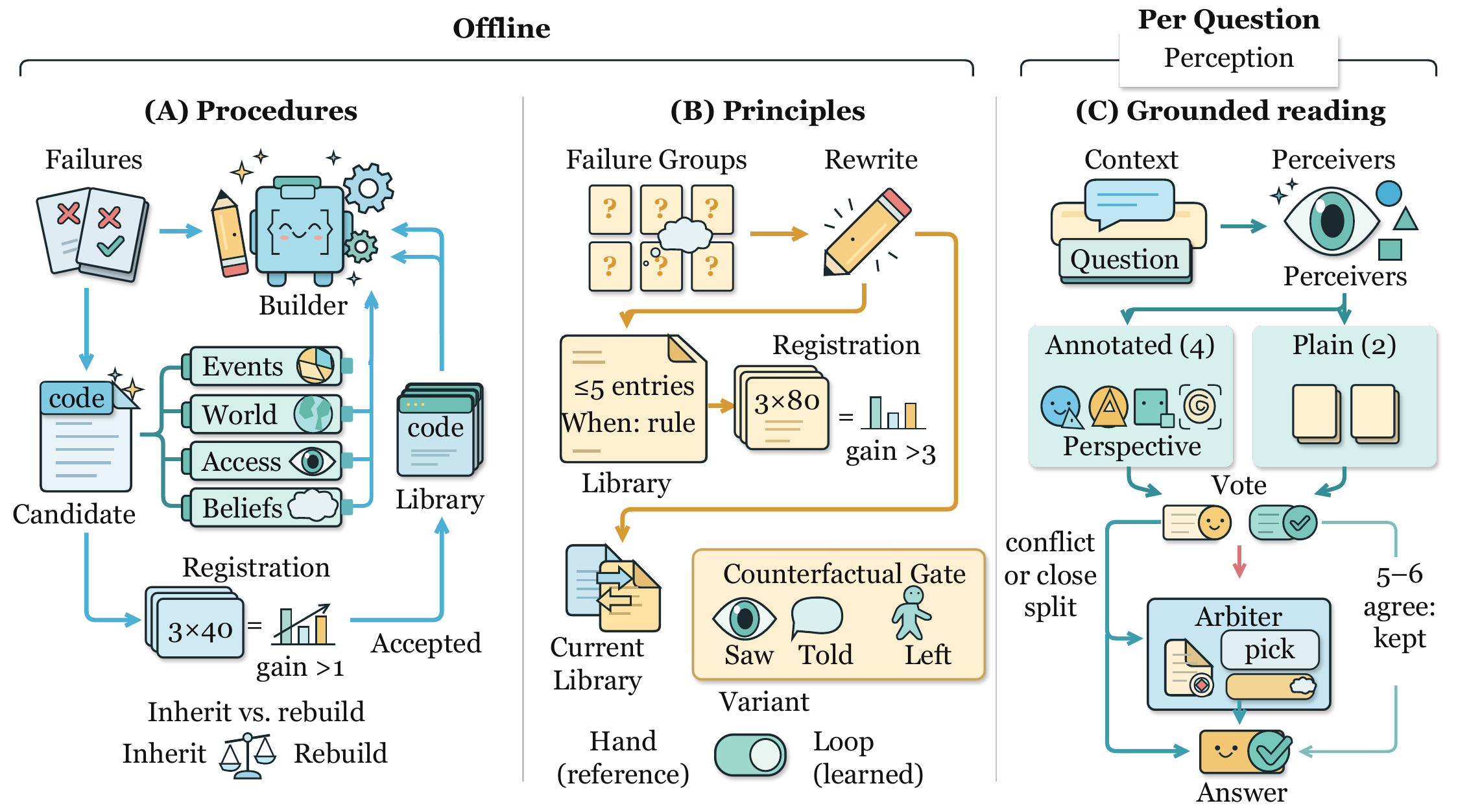}
\caption{\textbf{\cogrsi{}.} (A) RSI over procedures writes tools that export perceptual access. (B) RSI over principles
keeps a library of conditioned strategies and admits an update only if it passes the accuracy margin and, in the
gated variant, the counterfactual gate. (C) The reader applies the strategies at test time. The semantic audit
(\S\ref{sec:audit}) checks what the admitted strategies assert.}
\label{fig:method}
\end{figure*}
\cogrsi{} has three parts: a reader, loops that write principles for it, and an audit of what the loops write
(Figure~\ref{fig:method}, Algorithms~\ref{alg:rsi} and~\ref{alg:reader}).

\subsection{The Grounded Reader (C1)}
\label{sec:principle}\label{sec:arbitration}
One perception call per context labels each sentence or utterance with the characters who witnessed it. It uses the
perception rule of \citet{jung2024perceptom} with indexed outputs. PercepToM deletes what a character did not see. Our
reader keeps the whole story and adds who saw what. It draws six samples at temperature 0.7, the budget of
self-consistency with $k{=}6$. Four samples read the annotated story, or the questioned character's view for
conversations, and two read the plain story. If the two readings conflict or the samples split closely, one call picks
between the two leading answers under a \emph{principle} $P$. It never overturns an answer held by five or six samples
(Proposition~\ref{prop:local}). \reader{} uses an expert principle: \emph{a character knows only what the story shows
them witnessing, being told, or working out from what they themselves did}. The loops below write $P$ instead.

\subsection{Counterfactual-Guided Strategy Evolution (C2)}
\label{sec:loop}\label{sec:loop2}\label{sec:vc}
\paragraph{Strategies.} The loop keeps a library $L$ of up to five rules. Each rule has the form \emph{When}
condition\emph{:} rule, with at most 60 words. The library fills the principle slot of the reader. In each round the
builder sees $L$ and up to four failures from one group of 80 new development questions. It then proposes one change:
\textsc{add}, \textsc{revise} or \textsc{retire} a rule. The builder is never told what makes a rule valid. Up to four
candidate changes compete per round.

\paragraph{Calibrated registration.} The same principle varies by \RCohortSD{} points across trial cohorts of 80
questions. With four candidates per round, a 1-point margin keeps a candidate with no real gain in \RNullAdm{}\% of
rounds. We therefore set $\delta{=}3$ points and also require a gain of at least zero on two of the three cohorts.

\paragraph{Counterfactual guidance.} The gated variant also asks a candidate to keep the 96 gate pairs right. Let
$\Delta V$ be the change in correctly answered gate pairs and $\overline{\Delta A}$ the mean accuracy gain. Then
\begin{equation}
\begin{aligned}
\mathrm{reg}_{\mathrm{G}}\iff{}&\big[\mathrm{reg}\wedge \Delta V\ge-3\big]\\
\vee{}&\big[\Delta V\ge 6\wedge \overline{\Delta A}\ge-0.5\big],
\end{aligned}
\label{eq:vc}
\end{equation}
where the tolerances come from the reader's run-to-run noise. The second term keeps a candidate that fixes several
pairs at no accuracy cost. A \emph{pairs-as-data} variant uses the same 192 gate items as ordinary trial items with
accuracy-only registration. It sees the same information without the pair constraint. We also run an accuracy-only
library, a gated single principle, an accuracy-only loop with a 1-point margin that writes one principle, and GEPA
\citep{agrawal2025gepa} in the same slot. Appendix~\ref{app:algo-proc} reports a loop that writes executable tools.

\subsection{Semantic Audit (C3)}
\label{sec:audit}
\paragraph{A formal model of access.} The audit needs a ground truth for what characters believe. We use the account
that seeing leads to knowing \citep{wimmer1988access}, plus testimony. In one room, agents leave and return, fall asleep
and wake, move an object, whisper its location truly or falsely, peek into a container, or change a visible or hidden
property of the object. An event updates the beliefs of everyone who perceives it, and a whisper reaches only its
listener. A fixed sample of 6{,}000 small worlds gives about 85k questions with known answers.

\paragraph{Translation and counterexamples.} Two models translate each rule into code: Llama-3.3-70B and
Qwen3.5-122B, neither of which wrote the rules. Each writes \texttt{applies(world, q)} for the rule's condition and
\texttt{predict(world, q)} for its answer. They are told to follow the words literally and do not know which loop wrote
the rule or whether it was kept. A \emph{shared counterexample} is a question where both translations give the same
answer and the formal model gives a different one. Two independent readings then agree on what the rule says, and it is
wrong. A rule is \textbf{invalid} if it has a shared counterexample. It is \textbf{valid} if it has none and the two
translations agree on at least 80\% of questions. It is \textbf{vacuous} if it almost never applies and
\textbf{ambiguous} otherwise. Finally, the reader answers rendered counterexamples with and without the rule, to see
whether they change its behaviour.

\section{Experiments}
\label{sec:exp}
\begin{table*}[t]
\centering
\footnotesize
\setlength{\tabcolsep}{3pt}
\begin{tabular*}{\textwidth}{@{\extracolsep{\fill}}llccccc@{}}
\toprule
 & & \multicolumn{2}{c}{Accuracy} & \multicolumn{2}{c}{Pair validity} & Audit \\
\cmidrule(lr){3-4}\cmidrule(lr){5-6}\cmidrule(l){7-7}
Principle written by & Registration & Dev. & Test Main & Own pairs & BigToM-U & Invalid rules (\%) \\
\midrule
no principle & none & 59.7 & 70.1 & 91.7 & 94.0 & n.a. \\
hand (expert) & none & 59.7 & 73.1 & 91.7 & 94.5 & n.r. \\
\midrule
accuracy-only loop & 1-pt & 63.8 & 71.8 & 91.5 & 95.9 & 89.7 \\
GEPA (same slot) & Pareto & own val. & 67.0 & 75.3 & 84.4 & n.r. \\
accuracy-only library & 3-pt & 66.1 & 71.5$^\dagger$ & 91.2$^\dagger$ & 96.8 & 85.2$^\dagger$ \\
pairs as data, library & 3-pt & 63.1 & n.r. & n.r. & 96.6 & n.r. \\
gated single principle & 3-pt + gate & 62.1 & 71.9 & 98.1 & 96.5 & 82.7 \\
\rowcolor{ours} gated library & 3-pt + gate & 65.8 & 71.8 & 96.4 & 96.6 & 86.0 \\
\bottomrule
\end{tabular*}
\caption{\textbf{Who writes the principle.} Same reader, backbone and sampling; only the principle changes. Registration: accuracy margin in points, with the counterfactual gate where stated (Pareto: GEPA's own selection). Dev.: 300 unseen ExploreToM development questions per seed. Test Main: main test sample, eight benchmarks. Own pairs: 320 held-out pairs from our generator. BigToM-U: 800 BigToM pairs in conditions no system had seen. Invalid rules: share of the arm's proposed rules with a formal counterexample (\S\ref{sec:audit}). Seeds: 6 for loops and libraries, 3 for GEPA and the single principle; $^\dagger$seeds 0--2 only. n.r.: not run; n.a.: not applicable. \textbf{Every learned principle beats the expert one on BigToM-U; the gate changes validity on its own pairs only.}}
\label{tab:arms}
\end{table*}

We ask four questions. \textbf{RQ1}: How strong is the reader? \textbf{RQ2}: Do evolved strategies transfer?
\textbf{RQ3}: What does registration buy? \textbf{RQ4}: What do evolved strategies say?

\paragraph{Setup.} The main model is Qwen3.5-35B-A3B (FP8) \citep{qwen2026qwen35}. The reader also runs on
Qwen3.5-122B-A10B, Llama-3.3-70B-Instruct \citep{grattafiori2024llama3}, GLM-4.6V-Flash \citep{glmv2025} and
Qwen3.5-9B, served with vLLM \citep{kwon2023vllm} with thinking off. Baselines are zero-shot, chain of thought (CoT)
\citep{kojima2022large}, self-consistency (SC) \citep{wang2023selfconsistency} with $k{=}6$ and with $k{=}9$ (the
reader's token budget), retrieval, prompt-only GEPA, SimToM \citep{wilf2024simtom} and PercepToM
\citep{jung2024perceptom}. SimToM and PercepToM also run with our six samples. Each loop runs eight rounds on
ExploreToM development data. Every experiment after the design freeze was pre-registered with its hypotheses and
thresholds (Appendix~\ref{app:vc}), and we report every outcome.

\begin{figure*}[t]
\centering
\input{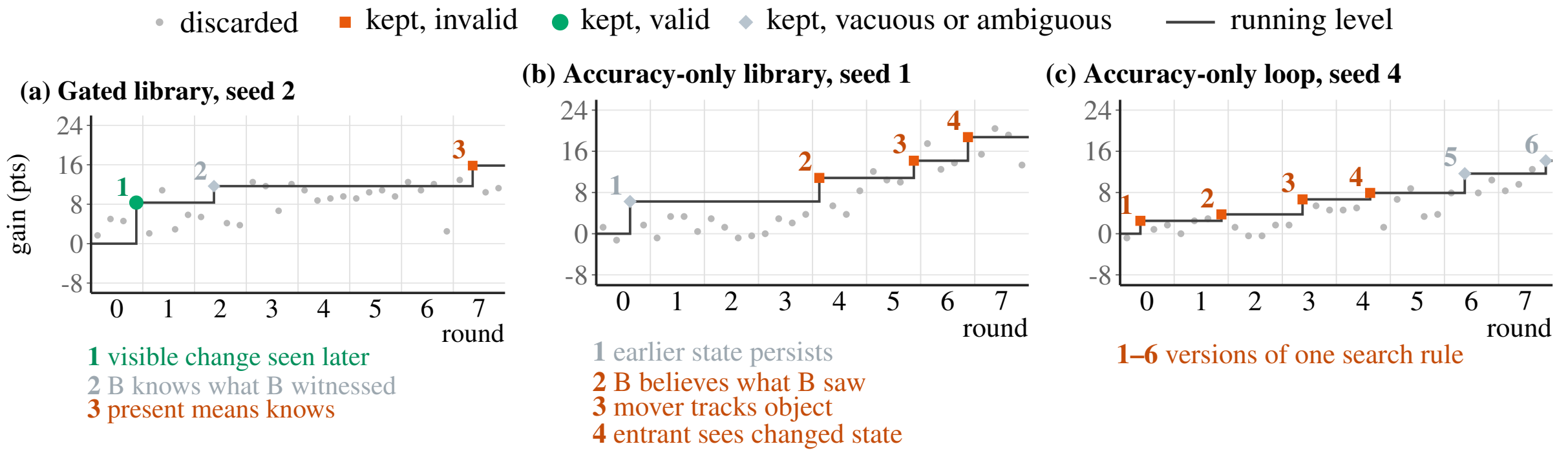}
\caption{\textbf{Registration as the loop sees it.} Every candidate that reached registration, in order (four per
round), placed at the incumbent's running level plus its mean trial gain. Kept candidates are coloured by the semantic
audit (\S\ref{sec:audit}). Levels are the loops' own trial estimates, which overstate held-out gains
(Table~\ref{tab:selection}). \textbf{Each loop keeps climbing on its own trials, and most of what it keeps is invalid.}}
\label{fig:progress}
\end{figure*}
\begin{figure*}[t]
\centering
\input{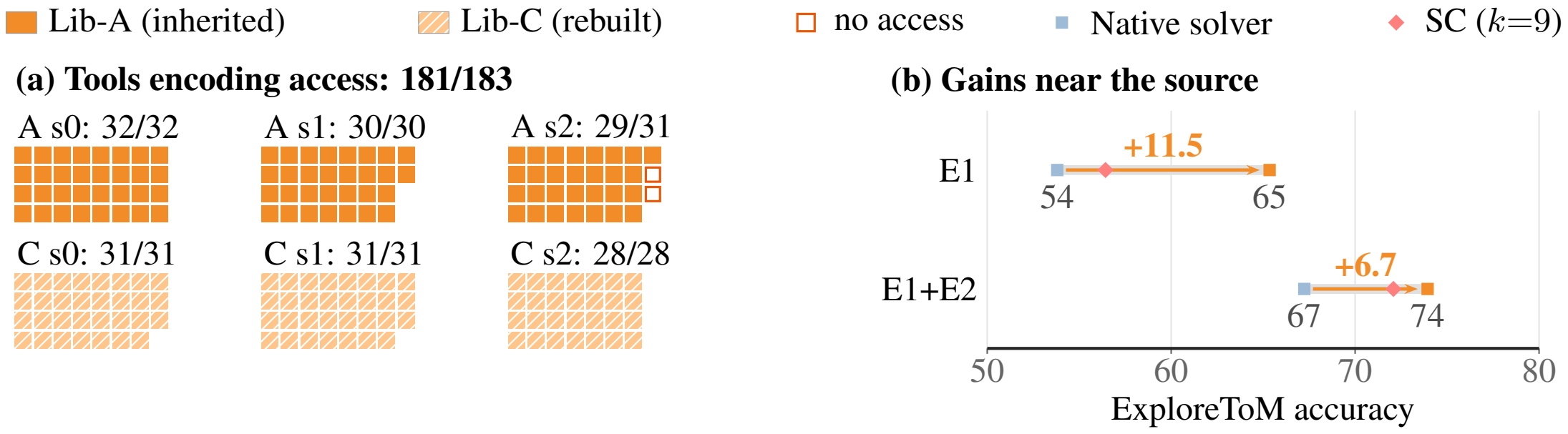}
\caption{\textbf{RSI over procedures.} (a) Candidate tools per E1 run, filled if they encode access. (b) ExploreToM accuracy, native solver $\to$ \libA{}. \textbf{Every run fills the access slot, and its library helps only near its source.}}
\label{fig:rsi}
\end{figure*}
\begin{figure*}[t]
\centering
\input{tables/fig_vc}
\caption{\textbf{Validity is a separate target.} (a) Test Main against pair validity on the 320 held-out counterfactual
pairs, means over seeds or runs. (b) Accuracy by access variant on the same items. \textbf{Accuracy-only
principles cluster near the reader with no principle, GEPA falls off both axes, and the gated library moves up but not
right.}}
\label{fig:vc}
\end{figure*}
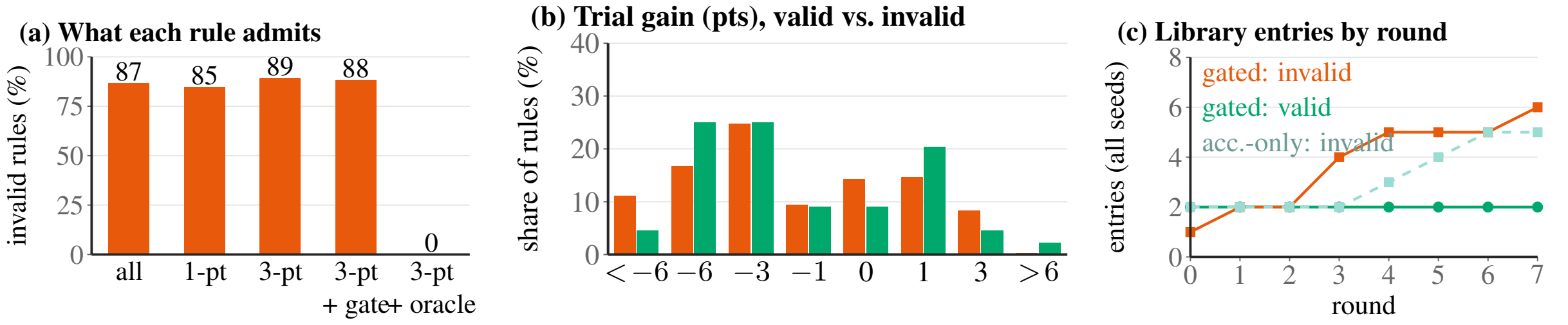
\begin{figure*}[t]
\centering
\begin{minipage}[b]{0.31\linewidth}\centering\fitbox{\begin{tikzpicture}[font=\figfont, x=1cm, y=1cm]
\draw[black!10] (0,0.000) -- (4.400,0.000);
\draw[black!60] (0,0.000) -- (-0.05,0.000) node[left, inner sep=1pt] {0};
\draw[black!10] (0,0.575) -- (4.400,0.575);
\draw[black!60] (0,0.575) -- (-0.05,0.575) node[left, inner sep=1pt] {25};
\draw[black!10] (0,1.150) -- (4.400,1.150);
\draw[black!60] (0,1.150) -- (-0.05,1.150) node[left, inner sep=1pt] {50};
\draw[black!10] (0,1.725) -- (4.400,1.725);
\draw[black!60] (0,1.725) -- (-0.05,1.725) node[left, inner sep=1pt] {75};
\draw[black!10] (0,2.300) -- (4.400,2.300);
\draw[black!60] (0,2.300) -- (-0.05,2.300) node[left, inner sep=1pt] {100};
\node[rotate=90, anchor=south] at (-0.6,1.150) {invalid rules (\%)};
\node[anchor=south west, font=\figfont\bfseries, inner sep=0pt] at (-0.85,2.420) {(a) What each rule admits};
\fill[cNeg] (0.198,0.000) rectangle (0.670,1.994);
\node[font=\figfont, anchor=south, inner sep=0.6pt] at (0.434,1.994) {87};
\node[anchor=north, inner sep=1.5pt, align=center] at (0.440,-0.02) {all};
\fill[cNeg] (1.078,0.000) rectangle (1.550,1.950);
\node[font=\figfont, anchor=south, inner sep=0.6pt] at (1.314,1.950) {85};
\node[anchor=north, inner sep=1.5pt, align=center] at (1.320,-0.02) {1-pt};
\fill[cNeg] (1.958,0.000) rectangle (2.430,2.054);
\node[font=\figfont, anchor=south, inner sep=0.6pt] at (2.194,2.054) {89};
\node[anchor=north, inner sep=1.5pt, align=center] at (2.200,-0.02) {3-pt};
\fill[cNeg] (2.838,0.000) rectangle (3.310,2.029);
\node[font=\figfont, anchor=south, inner sep=0.6pt] at (3.074,2.029) {88};
\node[anchor=north, inner sep=1.5pt, align=center] at (3.080,-0.02) {3-pt\\+ gate};
\fill[cNeg] (3.718,0.000) rectangle (4.190,0.000);
\node[font=\figfont, anchor=south, inner sep=0.6pt] at (3.954,0.000) {0};
\node[anchor=north, inner sep=1.5pt, align=center] at (3.960,-0.02) {3-pt\\+ oracle};
\draw[black!85, line width=0.8pt] (0,0) -- (0,2.300); \draw[black!85, line width=0.8pt] (0,0) -- (4.400,0);
\path (0,-0.75) (0,2.75);
\end{tikzpicture}}\end{minipage}\hfill
\begin{minipage}[b]{0.36\linewidth}\centering\fitbox{\begin{tikzpicture}[font=\figfont, x=1cm, y=1cm]
\draw[black!10] (0,0.000) -- (5.000,0.000);
\draw[black!60] (0,0.000) -- (-0.05,0.000) node[left, inner sep=1pt] {0};
\draw[black!10] (0,0.575) -- (5.000,0.575);
\draw[black!60] (0,0.575) -- (-0.05,0.575) node[left, inner sep=1pt] {10};
\draw[black!10] (0,1.150) -- (5.000,1.150);
\draw[black!60] (0,1.150) -- (-0.05,1.150) node[left, inner sep=1pt] {20};
\draw[black!10] (0,1.725) -- (5.000,1.725);
\draw[black!60] (0,1.725) -- (-0.05,1.725) node[left, inner sep=1pt] {30};
\draw[black!10] (0,2.300) -- (5.000,2.300);
\draw[black!60] (0,2.300) -- (-0.05,2.300) node[left, inner sep=1pt] {40};
\node[rotate=90, anchor=south] at (-0.6,1.150) {share of rules (\%)};
\node[anchor=south west, font=\figfont\bfseries, inner sep=0pt] at (-0.85,2.420) {(b) Trial gain (pts), valid vs.\ invalid};
\fill[cNeg] (0.062,0.000) rectangle (0.300,0.643);
\fill[cPos] (0.312,0.000) rectangle (0.550,0.261);
\node[anchor=north, inner sep=1.5pt, align=center] at (0.312,-0.02) {$<\!-6$};
\fill[cNeg] (0.688,0.000) rectangle (0.925,0.965);
\fill[cPos] (0.938,0.000) rectangle (1.175,1.438);
\node[anchor=north, inner sep=1.5pt, align=center] at (0.938,-0.02) {$-6$};
\fill[cNeg] (1.312,0.000) rectangle (1.550,1.427);
\fill[cPos] (1.562,0.000) rectangle (1.800,1.438);
\node[anchor=north, inner sep=1.5pt, align=center] at (1.562,-0.02) {$-3$};
\fill[cNeg] (1.938,0.000) rectangle (2.175,0.543);
\fill[cPos] (2.188,0.000) rectangle (2.425,0.523);
\node[anchor=north, inner sep=1.5pt, align=center] at (2.188,-0.02) {$-1$};
\fill[cNeg] (2.562,0.000) rectangle (2.800,0.824);
\fill[cPos] (2.812,0.000) rectangle (3.050,0.523);
\node[anchor=north, inner sep=1.5pt, align=center] at (2.812,-0.02) {0};
\fill[cNeg] (3.188,0.000) rectangle (3.425,0.844);
\fill[cPos] (3.438,0.000) rectangle (3.675,1.176);
\node[anchor=north, inner sep=1.5pt, align=center] at (3.438,-0.02) {1};
\fill[cNeg] (3.812,0.000) rectangle (4.051,0.483);
\fill[cPos] (4.062,0.000) rectangle (4.301,0.261);
\node[anchor=north, inner sep=1.5pt, align=center] at (4.062,-0.02) {3};
\fill[cNeg] (4.438,0.000) rectangle (4.676,0.020);
\fill[cPos] (4.688,0.000) rectangle (4.926,0.131);
\node[anchor=north, inner sep=1.5pt, align=center] at (4.688,-0.02) {$>\!6$};
\draw[black!85, line width=0.8pt] (0,0) -- (0,2.300); \draw[black!85, line width=0.8pt] (0,0) -- (5.000,0);
\path (0,-0.75) (0,2.75);
\end{tikzpicture}}\end{minipage}\hfill
\begin{minipage}[b]{0.29\linewidth}\centering\fitbox{\begin{tikzpicture}[font=\figfont, x=1cm, y=1cm]
\draw[black!10] (0,0.000) -- (4.000,0.000);
\draw[black!60] (0,0.000) -- (-0.05,0.000) node[left, inner sep=1pt] {0};
\draw[black!10] (0,0.575) -- (4.000,0.575);
\draw[black!60] (0,0.575) -- (-0.05,0.575) node[left, inner sep=1pt] {2};
\draw[black!10] (0,1.150) -- (4.000,1.150);
\draw[black!60] (0,1.150) -- (-0.05,1.150) node[left, inner sep=1pt] {4};
\draw[black!10] (0,1.725) -- (4.000,1.725);
\draw[black!60] (0,1.725) -- (-0.05,1.725) node[left, inner sep=1pt] {6};
\draw[black!10] (0,2.300) -- (4.000,2.300);
\draw[black!60] (0,2.300) -- (-0.05,2.300) node[left, inner sep=1pt] {8};
\node[rotate=90, anchor=south] at (-0.6,1.150) {entries (all seeds)};
\node[anchor=south west, font=\figfont\bfseries, inner sep=0pt] at (-0.85,2.420) {(c) Library entries by round};
\draw[black!60] (0.000,0) -- (0.000,-0.05) node[below, inner sep=1pt] {0};
\draw[black!60] (0.571,0) -- (0.571,-0.05) node[below, inner sep=1pt] {1};
\draw[black!60] (1.143,0) -- (1.143,-0.05) node[below, inner sep=1pt] {2};
\draw[black!60] (1.714,0) -- (1.714,-0.05) node[below, inner sep=1pt] {3};
\draw[black!60] (2.286,0) -- (2.286,-0.05) node[below, inner sep=1pt] {4};
\draw[black!60] (2.857,0) -- (2.857,-0.05) node[below, inner sep=1pt] {5};
\draw[black!60] (3.429,0) -- (3.429,-0.05) node[below, inner sep=1pt] {6};
\draw[black!60] (4.000,0) -- (4.000,-0.05) node[below, inner sep=1pt] {7};
\node[anchor=north] at (2.000,-0.33) {round};
\draw[cPos, line width=0.9pt] (0.000,0.575) -- (0.571,0.575) -- (1.143,0.575) -- (1.714,0.575) -- (2.286,0.575) -- (2.857,0.575) -- (3.429,0.575) -- (4.000,0.575);
\filldraw[draw=cPos, fill=cPos] (0.000,0.575) circle (1.6pt);
\filldraw[draw=cPos, fill=cPos] (0.571,0.575) circle (1.6pt);
\filldraw[draw=cPos, fill=cPos] (1.143,0.575) circle (1.6pt);
\filldraw[draw=cPos, fill=cPos] (1.714,0.575) circle (1.6pt);
\filldraw[draw=cPos, fill=cPos] (2.286,0.575) circle (1.6pt);
\filldraw[draw=cPos, fill=cPos] (2.857,0.575) circle (1.6pt);
\filldraw[draw=cPos, fill=cPos] (3.429,0.575) circle (1.6pt);
\filldraw[draw=cPos, fill=cPos] (4.000,0.575) circle (1.6pt);
\draw[cNeg, line width=0.9pt] (0.000,0.287) -- (0.571,0.575) -- (1.143,0.575) -- (1.714,1.150) -- (2.286,1.438) -- (2.857,1.438) -- (3.429,1.438) -- (4.000,1.725);
\filldraw[draw=cNeg, fill=cNeg] (-0.050,0.237) rectangle (0.050,0.337);
\filldraw[draw=cNeg, fill=cNeg] (0.521,0.525) rectangle (0.621,0.625);
\filldraw[draw=cNeg, fill=cNeg] (1.093,0.525) rectangle (1.193,0.625);
\filldraw[draw=cNeg, fill=cNeg] (1.664,1.100) rectangle (1.764,1.200);
\filldraw[draw=cNeg, fill=cNeg] (2.236,1.387) rectangle (2.336,1.488);
\filldraw[draw=cNeg, fill=cNeg] (2.807,1.387) rectangle (2.907,1.488);
\filldraw[draw=cNeg, fill=cNeg] (3.379,1.387) rectangle (3.479,1.488);
\filldraw[draw=cNeg, fill=cNeg] (3.950,1.675) rectangle (4.050,1.775);
\draw[cAO, line width=0.9pt, dashed] (0.000,0.575) -- (0.571,0.575) -- (1.143,0.575) -- (1.714,0.575) -- (2.286,0.862) -- (2.857,1.150) -- (3.429,1.438) -- (4.000,1.438);
\filldraw[draw=cAO, fill=cAO] (-0.050,0.525) rectangle (0.050,0.625);
\filldraw[draw=cAO, fill=cAO] (0.521,0.525) rectangle (0.621,0.625);
\filldraw[draw=cAO, fill=cAO] (1.093,0.525) rectangle (1.193,0.625);
\filldraw[draw=cAO, fill=cAO] (1.664,0.525) rectangle (1.764,0.625);
\filldraw[draw=cAO, fill=cAO] (2.236,0.812) rectangle (2.336,0.912);
\filldraw[draw=cAO, fill=cAO] (2.807,1.100) rectangle (2.907,1.200);
\filldraw[draw=cAO, fill=cAO] (3.379,1.387) rectangle (3.479,1.488);
\filldraw[draw=cAO, fill=cAO] (3.950,1.387) rectangle (4.050,1.488);
\node[anchor=west, text=cNeg, inner sep=2pt] at (0.057,2.099) {gated: invalid};
\node[anchor=west, text=cPos, inner sep=2pt] at (0.057,1.696) {gated: valid};
\node[anchor=west, text=cAO!70!black, inner sep=2pt] at (0.057,1.294) {acc.-only: invalid};
\draw[black!85, line width=0.8pt] (0,0) -- (0,2.300); \draw[black!85, line width=0.8pt] (0,0) -- (4.000,0);
\path (0,-0.75) (0,2.75);
\end{tikzpicture}}\end{minipage}
\caption{\textbf{Semantic audit of \QAudTexts{} evolved rules.} (a) Share of decided rules that are invalid among all
proposed rules and among those each registration rule would admit, replayed on the same candidates. (b) Trial gain of
valid and invalid rules. (c) Invalid and valid entries in the libraries by round, summed over seeds.
\textbf{Accuracy does not tell valid from invalid rules, and the gate does not change what is admitted.}}
\label{fig:audit}
\end{figure*}
\begin{figure*}[t]
\sbox{\rowboxA}{\begin{minipage}{0.578\textwidth}\centering\setlength{\tabcolsep}{2pt}%
\fitbox{%
\begin{tabular}{@{}cl*{5}{c}@{}}
\toprule
\multicolumn{2}{@{}l}{\textbf{(a) Whole system}} & \multicolumn{2}{c}{All questions} & \multicolumn{3}{c}{Narrative} \\
\cmidrule(lr){3-4}\cmidrule(l){5-7}
St. & Variant & Main & FAN & Belief & Non-lit. & Narr. \\
\midrule
1 & \libA{} library & 59.3\dn{13.8} & 2.3\dn{25.7} & 67.1\dn{15.1} & 76.3\dn{1.9} & 69.1\dn{13.0} \\
 & \libC{} (rebuilt) & 58.6\dn{14.5} & 2.7\dn{25.3} & 65.5\dn{16.6} & 76.6\dn{1.5} & 68.6\dn{13.5} \\
\midrule
2 & no principle & 70.1\dn{3.0} & 23.2\dn{4.8} & 78.6\dn{3.5} & 79.8\up{1.7} & 79.8\dn{2.4} \\
 & \readerL{} principle & 71.8\dn{1.3} & 25.3\dn{2.7} & 80.5\dn{1.6} & 78.4\up{0.3} & 80.8\dn{1.3} \\
\midrule
3 & $-$ close-split arb. & 72.6\dn{0.5} & 27.3\dn{0.7} & 81.9\dn{0.2} & 78.2\eq & 81.7\dn{0.4} \\
 & $-$ all arbitration & 72.2\dn{0.9} & 28.7\up{0.7} & 80.8\dn{1.3} & 75.5\dn{2.6} & 80.7\dn{1.4} \\
\midrule
\rowcolor{ours}  & \reader{} (full) & \textbf{73.1} & \textbf{28.0} & \textbf{82.1} & \textbf{78.2} & \textbf{82.1} \\
\midrule[\heavyrulewidth]
\multicolumn{2}{@{}l}{\textbf{(b) Stage-3 readings}} & \multicolumn{5}{c}{Narrative questions} \\
\cmidrule(l){3-7}
St. & Variant & Belief & Knowl. & Emot. & Non-lit. & Narr. \\
\midrule
3 & annotated only & 78.4\dn{3.7} & 80.1\dn{5.4} & 73.1\dn{2.6} & 72.9\dn{5.3} & 78.1\dn{4.0} \\
 & plain only & 76.9\dn{5.2} & 82.8\dn{2.7} & 77.2\up{1.4} & 80.5\up{2.4} & 78.6\dn{3.5} \\
 & vote of both & 80.8\dn{1.3} & 83.2\dn{2.4} & 74.8\dn{0.9} & 75.5\dn{2.6} & 80.7\dn{1.4} \\
\rowcolor{ours}  & \reader{} (full) & \textbf{82.1} & \textbf{85.5} & \textbf{75.7} & \textbf{78.2} & \textbf{82.1} \\
\bottomrule
\end{tabular}}
\end{minipage}}%
\setlength{\rowh}{\dimexpr\ht\rowboxA+\dp\rowboxA\relax}%
\setlength{\radunit}{1cm}\sbox{\rowboxB}{\vbox{\offinterlineskip\halign{\hfil#\hfil\cr
\begin{tikzpicture}[font=\figfont, x=1cm, y=1cm]
\draw[cOurs, line width=0.9pt] (-2.46,-0.00) -- (-2.10,-0.00);\filldraw[draw=cOurs, fill=cOurs] (-2.280,-0.000) circle (1.7pt);
\node[anchor=west, inner sep=1pt] at (-2.06,-0.00) {\reader{}};
\draw[cSCnine, line width=0.9pt] (-0.63,-0.00) -- (-0.27,-0.00);\filldraw[draw=cSCnine, fill=cSCnine] (-0.450,0.065) -- (-0.385,-0.000) -- (-0.450,-0.065) -- (-0.515,-0.000) -- cycle;
\node[anchor=west, inner sep=1pt] at (-0.23,-0.00) {best baseline};
\draw[cNative, line width=0.8pt, dashed] (-1.54,-0.32) -- (-1.19,-0.32);
\node[anchor=west, inner sep=1pt] at (-1.15,-0.32) {Native solver};
\end{tikzpicture}\cr\noalign{\kern6pt}
\begin{tikzpicture}[font=\figfont, x=\radunit, y=\radunit]
\draw[black!10] (0.000,0.250) -- (0.177,0.177) -- (0.250,0.000) -- (0.177,-0.177) -- (0.000,-0.250) -- (-0.177,-0.177) -- (-0.250,-0.000) -- (-0.177,0.177) -- cycle;
\draw[black!10] (0.000,0.500) -- (0.354,0.354) -- (0.500,0.000) -- (0.354,-0.354) -- (0.000,-0.500) -- (-0.354,-0.354) -- (-0.500,-0.000) -- (-0.354,0.354) -- cycle;
\draw[black!10] (0.000,0.750) -- (0.530,0.530) -- (0.750,0.000) -- (0.530,-0.530) -- (0.000,-0.750) -- (-0.530,-0.530) -- (-0.750,-0.000) -- (-0.530,0.530) -- cycle;
\draw[black!30] (0.000,1.000) -- (0.707,0.707) -- (1.000,0.000) -- (0.707,-0.707) -- (0.000,-1.000) -- (-0.707,-0.707) -- (-1.000,-0.000) -- (-0.707,0.707) -- cycle;
\draw[black!10] (0,0) -- (0.000,1.000);
\node[inner sep=0pt] at (0.000,1.180) {FAN};
\draw[black!10] (0,0) -- (0.707,0.707);
\node[inner sep=0pt] at (0.834,0.834) {Simple};
\draw[black!10] (0,0) -- (1.000,0.000);
\node[inner sep=0pt] at (1.180,0.000) {Explore};
\draw[black!10] (0,0) -- (0.707,-0.707);
\node[inner sep=0pt] at (0.834,-0.834) {Hi};
\draw[black!10] (0,0) -- (0.000,-1.000);
\node[inner sep=0pt] at (0.000,-1.180) {Open};
\draw[black!10] (0,0) -- (-0.707,-0.707);
\node[inner sep=0pt] at (-0.834,-0.834) {TBench};
\draw[black!10] (0,0) -- (-1.000,-0.000);
\node[inner sep=0pt] at (-1.180,-0.000) {Cog};
\draw[black!10] (0,0) -- (-0.707,0.707);
\node[inner sep=0pt] at (-0.834,0.834) {Big};
\draw[cOurs, line width=1pt, fill=cOurs, fill opacity=0.12] (0.000,0.280) -- (0.637,0.637) -- (0.725,0.000) -- (0.490,-0.490) -- (0.000,-0.758) -- (-0.524,-0.524) -- (-0.849,-0.000) -- (-0.639,0.639) -- cycle;
\filldraw[draw=cOurs, fill=cOurs] (0.000,0.280) circle (1.3pt);
\filldraw[draw=cOurs, fill=cOurs] (0.637,0.637) circle (1.3pt);
\filldraw[draw=cOurs, fill=cOurs] (0.725,0.000) circle (1.3pt);
\filldraw[draw=cOurs, fill=cOurs] (0.490,-0.490) circle (1.3pt);
\filldraw[draw=cOurs, fill=cOurs] (0.000,-0.758) circle (1.3pt);
\filldraw[draw=cOurs, fill=cOurs] (-0.524,-0.524) circle (1.3pt);
\filldraw[draw=cOurs, fill=cOurs] (-0.849,-0.000) circle (1.3pt);
\filldraw[draw=cOurs, fill=cOurs] (-0.639,0.639) circle (1.3pt);
\draw[cSCnine, line width=0.9pt] (0.000,0.237) -- (0.619,0.619) -- (0.752,0.000) -- (0.459,-0.459) -- (0.000,-0.750) -- (-0.523,-0.523) -- (-0.859,-0.000) -- (-0.632,0.632) -- cycle;
\filldraw[draw=cSCnine, fill=cSCnine] (0.000,0.302) -- (0.065,0.237) -- (0.000,0.172) -- (-0.065,0.237) -- cycle;
\filldraw[draw=cSCnine, fill=cSCnine] (0.619,0.684) -- (0.684,0.619) -- (0.619,0.554) -- (0.554,0.619) -- cycle;
\filldraw[draw=cSCnine, fill=cSCnine] (0.752,0.065) -- (0.817,0.000) -- (0.752,-0.065) -- (0.687,0.000) -- cycle;
\filldraw[draw=cSCnine, fill=cSCnine] (0.459,-0.394) -- (0.524,-0.459) -- (0.459,-0.524) -- (0.394,-0.459) -- cycle;
\filldraw[draw=cSCnine, fill=cSCnine] (0.000,-0.685) -- (0.065,-0.750) -- (0.000,-0.815) -- (-0.065,-0.750) -- cycle;
\filldraw[draw=cSCnine, fill=cSCnine] (-0.523,-0.458) -- (-0.458,-0.523) -- (-0.523,-0.588) -- (-0.588,-0.523) -- cycle;
\filldraw[draw=cSCnine, fill=cSCnine] (-0.859,0.065) -- (-0.794,-0.000) -- (-0.859,-0.065) -- (-0.924,-0.000) -- cycle;
\filldraw[draw=cSCnine, fill=cSCnine] (-0.632,0.697) -- (-0.567,0.632) -- (-0.632,0.567) -- (-0.697,0.632) -- cycle;
\draw[cNative, line width=0.6pt, dashed] (0.000,0.037) -- (0.360,0.360) -- (0.672,0.000) -- (0.394,-0.394) -- (0.000,-0.665) -- (-0.496,-0.496) -- (-0.797,-0.000) -- (-0.589,0.589) -- cycle;
\node[anchor=north, font=\figfont\bfseries, text=cOurs] at (0,-1.3) {FAN: +4.3 over the best baseline};
\end{tikzpicture}\cr}}}\dimen2=\dimexpr\ht\rowboxB+\dp\rowboxB\relax
\setlength{\radunit}{0.5cm}\sbox{\rowboxB}{\vbox{\offinterlineskip\halign{\hfil#\hfil\cr
\begin{tikzpicture}[font=\figfont, x=1cm, y=1cm]
\draw[cOurs, line width=0.9pt] (-2.46,-0.00) -- (-2.10,-0.00);\filldraw[draw=cOurs, fill=cOurs] (-2.280,-0.000) circle (1.7pt);
\node[anchor=west, inner sep=1pt] at (-2.06,-0.00) {\reader{}};
\draw[cSCnine, line width=0.9pt] (-0.63,-0.00) -- (-0.27,-0.00);\filldraw[draw=cSCnine, fill=cSCnine] (-0.450,0.065) -- (-0.385,-0.000) -- (-0.450,-0.065) -- (-0.515,-0.000) -- cycle;
\node[anchor=west, inner sep=1pt] at (-0.23,-0.00) {best baseline};
\draw[cNative, line width=0.8pt, dashed] (-1.54,-0.32) -- (-1.19,-0.32);
\node[anchor=west, inner sep=1pt] at (-1.15,-0.32) {Native solver};
\end{tikzpicture}\cr\noalign{\kern6pt}
\begin{tikzpicture}[font=\figfont, x=\radunit, y=\radunit]
\draw[black!10] (0.000,0.250) -- (0.177,0.177) -- (0.250,0.000) -- (0.177,-0.177) -- (0.000,-0.250) -- (-0.177,-0.177) -- (-0.250,-0.000) -- (-0.177,0.177) -- cycle;
\draw[black!10] (0.000,0.500) -- (0.354,0.354) -- (0.500,0.000) -- (0.354,-0.354) -- (0.000,-0.500) -- (-0.354,-0.354) -- (-0.500,-0.000) -- (-0.354,0.354) -- cycle;
\draw[black!10] (0.000,0.750) -- (0.530,0.530) -- (0.750,0.000) -- (0.530,-0.530) -- (0.000,-0.750) -- (-0.530,-0.530) -- (-0.750,-0.000) -- (-0.530,0.530) -- cycle;
\draw[black!30] (0.000,1.000) -- (0.707,0.707) -- (1.000,0.000) -- (0.707,-0.707) -- (0.000,-1.000) -- (-0.707,-0.707) -- (-1.000,-0.000) -- (-0.707,0.707) -- cycle;
\draw[black!10] (0,0) -- (0.000,1.000);
\node[inner sep=0pt] at (0.000,1.180) {FAN};
\draw[black!10] (0,0) -- (0.707,0.707);
\node[inner sep=0pt] at (0.834,0.834) {Simple};
\draw[black!10] (0,0) -- (1.000,0.000);
\node[inner sep=0pt] at (1.180,0.000) {Explore};
\draw[black!10] (0,0) -- (0.707,-0.707);
\node[inner sep=0pt] at (0.834,-0.834) {Hi};
\draw[black!10] (0,0) -- (0.000,-1.000);
\node[inner sep=0pt] at (0.000,-1.180) {Open};
\draw[black!10] (0,0) -- (-0.707,-0.707);
\node[inner sep=0pt] at (-0.834,-0.834) {TBench};
\draw[black!10] (0,0) -- (-1.000,-0.000);
\node[inner sep=0pt] at (-1.180,-0.000) {Cog};
\draw[black!10] (0,0) -- (-0.707,0.707);
\node[inner sep=0pt] at (-0.834,0.834) {Big};
\draw[cOurs, line width=1pt, fill=cOurs, fill opacity=0.12] (0.000,0.280) -- (0.637,0.637) -- (0.725,0.000) -- (0.490,-0.490) -- (0.000,-0.758) -- (-0.524,-0.524) -- (-0.849,-0.000) -- (-0.639,0.639) -- cycle;
\filldraw[draw=cOurs, fill=cOurs] (0.000,0.280) circle (1.3pt);
\filldraw[draw=cOurs, fill=cOurs] (0.637,0.637) circle (1.3pt);
\filldraw[draw=cOurs, fill=cOurs] (0.725,0.000) circle (1.3pt);
\filldraw[draw=cOurs, fill=cOurs] (0.490,-0.490) circle (1.3pt);
\filldraw[draw=cOurs, fill=cOurs] (0.000,-0.758) circle (1.3pt);
\filldraw[draw=cOurs, fill=cOurs] (-0.524,-0.524) circle (1.3pt);
\filldraw[draw=cOurs, fill=cOurs] (-0.849,-0.000) circle (1.3pt);
\filldraw[draw=cOurs, fill=cOurs] (-0.639,0.639) circle (1.3pt);
\draw[cSCnine, line width=0.9pt] (0.000,0.237) -- (0.619,0.619) -- (0.752,0.000) -- (0.459,-0.459) -- (0.000,-0.750) -- (-0.523,-0.523) -- (-0.859,-0.000) -- (-0.632,0.632) -- cycle;
\filldraw[draw=cSCnine, fill=cSCnine] (0.000,0.302) -- (0.065,0.237) -- (0.000,0.172) -- (-0.065,0.237) -- cycle;
\filldraw[draw=cSCnine, fill=cSCnine] (0.619,0.684) -- (0.684,0.619) -- (0.619,0.554) -- (0.554,0.619) -- cycle;
\filldraw[draw=cSCnine, fill=cSCnine] (0.752,0.065) -- (0.817,0.000) -- (0.752,-0.065) -- (0.687,0.000) -- cycle;
\filldraw[draw=cSCnine, fill=cSCnine] (0.459,-0.394) -- (0.524,-0.459) -- (0.459,-0.524) -- (0.394,-0.459) -- cycle;
\filldraw[draw=cSCnine, fill=cSCnine] (0.000,-0.685) -- (0.065,-0.750) -- (0.000,-0.815) -- (-0.065,-0.750) -- cycle;
\filldraw[draw=cSCnine, fill=cSCnine] (-0.523,-0.458) -- (-0.458,-0.523) -- (-0.523,-0.588) -- (-0.588,-0.523) -- cycle;
\filldraw[draw=cSCnine, fill=cSCnine] (-0.859,0.065) -- (-0.794,-0.000) -- (-0.859,-0.065) -- (-0.924,-0.000) -- cycle;
\filldraw[draw=cSCnine, fill=cSCnine] (-0.632,0.697) -- (-0.567,0.632) -- (-0.632,0.567) -- (-0.697,0.632) -- cycle;
\draw[cNative, line width=0.6pt, dashed] (0.000,0.037) -- (0.360,0.360) -- (0.672,0.000) -- (0.394,-0.394) -- (0.000,-0.665) -- (-0.496,-0.496) -- (-0.797,-0.000) -- (-0.589,0.589) -- cycle;
\node[anchor=north, font=\figfont\bfseries, text=cOurs] at (0,-1.3) {FAN: +4.3 over the best baseline};
\end{tikzpicture}\cr}}}\dimen4=\dimexpr\ht\rowboxB+\dp\rowboxB\relax
\pgfmathsetlength{\radunit}{14.22638pt+14.22638pt*(\the\rowh-\the\dimen4)/(\the\dimen2-\the\dimen4)}%
\sbox{\rowboxB}{\vbox{\offinterlineskip\halign{\hfil#\hfil\cr
\begin{tikzpicture}[font=\figfont, x=1cm, y=1cm]
\draw[cOurs, line width=0.9pt] (-2.46,-0.00) -- (-2.10,-0.00);\filldraw[draw=cOurs, fill=cOurs] (-2.280,-0.000) circle (1.7pt);
\node[anchor=west, inner sep=1pt] at (-2.06,-0.00) {\reader{}};
\draw[cSCnine, line width=0.9pt] (-0.63,-0.00) -- (-0.27,-0.00);\filldraw[draw=cSCnine, fill=cSCnine] (-0.450,0.065) -- (-0.385,-0.000) -- (-0.450,-0.065) -- (-0.515,-0.000) -- cycle;
\node[anchor=west, inner sep=1pt] at (-0.23,-0.00) {best baseline};
\draw[cNative, line width=0.8pt, dashed] (-1.54,-0.32) -- (-1.19,-0.32);
\node[anchor=west, inner sep=1pt] at (-1.15,-0.32) {Native solver};
\end{tikzpicture}\cr\noalign{\kern6pt}
\begin{tikzpicture}[font=\figfont, x=\radunit, y=\radunit]
\draw[black!10] (0.000,0.250) -- (0.177,0.177) -- (0.250,0.000) -- (0.177,-0.177) -- (0.000,-0.250) -- (-0.177,-0.177) -- (-0.250,-0.000) -- (-0.177,0.177) -- cycle;
\draw[black!10] (0.000,0.500) -- (0.354,0.354) -- (0.500,0.000) -- (0.354,-0.354) -- (0.000,-0.500) -- (-0.354,-0.354) -- (-0.500,-0.000) -- (-0.354,0.354) -- cycle;
\draw[black!10] (0.000,0.750) -- (0.530,0.530) -- (0.750,0.000) -- (0.530,-0.530) -- (0.000,-0.750) -- (-0.530,-0.530) -- (-0.750,-0.000) -- (-0.530,0.530) -- cycle;
\draw[black!30] (0.000,1.000) -- (0.707,0.707) -- (1.000,0.000) -- (0.707,-0.707) -- (0.000,-1.000) -- (-0.707,-0.707) -- (-1.000,-0.000) -- (-0.707,0.707) -- cycle;
\draw[black!10] (0,0) -- (0.000,1.000);
\node[inner sep=0pt] at (0.000,1.180) {FAN};
\draw[black!10] (0,0) -- (0.707,0.707);
\node[inner sep=0pt] at (0.834,0.834) {Simple};
\draw[black!10] (0,0) -- (1.000,0.000);
\node[inner sep=0pt] at (1.180,0.000) {Explore};
\draw[black!10] (0,0) -- (0.707,-0.707);
\node[inner sep=0pt] at (0.834,-0.834) {Hi};
\draw[black!10] (0,0) -- (0.000,-1.000);
\node[inner sep=0pt] at (0.000,-1.180) {Open};
\draw[black!10] (0,0) -- (-0.707,-0.707);
\node[inner sep=0pt] at (-0.834,-0.834) {TBench};
\draw[black!10] (0,0) -- (-1.000,-0.000);
\node[inner sep=0pt] at (-1.180,-0.000) {Cog};
\draw[black!10] (0,0) -- (-0.707,0.707);
\node[inner sep=0pt] at (-0.834,0.834) {Big};
\draw[cOurs, line width=1pt, fill=cOurs, fill opacity=0.12] (0.000,0.280) -- (0.637,0.637) -- (0.725,0.000) -- (0.490,-0.490) -- (0.000,-0.758) -- (-0.524,-0.524) -- (-0.849,-0.000) -- (-0.639,0.639) -- cycle;
\filldraw[draw=cOurs, fill=cOurs] (0.000,0.280) circle (1.3pt);
\filldraw[draw=cOurs, fill=cOurs] (0.637,0.637) circle (1.3pt);
\filldraw[draw=cOurs, fill=cOurs] (0.725,0.000) circle (1.3pt);
\filldraw[draw=cOurs, fill=cOurs] (0.490,-0.490) circle (1.3pt);
\filldraw[draw=cOurs, fill=cOurs] (0.000,-0.758) circle (1.3pt);
\filldraw[draw=cOurs, fill=cOurs] (-0.524,-0.524) circle (1.3pt);
\filldraw[draw=cOurs, fill=cOurs] (-0.849,-0.000) circle (1.3pt);
\filldraw[draw=cOurs, fill=cOurs] (-0.639,0.639) circle (1.3pt);
\draw[cSCnine, line width=0.9pt] (0.000,0.237) -- (0.619,0.619) -- (0.752,0.000) -- (0.459,-0.459) -- (0.000,-0.750) -- (-0.523,-0.523) -- (-0.859,-0.000) -- (-0.632,0.632) -- cycle;
\filldraw[draw=cSCnine, fill=cSCnine] (0.000,0.302) -- (0.065,0.237) -- (0.000,0.172) -- (-0.065,0.237) -- cycle;
\filldraw[draw=cSCnine, fill=cSCnine] (0.619,0.684) -- (0.684,0.619) -- (0.619,0.554) -- (0.554,0.619) -- cycle;
\filldraw[draw=cSCnine, fill=cSCnine] (0.752,0.065) -- (0.817,0.000) -- (0.752,-0.065) -- (0.687,0.000) -- cycle;
\filldraw[draw=cSCnine, fill=cSCnine] (0.459,-0.394) -- (0.524,-0.459) -- (0.459,-0.524) -- (0.394,-0.459) -- cycle;
\filldraw[draw=cSCnine, fill=cSCnine] (0.000,-0.685) -- (0.065,-0.750) -- (0.000,-0.815) -- (-0.065,-0.750) -- cycle;
\filldraw[draw=cSCnine, fill=cSCnine] (-0.523,-0.458) -- (-0.458,-0.523) -- (-0.523,-0.588) -- (-0.588,-0.523) -- cycle;
\filldraw[draw=cSCnine, fill=cSCnine] (-0.859,0.065) -- (-0.794,-0.000) -- (-0.859,-0.065) -- (-0.924,-0.000) -- cycle;
\filldraw[draw=cSCnine, fill=cSCnine] (-0.632,0.697) -- (-0.567,0.632) -- (-0.632,0.567) -- (-0.697,0.632) -- cycle;
\draw[cNative, line width=0.6pt, dashed] (0.000,0.037) -- (0.360,0.360) -- (0.672,0.000) -- (0.394,-0.394) -- (0.000,-0.665) -- (-0.496,-0.496) -- (-0.797,-0.000) -- (-0.589,0.589) -- cycle;
\node[anchor=north, font=\figfont\bfseries, text=cOurs] at (0,-1.3) {FAN: +4.3 over the best baseline};
\end{tikzpicture}\cr}}}%
\begin{minipage}[t]{0.578\textwidth}\vspace{0pt}\centering\usebox{\rowboxA}\end{minipage}\hfill
\begin{minipage}[t]{0.402\textwidth}\vspace{0pt}\centering\fitbox{\usebox{\rowboxB}}\end{minipage}\par\smallskip
\begin{minipage}[t]{0.578\textwidth}\vspace{0pt}
\input{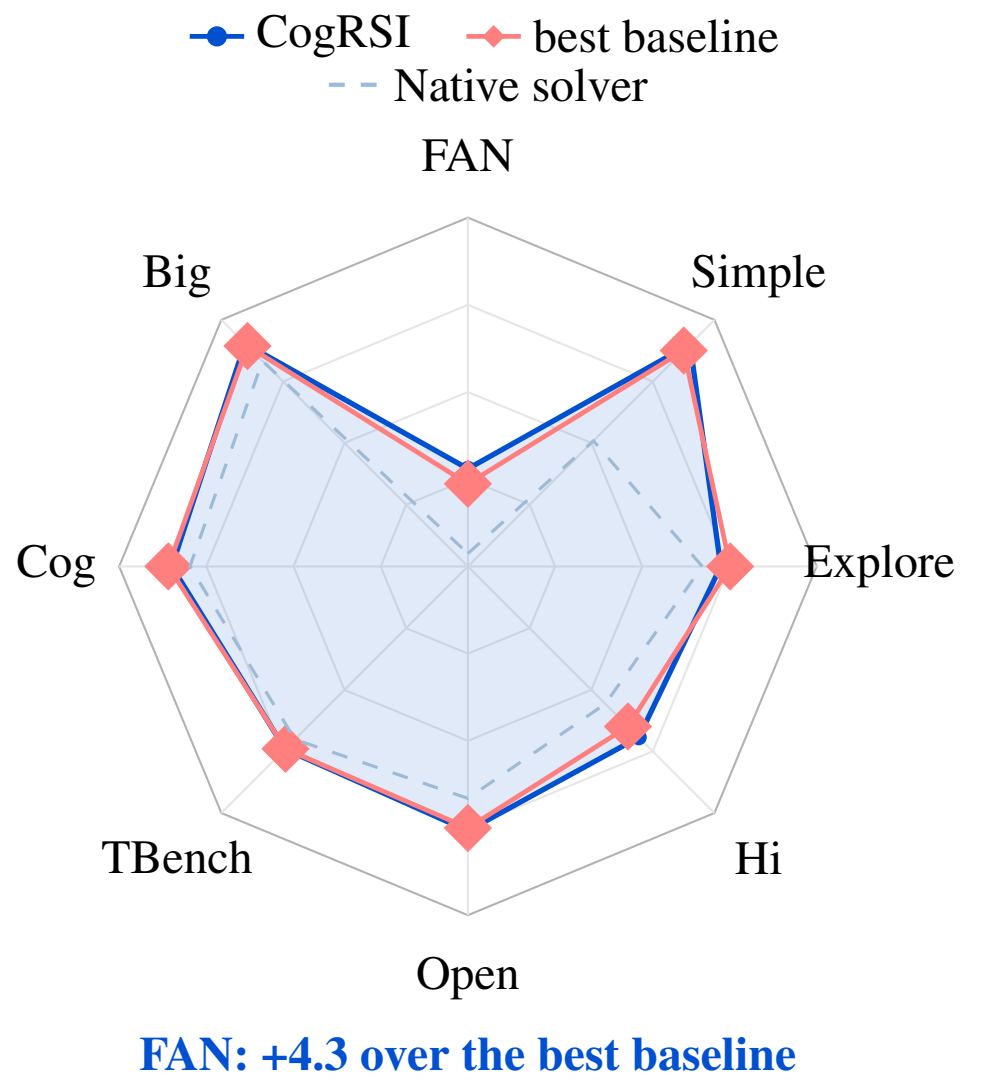}
\end{minipage}\hfill
\begin{minipage}[t]{0.402\textwidth}\vspace{0pt}
\captionof{figure}{\textbf{Table~\ref{tab:t1}(A) by benchmark} (main backbone; rings at 25, 50, 75, 100). \textbf{\reader{} encloses every baseline on FAN, Simple and Hi.}}
\label{fig:radar}
\end{minipage}
\end{figure*}
\subsection{The Reader (RQ1)}
\label{sec:rq1}\label{sec:rq5}
With the expert principle, \reader{} scores \OvVsix{} Main, against \OvSCnine{} for SC ($k{=}9$) at the same token cost
(Table~\ref{tab:t1}). It scores \HOvVsix{} against \HOvSCnine{} on the confirmatory sample and beats SC ($k{=}9$) on all
five backbones. FANToM and SimpleToM carry \ShareMainFS{}\% of the gain. On the seven test-only benchmarks it scores
\NewMeanVsix{}, against \NewMeanSCnine{} for SC ($k{=}9$). Inside the same reader, our annotated view beats PercepToM's
filtered view by \RViewPercep{} narrative Main points (\RViewciPercep{}) and SimToM's view by \RViewSimToM{}
(\RViewciSimToM{}). With learned principles, the reader scores \RMainLoop{}--\RMainVCs{} Main (Table~\ref{tab:arms}).

\paragraph{Each part adds, at equal cost.} The expert principle adds \MacroHandEmpty{} Main points over no principle
(\MacroHandEmptyCI{}). Even without a principle, the reader beats SC ($k{=}9$) by \MacroEmptySCnine{}
(\MacroEmptySCnineCI{}). Each reading and each arbitration step raises belief accuracy (Table~\ref{tab:ablation},
Figure~\ref{fig:radar}). \reader{} makes \CallsVsix{} calls per question and uses \TokRatioVsix{}$\times$ the tokens of SC
($k{=}6$). Against SC ($k{=}9$) at the same token cost, the paired intervals exclude zero on FANToM, SimpleToM and
Hi-ToM (Figure~\ref{fig:cost}, Table~\ref{tab:paired-small}). In a ToMBench case (Figure~\ref{fig:case}), every other
system calls a lie true. The two readings of \reader{} disagree, and arbitration picks ``false'' in all three runs.

\begin{table*}[t]
\centering
\footnotesize
\setlength{\tabcolsep}{1.5pt}
\renewcommand{\arraystretch}{1.02}
\begin{tabular*}{\textwidth}{@{\extracolsep{\fill}}l*{9}{>{\centering\arraybackslash}p{1.16cm}}@{}}
\toprule
\multicolumn{10}{@{}l}{\textbf{(A) Main test sample} (held-out sample~2): eight benchmarks, three backbones} \\
\cmidrule(lr){2-9}\cmidrule(l){10-10}
Method & FAN & Simple & Explore & Hi & Open & TBench & Cog & Big & \textbf{Main} \\
\midrule
\multicolumn{10}{@{}l}{\cellcolor{card}\textit{Qwen3.5-35B-A3B} (main backbone; MoE 35B / 3B active, FP8, thinking off)} \\
Native solver (floor) & 3.7 & 50.9 & 67.2 & 55.8 & 66.5 & 70.1 & 79.7 & 83.3 & 59.7 \\
Zero-shot & 8.0\up{4.3} & 60.8\up{9.9} & 64.9\dn{2.4} & 58.4\up{2.6} & 68.4\up{1.9} & 72.1\up{2.0} & 79.7\eq & 84.9\up{1.7} & 62.1\up{2.5} \\
CoT & 13.2\up{9.4} & 69.1\up{18.2} & 69.6\up{2.3} & 51.8\dn{4.0} & 72.5\up{5.9} & 71.7\up{1.7} & 82.5\up{2.9} & 88.4\up{5.1} & 64.8\up{5.2} \\
SC ($k{=}6$) & 17.6\up{13.9} & 70.2\up{19.3} & 72.6\up{5.4} & 62.2\up{6.4} & 75.0\up{8.5} & 73.9\up{3.8} & \underline{85.5}\up{5.9} & 89.2\up{5.9} & 68.3\up{8.6} \\
SC ($k{=}9$) & 18.0\up{14.3} & 71.2\up{20.3} & 72.1\up{4.8} & 64.9\up{9.2} & 74.4\up{7.9} & 73.3\up{3.2} & \textbf{85.9}\up{6.3} & 89.4\up{6.1} & 68.7\up{9.0} \\
Retrieval & 1.8\dn{2.0} & 49.4\dn{1.5} & 68.8\up{1.6} & 58.6\up{2.8} & 65.9\dn{0.7} & 70.8\up{0.7} & 76.4\dn{3.3} & 83.1\dn{0.1} & 59.3\dn{0.3} \\
GEPA & 0.4\dn{3.4} & 45.7\dn{5.2} & 71.8\up{4.6} & 46.5\dn{9.2} & 65.9\dn{0.6} & 67.5\dn{2.6} & 74.8\dn{4.9} & 81.7\dn{1.6} & 56.8\dn{2.9} \\
PercepToM & 2.0\dn{1.8} & \underline{87.5}\up{36.6} & 67.7\up{0.5} & 48.2\dn{7.6} & 68.4\up{1.9} & 72.0\up{1.9} & 79.7\up{0.1} & 84.8\up{1.5} & 63.8\up{4.1} \\
PercepToM + SC (belief)$^\ddagger$ & 23.7\up{20.0} & --- & --- & --- & --- & --- & --- & --- & --- \\
PercepToM + SC & 3.7\eq & 86.2\up{35.3} & 62.2\dn{5.0} & 55.7\dn{0.1} & 75.6\up{9.1} & \textbf{74.3}\up{4.2} & 84.8\up{5.1} & 89.5\up{6.2} & 66.5\up{6.9} \\
SimToM & 1.1\dn{2.7} & 78.0\up{27.1} & \underline{75.2}\up{8.0} & 37.6\dn{18.2} & 74.1\up{7.5} & 67.9\dn{2.2} & 75.7\dn{4.0} & 85.1\up{1.8} & 61.8\up{2.2} \\
SimToM + SC & 1.8\dn{2.0} & 79.0\up{28.1} & \textbf{76.9}\up{9.6} & 41.6\dn{14.1} & \textbf{76.5}\up{9.9} & 69.6\dn{0.5} & 80.5\up{0.9} & 84.9\up{1.6} & 63.9\up{4.2} \\
\libC{} (rebuilt library) & 2.7\dn{1.1} & 51.8\up{0.9} & 72.2\up{4.9} & 53.9\dn{1.9} & 66.6\up{0.1} & 68.3\dn{1.8} & 72.3\dn{7.3} & 80.9\dn{2.4} & 58.6\dn{1.1} \\
\libA{} (inherited library) & 2.3\dn{1.4} & 54.6\up{3.7} & 73.9\up{6.7} & 53.1\dn{2.7} & 65.9\dn{0.7} & 68.8\dn{1.3} & 75.6\dn{4.1} & 80.4\dn{2.9} & 59.3\dn{0.3} \\
\rowcolor{ours} \readerL{} (loop-written principle) & \underline{25.3}\up{21.6} & 81.7\up{30.8} & 74.0\up{6.8} & \textbf{70.0}\up{14.2} & \textbf{76.5}\up{9.9} & 72.8\up{2.7} & 84.7\up{5.0} & \underline{89.6}\up{6.3} & \underline{71.8}\up{12.2} \\
\rowcolor{ours} \reader{} & \textbf{28.0}\up{24.2} & \textbf{90.0}\up{39.1} & 72.5\up{5.3} & \underline{69.3}\up{13.5} & \underline{75.8}\up{9.3} & \underline{74.1}\up{4.0} & 84.9\up{5.2} & \textbf{90.3}\up{7.1} & \textbf{73.1}\up{13.5} \\
\midrule
\multicolumn{10}{@{}l}{\cellcolor{card}\textit{Qwen3.5-122B-A10B} (MoE 122B / 10B active, FP8)} \\
Zero-shot (floor) & 3.0 & 70.0 & 70.0 & 68.8 & 64.6 & 77.2 & 79.1 & 85.9 & 64.8 \\
CoT & 11.6\up{8.6} & 78.5\up{8.6} & 72.9\up{2.9} & 55.6\dn{13.1} & 73.1\up{8.5} & 74.8\dn{2.4} & 84.9\up{5.7} & 89.0\up{3.1} & 67.5\up{2.7} \\
SC ($k{=}6$) & \underline{18.2}\up{15.2} & 80.8\up{10.9} & \underline{77.5}\up{7.6} & 68.0\dn{0.8} & \underline{76.3}\up{11.7} & \underline{78.0}\up{0.7} & 87.8\up{8.7} & \textbf{89.9}\up{4.0} & \underline{72.1}\up{7.2} \\
SC ($k{=}9$) & 16.9\up{13.9} & 81.9\up{12.0} & \underline{77.5}\up{7.5} & \underline{68.7}\dn{0.1} & 76.0\up{11.4} & \underline{78.0}\up{0.8} & \textbf{88.5}\up{9.3} & 89.1\up{3.2} & \underline{72.1}\up{7.3} \\
PercepToM & 6.2\up{3.2} & \underline{88.2}\up{18.3} & 70.4\up{0.5} & 48.0\dn{20.8} & 63.7\dn{0.9} & 77.8\up{0.5} & 78.9\dn{0.3} & 86.2\up{0.3} & 64.9\up{0.1} \\
\rowcolor{ours} \reader{} & \textbf{25.3}\up{22.3} & \textbf{94.0}\up{24.1} & \textbf{78.4}\up{8.5} & \textbf{72.4}\up{3.7} & \textbf{78.6}\up{14.0} & \textbf{78.2}\up{1.0} & \underline{87.9}\up{8.8} & \underline{89.7}\up{3.8} & \textbf{75.6}\up{10.8} \\
\midrule
\multicolumn{10}{@{}l}{\cellcolor{card}\textit{Llama-3.3-70B-Instruct} (dense 70B, AWQ 4-bit)} \\
Zero-shot (floor) & 0.0 & 72.5 & 70.9 & 65.9 & 69.0 & 75.8 & 81.4 & 87.7 & 65.4 \\
CoT & 1.6\up{1.6} & 86.0\up{13.6} & \underline{77.9}\up{6.9} & 65.5\dn{0.4} & 72.1\up{3.1} & \textbf{76.4}\up{0.6} & \textbf{85.2}\up{3.8} & \underline{88.0}\up{0.3} & 69.1\up{3.7} \\
SC ($k{=}6$) & 3.7\up{3.7} & \underline{87.1}\up{14.7} & 77.0\up{6.1} & 68.9\up{2.9} & \underline{72.9}\up{3.9} & 75.4\dn{0.4} & 83.8\up{2.4} & \underline{88.0}\up{0.3} & 69.6\up{4.2} \\
SC ($k{=}9$) & \underline{4.3}\up{4.3} & 86.8\up{14.4} & 77.2\up{6.2} & \underline{69.0}\up{3.1} & 72.8\up{3.8} & \underline{76.2}\up{0.3} & \underline{83.9}\up{2.5} & 87.9\up{0.2} & \underline{69.8}\up{4.3} \\
PercepToM & 1.1\up{1.1} & 85.0\up{12.6} & \textbf{78.0}\up{7.1} & 59.1\dn{6.8} & 69.0\eq & 75.8\eq & 81.4\eq & 87.1\dn{0.6} & 67.1\up{1.7} \\
\rowcolor{ours} \reader{} & \textbf{12.8}\up{12.8} & \textbf{96.6}\up{24.2} & 75.9\up{5.0} & \textbf{69.7}\up{3.8} & \textbf{75.7}\up{6.7} & 74.9\dn{0.9} & 83.1\up{1.7} & \textbf{89.4}\up{1.7} & \textbf{72.3}\up{6.9} \\
\bottomrule
\end{tabular*}
\par\medskip
\begin{tabular*}{\textwidth}{@{\extracolsep{\fill}}l*{9}{>{\centering\arraybackslash}p{1.16cm}}@{}}
\toprule
\multicolumn{10}{@{}l}{\textbf{(B) Confirmatory sample} (held-out sample~3, drawn after every design decision; main backbone)} \\
\cmidrule(lr){2-9}\cmidrule(l){10-10}
Method & FAN & Simple & Explore & Hi & Open & TBench & Cog & Big & \textbf{Main} \\
\midrule
CoT & 14.8 & 66.4 & 63.9 & 60.0 & 70.2 & 78.6 & 85.5 & 91.7 & 66.4 \\
SC ($k{=}6$) & 17.6 & 66.8 & 66.7 & 64.3 & 72.2 & \textbf{81.4} & 87.3 & \underline{93.3} & 68.7 \\
SC ($k{=}9$) & 16.6 & 66.7 & 67.1 & 65.2 & 71.5 & \underline{81.1} & 87.7 & 91.7 & 68.5 \\
PercepToM & 2.1 & 84.7 & \underline{69.4} & 51.4 & 65.0 & 76.8 & 81.7 & 90.0 & 65.1 \\
PercepToM + SC (belief)$^\ddagger$ & 26.2 & --- & --- & --- & --- & --- & --- & --- & --- \\
SimToM & 2.1 & 72.8 & 68.9 & 31.4 & \underline{74.8} & 71.0 & 79.9 & 88.3 & 61.2 \\
\reader{} (vote only) & 25.7 & 87.1 & 68.3 & \underline{65.7} & 74.6 & 79.8 & 86.7 & 88.3 & 72.0 \\
\reader{} (no close-split) & 27.2 & \underline{88.3} & 66.9 & 62.9 & \textbf{75.2} & 80.6 & \textbf{88.8} & 91.7 & \underline{72.7} \\
\rowcolor{ours} \readerL{} (loop-written principle) & \textbf{29.3} & 80.4 & \textbf{70.6} & \textbf{66.7} & 73.9 & 79.9 & \underline{88.3} & 92.5 & \underline{72.7} \\
\rowcolor{ours} \reader{} & \underline{28.8} & \textbf{88.8} & 67.5 & \underline{65.7} & 74.5 & 80.4 & \underline{88.3} & \textbf{93.9} & \textbf{73.5} \\
\bottomrule
\end{tabular*}
\caption{\textbf{Main results.} (A) Main test sample, three backbones. (B) Confirmatory sample. Accuracy (\%) is the mean of runs: three in A, and in B three for \reader{} and SC ($k{=}9$), six loop seeds for \readerL{}, one for the rest. Main averages the eight. Subscripts are gaps to the block floor. Bold is best, underline second. FAN is FANToM All$^*$. \textbf{\reader{} has the best Main in every block.} $^\ddagger$FANToM only. ---: does not apply.}
\label{tab:t1}
\end{table*}

\begin{figure*}[t]
\sbox{\rowboxA}{\begin{minipage}{0.66\textwidth}\centering\centering \begin{tikzpicture}[font=\figfont, x=1cm, y=1cm]
\filldraw[draw=cOurs, fill=cOurs] (-4.020,-0.000) circle (2.0pt);
\node[anchor=west, inner sep=1pt] at (-3.80,-0.00) {\reader{}};
\filldraw[draw=cOursL, fill=cOursL] (-2.190,-0.000) circle (2.0pt);
\node[anchor=west, inner sep=1pt] at (-1.97,-0.00) {\readerL{}};
\filldraw[draw=cSCnine, fill=cSCnine] (0.000,0.065) -- (0.065,-0.000) -- (0.000,-0.065) -- (-0.065,-0.000) -- cycle;
\node[anchor=west, inner sep=1pt] at (0.22,-0.00) {SC ($k{=}9$)};
\filldraw[draw=cSCsix, fill=cSCsix] (2.190,0.065) -- (2.255,-0.000) -- (2.190,-0.065) -- (2.125,-0.000) -- cycle;
\node[anchor=west, inner sep=1pt] at (2.41,-0.00) {SC ($k{=}6$)};
\filldraw[draw=cPercep, fill=cPercep] (-4.755,-0.255) -- (-4.815,-0.365) -- (-4.695,-0.365) -- cycle;
\node[anchor=west, inner sep=1pt] at (-4.53,-0.32) {PercepToM};
\filldraw[draw=cCoT, fill=cCoT] (-2.385,-0.255) -- (-2.320,-0.320) -- (-2.385,-0.385) -- (-2.450,-0.320) -- cycle;
\node[anchor=west, inner sep=1pt] at (-2.17,-0.32) {CoT};
\filldraw[draw=cZS, fill=cZS] (-1.095,-0.255) -- (-1.030,-0.320) -- (-1.095,-0.385) -- (-1.160,-0.320) -- cycle;
\node[anchor=west, inner sep=1pt] at (-0.87,-0.32) {Zero-shot};
\filldraw[draw=cRetr, fill=cRetr] (1.275,-0.255) -- (1.340,-0.320) -- (1.275,-0.385) -- (1.210,-0.320) -- cycle;
\node[anchor=west, inner sep=1pt] at (1.49,-0.32) {Retrieval};
\filldraw[draw=cGEPA, fill=cGEPA] (3.645,-0.255) -- (3.710,-0.320) -- (3.645,-0.385) -- (3.580,-0.320) -- cycle;
\node[anchor=west, inner sep=1pt] at (3.86,-0.32) {GEPA};
\filldraw[draw=cZS, fill=cZS] (-1.905,-0.575) -- (-1.840,-0.640) -- (-1.905,-0.705) -- (-1.970,-0.640) -- cycle;
\node[anchor=west, inner sep=1pt] at (-1.69,-0.64) {best other baseline};
\end{tikzpicture}\\[3pt]
\begin{minipage}[b]{0.490\linewidth}\centering\fitbox{\begin{tikzpicture}[font=\figfont, x=1cm, y=1cm]
\draw[black!10] (0,0.000) -- (3.200,0.000);
\draw[black!60] (0,0.000) -- (-0.05,0.000) node[left, inner sep=1pt] {55};
\draw[black!10] (0,0.550) -- (3.200,0.550);
\draw[black!60] (0,0.550) -- (-0.05,0.550) node[left, inner sep=1pt] {60};
\draw[black!10] (0,1.100) -- (3.200,1.100);
\draw[black!60] (0,1.100) -- (-0.05,1.100) node[left, inner sep=1pt] {65};
\draw[black!10] (0,1.650) -- (3.200,1.650);
\draw[black!60] (0,1.650) -- (-0.05,1.650) node[left, inner sep=1pt] {70};
\draw[black!10] (0,2.200) -- (3.200,2.200);
\draw[black!60] (0,2.200) -- (-0.05,2.200) node[left, inner sep=1pt] {75};
\node[rotate=90, anchor=south] at (-0.6,1.100) {Main accuracy};
\node[anchor=south west, font=\figfont\bfseries, inner sep=0pt] at (-0.85,2.320) {(a) Cost vs.\ accuracy};
\draw[black!60] (0.619,0) -- (0.619,-0.05) node[below, inner sep=1pt] {0.1};
\draw[black!60] (1.600,0) -- (1.600,-0.05) node[below, inner sep=1pt] {0.3};
\draw[black!60] (2.675,0) -- (2.675,-0.05) node[below, inner sep=1pt] {1};
\node[anchor=north] at (1.600,-0.33) {tokens per question ($\times$ \reader{})};
\filldraw[draw=cZS, fill=cZS] (0.148,0.851) -- (0.213,0.786) -- (0.148,0.721) -- (0.083,0.786) -- cycle;
\filldraw[draw=cCoT, fill=cCoT] (0.689,1.147) -- (0.754,1.082) -- (0.689,1.017) -- (0.624,1.082) -- cycle;
\filldraw[draw=cSCsix, fill=cSCsix] (2.309,1.526) -- (2.374,1.461) -- (2.309,1.396) -- (2.244,1.461) -- cycle;
\filldraw[draw=cSCnine, fill=cSCnine] (2.671,1.567) -- (2.736,1.502) -- (2.671,1.437) -- (2.606,1.502) -- cycle;
\filldraw[draw=cRetr, fill=cRetr] (1.210,0.543) -- (1.275,0.478) -- (1.210,0.413) -- (1.145,0.478) -- cycle;
\filldraw[draw=cGEPA, fill=cGEPA] (1.944,0.263) -- (2.009,0.198) -- (1.944,0.133) -- (1.879,0.198) -- cycle;
\filldraw[draw=cPercep, fill=cPercep] (0.584,1.033) -- (0.524,0.923) -- (0.644,0.923) -- cycle;
\filldraw[draw=cOursL, fill=cOursL] (2.707,1.851) circle (2.0pt);
\filldraw[draw=cOurs, fill=cOurs] (2.675,1.994) circle (2.0pt);
\node[anchor=south west, inner sep=3.6pt] at (0.689,1.082) {CoT};
\node[anchor=north, inner sep=3.6pt] at (2.671,1.342) {SC ($k{=}9$)};
\node[anchor=west, inner sep=3.6pt] at (1.944,0.198) {GEPA};
\node[anchor=west, inner sep=3.6pt] at (2.675,1.994) {\reader{}};
\draw[cOurs, -{Stealth[length=3pt]}, line width=0.6pt] (2.671,1.582) -- (2.675,1.914);
\node[anchor=north west, font=\figfont\bfseries, text=cOurs, inner sep=1pt, align=center] at (0.120,2.200) {+4.4 at\\ equal tokens};
\draw[black!85, line width=0.8pt] (0,0) -- (0,2.200); \draw[black!85, line width=0.8pt] (0,0) -- (3.200,0);
\path (0,-0.75) (0,2.70);
\end{tikzpicture}}\end{minipage}\hfill
\begin{minipage}[b]{0.490\linewidth}\centering\fitbox{\begin{tikzpicture}[font=\figfont, x=1cm, y=1cm]
\draw[black!10] (0,0.000) -- (3.200,0.000);
\draw[black!60] (0,0.000) -- (-0.05,0.000) node[left, inner sep=1pt] {30};
\draw[black!10] (0,0.978) -- (3.200,0.978);
\draw[black!60] (0,0.978) -- (-0.05,0.978) node[left, inner sep=1pt] {50};
\draw[black!10] (0,1.956) -- (3.200,1.956);
\draw[black!60] (0,1.956) -- (-0.05,1.956) node[left, inner sep=1pt] {70};
\node[rotate=90, anchor=south] at (-0.6,1.100) {accuracy};
\node[anchor=south west, font=\figfont\bfseries, inner sep=0pt] at (-0.85,2.320) {(b) Every backbone};
\draw[black!60] (0.209,0) -- (0.209,-0.05) node[below, inner sep=1pt] {9B};
\draw[black!60] (0.904,0) -- (0.904,-0.05) node[below, inner sep=1pt] {GLM};
\draw[black!60] (1.600,0) -- (1.600,-0.05) node[below, inner sep=1pt] {35B};
\draw[black!60] (2.296,0) -- (2.296,-0.05) node[below, inner sep=1pt] {70B};
\draw[black!60] (2.991,0) -- (2.991,-0.05) node[below, inner sep=1pt] {122B};
\node[anchor=north] at (1.600,-0.33) {backbone (small $\to$ large)};
\fill[cOurs, opacity=0.12] (0.209,1.509) -- (0.904,1.216) -- (1.600,1.927) -- (2.296,1.816) -- (2.991,1.974) -- (2.991,1.671) -- (2.296,1.564) -- (1.600,1.509) -- (0.904,0.876) -- (0.209,1.091) -- cycle;
\draw[cZS, line width=0.7pt] (0.209,0.929) -- (0.904,0.831) -- (1.600,1.253) -- (2.296,1.481) -- (2.991,1.403);
\filldraw[draw=cZS, fill=cZS] (0.209,0.994) -- (0.274,0.929) -- (0.209,0.864) -- (0.144,0.929) -- cycle;
\filldraw[draw=cZS, fill=cZS] (0.904,0.896) -- (0.969,0.831) -- (0.904,0.766) -- (0.839,0.831) -- cycle;
\filldraw[draw=cZS, fill=cZS] (1.600,1.318) -- (1.665,1.253) -- (1.600,1.188) -- (1.535,1.253) -- cycle;
\filldraw[draw=cZS, fill=cZS] (2.296,1.546) -- (2.361,1.481) -- (2.296,1.416) -- (2.231,1.481) -- cycle;
\filldraw[draw=cZS, fill=cZS] (2.991,1.468) -- (3.056,1.403) -- (2.991,1.338) -- (2.926,1.403) -- cycle;
\draw[cSCsix, line width=0.7pt] (0.209,1.045) -- (0.904,0.878) -- (1.600,1.457) -- (2.296,1.561) -- (2.991,1.674);
\filldraw[draw=cSCsix, fill=cSCsix] (0.209,1.110) -- (0.274,1.045) -- (0.209,0.980) -- (0.144,1.045) -- cycle;
\filldraw[draw=cSCsix, fill=cSCsix] (0.904,0.943) -- (0.969,0.878) -- (0.904,0.813) -- (0.839,0.878) -- cycle;
\filldraw[draw=cSCsix, fill=cSCsix] (1.600,1.522) -- (1.665,1.457) -- (1.600,1.392) -- (1.535,1.457) -- cycle;
\filldraw[draw=cSCsix, fill=cSCsix] (2.296,1.626) -- (2.361,1.561) -- (2.296,1.496) -- (2.231,1.561) -- cycle;
\filldraw[draw=cSCsix, fill=cSCsix] (2.991,1.739) -- (3.056,1.674) -- (2.991,1.609) -- (2.926,1.674) -- cycle;
\draw[cSCnine, line width=0.9pt] (0.209,1.091) -- (0.904,0.876) -- (1.600,1.509) -- (2.296,1.564) -- (2.991,1.671);
\filldraw[draw=cSCnine, fill=cSCnine] (0.209,1.156) -- (0.274,1.091) -- (0.209,1.026) -- (0.144,1.091) -- cycle;
\filldraw[draw=cSCnine, fill=cSCnine] (0.904,0.941) -- (0.969,0.876) -- (0.904,0.811) -- (0.839,0.876) -- cycle;
\filldraw[draw=cSCnine, fill=cSCnine] (1.600,1.574) -- (1.665,1.509) -- (1.600,1.444) -- (1.535,1.509) -- cycle;
\filldraw[draw=cSCnine, fill=cSCnine] (2.296,1.629) -- (2.361,1.564) -- (2.296,1.499) -- (2.231,1.564) -- cycle;
\filldraw[draw=cSCnine, fill=cSCnine] (2.991,1.736) -- (3.056,1.671) -- (2.991,1.606) -- (2.926,1.671) -- cycle;
\draw[cOurs, line width=1.3pt] (0.209,1.509) -- (0.904,1.216) -- (1.600,1.927) -- (2.296,1.816) -- (2.991,1.974);
\filldraw[draw=cOurs, fill=cOurs] (0.209,1.509) circle (1.9pt);
\filldraw[draw=cOurs, fill=cOurs] (0.904,1.216) circle (1.9pt);
\filldraw[draw=cOurs, fill=cOurs] (1.600,1.927) circle (1.9pt);
\filldraw[draw=cOurs, fill=cOurs] (2.296,1.816) circle (1.9pt);
\filldraw[draw=cOurs, fill=cOurs] (2.991,1.974) circle (1.9pt);
\draw[black!85, line width=0.8pt] (0,0) -- (0,2.200); \draw[black!85, line width=0.8pt] (0,0) -- (3.200,0);
\path (0,-0.75) (0,2.70);
\node[anchor=south east, font=\figfont\bfseries, text=cOurs, align=right] at (3.2,0.04) {+5.1 to +8.6\\ over SC ($k{=}9$)};
\end{tikzpicture}}\end{minipage}\end{minipage}}%
\setlength{\rowh}{\dimexpr\ht\rowboxA+\dp\rowboxA\relax}%
\sbox{\rowboxB}{\begin{tabular}{@{}lcc@{}}
\toprule
Bench. & $\Delta$ & 95\% CI \\
\midrule
FAN & \textbf{+10.0} & \textbf{[+4.0, +15.7]} \\
Simple & \textbf{+18.9} & \textbf{[+15.0, +22.7]} \\
Explore & +0.5 & [$-$1.5, +2.3] \\
Hi & \textbf{+4.4} & \textbf{[+1.2, +7.6]} \\
Open & +1.4 & [$-$0.4, +3.1] \\
TBench & +0.8 & [$-$0.7, +2.2] \\
Cog & $-$1.1 & [$-$2.3, +0.1] \\
Big & +1.0 & [$-$0.1, +2.2] \\
\bottomrule
\end{tabular}
}\dimen2=\dimexpr\ht\rowboxB+\dp\rowboxB\relax
\sbox{\rowboxB}{\renewcommand{\arraystretch}{2}\begin{tabular}{@{}lcc@{}}
\toprule
Bench. & $\Delta$ & 95\% CI \\
\midrule
FAN & \textbf{+10.0} & \textbf{[+4.0, +15.7]} \\
Simple & \textbf{+18.9} & \textbf{[+15.0, +22.7]} \\
Explore & +0.5 & [$-$1.5, +2.3] \\
Hi & \textbf{+4.4} & \textbf{[+1.2, +7.6]} \\
Open & +1.4 & [$-$0.4, +3.1] \\
TBench & +0.8 & [$-$0.7, +2.2] \\
Cog & $-$1.1 & [$-$2.3, +0.1] \\
Big & +1.0 & [$-$0.1, +2.2] \\
\bottomrule
\end{tabular}
}\dimen4=\dimexpr\ht\rowboxB+\dp\rowboxB\relax
\pgfmathsetmacro{\pairstretch}{1+(\the\rowh-\the\dimen2)/(\the\dimen4-\the\dimen2)}%
\sbox{\rowboxB}{\renewcommand{\arraystretch}{\pairstretch}\begin{tabular}{@{}lcc@{}}
\toprule
Bench. & $\Delta$ & 95\% CI \\
\midrule
FAN & \textbf{+10.0} & \textbf{[+4.0, +15.7]} \\
Simple & \textbf{+18.9} & \textbf{[+15.0, +22.7]} \\
Explore & +0.5 & [$-$1.5, +2.3] \\
Hi & \textbf{+4.4} & \textbf{[+1.2, +7.6]} \\
Open & +1.4 & [$-$0.4, +3.1] \\
TBench & +0.8 & [$-$0.7, +2.2] \\
Cog & $-$1.1 & [$-$2.3, +0.1] \\
Big & +1.0 & [$-$0.1, +2.2] \\
\bottomrule
\end{tabular}
}%
\begin{minipage}[t]{0.66\textwidth}\vspace{0pt}\centering\usebox{\rowboxA}\end{minipage}\hfill
\begin{minipage}[t]{0.31\textwidth}\vspace{0pt}\centering\fitbox{\usebox{\rowboxB}}\end{minipage}\par\smallskip
\begin{minipage}[t]{0.66\textwidth}\vspace{0pt}
\captionof{figure}{\textbf{Cost and robustness.} (a) Main accuracy vs.\ tokens (log, relative to \reader{}). (b) Four-benchmark mean by backbone, with SC ($k{=}6$) and ($k{=}9$) (70B is Llama-3.3). \textbf{\reader{} is on the frontier and leads SC at equal tokens on every backbone.}}
\label{fig:cost}
\end{minipage}\hfill
\begin{minipage}[t]{0.31\textwidth}\vspace{0pt}
\captionof{table}{\textbf{Paired test}: \reader{} $-$ SC ($k{=}9$) at equal tokens. \textbf{Significant on FAN, Simple and Hi; Hi not after Holm correction.}}
\label{tab:paired-small}
\end{minipage}
\end{figure*}
\begin{figure}[t]
\centering
\begin{tikzpicture}
\node[draw=boxrule, line width=0.6pt, rounded corners=3pt, inner sep=4pt] {\begin{minipage}{\dimexpr\linewidth-8pt-1.2pt\relax}
\centering\begin{tikzpicture}[font=\figfont, x=1cm, y=1cm,
 chip/.style={draw=black!25, rounded corners=1.5pt, minimum width=0.42cm, minimum height=0.42cm, inner sep=0pt},
 lab/.style={anchor=west, inner sep=0pt}, arr/.style={-{Stealth[length=4pt]}, black!55, line width=0.6pt}]
\node[anchor=north west, text width=6.9cm, fill=card, draw=black!12, rounded corners=2pt, inner sep=4pt, align=left] (story) at (0,0)
 {{\bfseries Story} (abridged). Asked about her company's new strategy, Miss Zhou says her team has developed a breakthrough method but cannot say more. \textcolor{cOurs}{After the gathering, she has her team start planning the strategy this week.}\\[2pt]
 Is what Miss Zhou says true? Gold: B (No).};
\path (story.south west) ++(0,-0.4) coordinate (r);
\path (r) ++(0,-0.00) node[lab] {Annotated};
\path (r) ++(1.85,-0.00) node[chip, fill=cAnn] {A};
\path (r) ++(2.35,-0.00) node[chip, fill=cAnn] {B};
\path (r) ++(2.85,-0.00) node[chip, fill=cAnn] {B};
\path (r) ++(3.35,-0.00) node[chip, fill=cAnn] {B};
\path (r) ++(3.95,-0.00) coordinate (a0); \path (r) ++(4.4,-0.00) coordinate (b0); \draw[arr] (a0) -- (b0);
\path (r) ++(4.5,-0.00) node[lab] {B (3 of 4)};
\path (r) ++(0,-0.55) node[lab] {Plain};
\path (r) ++(1.85,-0.55) node[chip, fill=cPlain] {A};
\path (r) ++(2.35,-0.55) node[chip, fill=cPlain] {A};
\path (r) ++(3.95,-0.55) coordinate (a1); \path (r) ++(4.4,-0.55) coordinate (b1); \draw[arr] (a1) -- (b1);
\path (r) ++(4.5,-0.55) node[lab] {A (2 of 2)};
\path (r) ++(0,-1.2) node[lab] {Vote of all six (3--3)};
\path (r) ++(3.95,-1.2) coordinate (a2); \path (r) ++(4.4,-1.2) coordinate (b2); \draw[arr] (a2) -- (b2);
\path (r) ++(4.5,-1.2) node[lab] {A \textcolor{cNeg}{$\times$}};
\path (r) ++(0,-1.75) node[lab] {Arbiter + principle};
\path (r) ++(3.95,-1.75) coordinate (a3); \path (r) ++(4.4,-1.75) coordinate (b3); \draw[arr, cOurs] (a3) -- (b3);
\path (r) ++(4.5,-1.75) node[lab, fill=cOurs!12, inner sep=2pt] {\textcolor{cOurs}{\reader{}: B} \textcolor{cPos}{$\surd$}};
\path (r) ++(0,-2.15) coordinate (s1); \path (r) ++(6.98,-2.15) coordinate (s2); \draw[black!15] (s1) -- (s2);
\path (r) ++(0,-2.5) node[lab] {Every run of the other systems};
\path (r) ++(0.00,-2.95) node[lab] {SC ($k{=}6$): A \textcolor{cNeg}{$\times$}};
\path (r) ++(3.70,-2.95) node[lab] {PercepToM: A \textcolor{cNeg}{$\times$}};
\path (r) ++(0.00,-3.40) node[lab] {\libA{}: A \textcolor{cNeg}{$\times$}};
\path (r) ++(3.70,-3.40) node[lab] {\readerL{}: A \textcolor{cNeg}{$\times$}};
\end{tikzpicture}
\end{minipage}};
\end{tikzpicture}
\caption{\textbf{Case study.} A ToMBench egocentric lie, abridged (full text and logs in
Table~\ref{tab:case}). Chips are \reader{}'s six samples in run~0. \textbf{The readings disagree, and arbitration
picks the right answer.}}
\label{fig:case}
\end{figure}
\subsection{Evolved Strategies Transfer (RQ2)}
\label{sec:rq2}
\paragraph{Beyond the expert on independent pairs.} On BigToM-U, every learned principle scores above the expert one
(Table~\ref{tab:arms}). The gated library reaches \QBUVC{}\% pair validity, against \QBUHand{} for the expert
principle (\QBUdVCHand{}, \QBUciVCHand{}) and \QBUNone{} with no principle (\QBUdVCNone{}, \QBUciVCNone{}). The
accuracy-only library gains \QBUdAONone{} over no principle. The gain comes from flip pairs, where the agent does or
does not see the change (\QBUFlipVC{} vs.\ \QBUFlipHand{}). On same pairs every principle stays near ceiling
(\QBUSameVC{} vs.\ \QBUSameNone{}), so learned rules do not make the reader see missed events where there are none. SC
($k{=}9$) reaches \QBUSCnine{}, below the reader without a principle.

\paragraph{The gate adds nothing here.} The pre-registered primary contrast, gated minus accuracy-only library, is
\QBUdVCAO{} (\QBUciVCAO{}). Gated minus pairs-as-data is \QBUdVCB{} (\QBUciVCB{}). The transfer comes from evolving the
principle.

\paragraph{Development gains do not reach test Main.} The accuracy-only loop's principle beats the expert one on new
development questions in most seeds (\RDevLoop{} vs.\ \RDevHand{}, Figure~\ref{fig:rsip}), yet it loses on test Main.
A loop that writes executable tools shows the same pattern. Its library lifts ExploreToM from \RExploreNative{} to
\RExploreRSIA{} but leaves FANToM flat (Figure~\ref{fig:rsi}), and inheriting it instead of rebuilding it changes
FANToM All$^*$ by \RInhMC{} (\RInhciMC{}).

\paragraph{Limits.} On test Main the gated library trails the expert principle by \RdVCHand{} (\RciVCHand{}) and is
level with the accuracy-only loop (\RdVCLoop{}). GEPA writes prompts of \RGEPAWordsLo{}--\RGEPAWordsHi{} words. They
reach \RExploreSlot{} on ExploreToM, \RMainGEPA{} Main and \QBUGEPA{} on BigToM-U. When a belief must be inferred
backwards from an action, every system stays near \QBWVC{}\% pair validity (\QBWFlipVC{}\% on flip pairs), and no
principle helps (\QBWdVCNone{}, \QBWciVCNone{}).

\subsection{What Registration Buys (RQ3)}
\label{sec:rq3}\label{sec:rq4b}
\begin{table}[t]
\centering
\footnotesize
\setlength{\tabcolsep}{0pt}
\begin{tabular*}{\columnwidth}{@{\extracolsep{\fill}}lccccc@{}}
\toprule
Registration & Adm. & Prec. & Harm. & Recall & Missed \\
\midrule
1-pt margin & 46 & 63.6 & 31.8 & 23.7 & 22.9 \\
3-pt margin & 25 & 83.3 & 16.7 & 21.7 & 29.2 \\
3-pt + gate (single) & 25 & 83.3 & 16.7 & 17.2 & 25.0 \\
\rowcolor{ours} 3-pt + gate (library) & 25 & 83.3 & 16.7 & 13.0 & 43.8 \\
\bottomrule
\end{tabular*}
\caption{\textbf{Selection quality} (\%, development data). Every candidate scored on 120 held-out questions. Adm.: rounds with an admission; Prec./Harm.: admitted updates that help/hurt; Recall: helpful candidates admitted; Missed: rounds with a helpful candidate and no admission. \textbf{The calibrated margin, not the gate, makes selection precise.}}
\label{tab:selection}
\end{table}

\paragraph{A calibrated margin makes admissions precise.} We scored every candidate on 120 held-out development
questions. Of the updates the 1-point loop kept, \RPrecLoop{}\% help. With the 3-point margin, \RPrecAO{}\% help, with or
without the gate (\RNAdmAO{} and \RNAdmVC{} kept updates, Table~\ref{tab:selection}). The margin halves the share of
harmful updates, from \RHarmLoop{} to \RHarmAO{}\%, and keeps fewer helpful ones.

\paragraph{Counterfactual items help, with or without the gate.} On the evaluation split of our own generator, the
gated library keeps \RCFVC{}\% of pairs right, against \RCFAO{} for the accuracy-only library (\RdCFVCAO{},
\RciCFVCAO{}). Almost all of this gain is on one case: a character misses a move and later looks (Figure~\ref{fig:vc}).
On the development gate pairs, the pairs-as-data library keeps \QGateB{} of 96 right and the gated library \QGateC{}.
The accuracy-only library keeps \QGateA{}, with seeds ranging from \QGateALo{} to \QGateAHi{}. The counterfactual items
carry the gain, and the pair constraint adds little. We cannot show that the gated library is less accurate on
development (\QDevCB{}, \QDevCBci{} against pairs-as-data).

\subsection{What Evolved Strategies Say (RQ4)}
\label{sec:rq4}
\paragraph{Most evolved rules fail somewhere.} The principle loops proposed \QAudTexts{} rules. \QAudInv{} are invalid:
for each, there is a story it claims to cover where following it gives the wrong answer. \QAudVal{} are valid,
\QAudAmb{} are ambiguous because the two translations disagree, and \QAudVac{} almost never apply. Among the rules the
audit could decide, \QAudPropShare{}\% (\QAudPropCI{}) are invalid, in every loop (Table~\ref{tab:arms}). In the
pre-registered run, one translator hit its output limit on \QAudTrunc{} rules. We recompiled them with a larger limit
and report both sets of verdicts (before: \QAudPreInv{} invalid and \QAudPreAmb{} ambiguous, Appendix~\ref{app:vc}).

\paragraph{Accuracy cannot filter them.} We replayed each registration rule on the same candidates. Among the rules a
1-point margin would keep, \QRepOne{}\% are invalid. The 3-point margin gives \QRepThree{}\% and the gate
\QRepGate{}\% (Figure~\ref{fig:audit}a). Valid and invalid rules have the same trial gain: \QGainVal{} against
\QGainInv{} points, a difference of \QGainDiff{} (\QGainCI{}, Figure~\ref{fig:audit}b). The two gain distributions
differ by only \QTV{} in total variation. Any rule that keeps candidates by trial gain then mislabels valid and invalid
rules at rates that sum to at least \QBound{} \citep{guo2026seal}. The training data explains part of this.
\QSrcOut{} of the \QSrcInv{} invalid rules fail only in situations ExploreToM never contains, such as sleeping, private
telling, lies and peeking. Invalid rules also pile up in the libraries over rounds (Figure~\ref{fig:audit}c). Only
\QRepThreeValid{} of the \QRepThreeDec{} decided rules that pass the 3-point margin are valid.

\paragraph{What each loop kept.} Figure~\ref{fig:progress} follows three loops round by round. Among kept rules the
audit could decide, the invalid ones number \QAdmLoop{} for the accuracy-only loop, \QAdmAO{} for the accuracy-only
library, \QAdmVC{} for the gated library and \QAdmVCs{} for the gated single principle.

\paragraph{An example.} One kept rule of a gated library reads \emph{when a question asks whether A believes B knows the
resulting state, and B was present when the event occurred: assume A believes B knows it} (\boxref{box:vclib}). Both
translations answer ``yes'' when B was in the room but asleep. The formal model answers ``no'', because being there is
not seeing. ExploreToM never puts a character to sleep, so no trial cohort could penalise this rule. It also passed the
gate, because none of the 96 gate pairs asks this second-order question.

\paragraph{Counterexamples matter, but only a little.} On \QConfNCx{} rendered counterexamples, the rule turns a right
answer wrong in \QInducedCx{}\% of cases. On \QConfNCt{} random control stories it does so in \QInducedCt{}\%.

\paragraph{The conditions do not decide when a rule applies.} Removing every condition from the gated libraries changes
BigToM-U pair validity by \QActdNoCond{} (\QActciNoCond{}). Giving each rule another rule's condition changes it by
\QActdSwap{} (\QActciSwap{}). The reader treats the rules as general hints and weighs them against the story.

\subsection{Discussion: What Does the Loop Learn?}
\label{sec:discussion}
The evolved strategies are useful, and most of them are wrong as rules. The reader explains how both can hold. It
weighs a rule against a story in which perception is already marked. It can follow a wrong rule where it helps and
override it where the story disagrees, which fits the small effect of counterexamples and the irrelevant conditions. A
strategy handed to another reader or a person carries the rule itself, so the rule needs its own check. The expert
principle is the opposite case. It is correct and leads on test Main, where conversations dominate, but it loses on
BigToM-U, where seeing a change is the whole question. The loop has learned behaviour that helps this reader, and it
has written that behaviour down as rules that do not hold on their own.

\section{Conclusion}
\cogrsi{} evolves strategies for reasoning about other minds. The evolved strategies transfer to independent BigToM
pairs, where the gated library reaches \QBUVC{}\% pair validity against \QBUHand{}\% for an expert principle. A
calibrated margin raises the share of helpful updates from \RPrecLoop{}\% to \RPrecAO{}\%. A semantic audit shows that
\QAudPropShare{}\% of the rules the loops propose give a wrong answer in some situation they claim to cover. Neither accuracy nor counterfactual
gating catches this. Self-improving systems that reason about minds should check both what their strategies do and
what they say, and using the audit during registration is the natural next step. Our replay suggests that few
current candidates would pass, which moves the bottleneck from selecting strategies to proposing correct ones.

\clearpage
\section*{Limitations}
\begin{itemize}[leftmargin=*, itemsep=1pt, topsep=2pt]
\item All loops learn from one story generator. The formal model covers physical access and testimony in one room,
and other mental states are left for future work.
\item The audit relies on two models translating rules into code. Shared counterexamples guard against a single
mistranslation, and both translations can still miss what a rule's author meant.
\item Loops run three to six seeds, and some arms were not evaluated on every test set (Table~\ref{tab:arms}).
\item All benchmarks are in English and mostly synthetic.
\end{itemize}

\section*{Ethics, AI-Use and Reproducibility Statements}
\paragraph{Ethics.} Systems that infer what a person knows or believes could be used to profile real people from their
conversations. We use only fictional stories and synthetic dialogues and train no models.

\paragraph{Use of AI assistants.} AI assistants helped write and review code, run the experiments, and draft and edit
text. The authors directed and checked this work and are responsible for all content.

\paragraph{Reproducibility.} Our repository contains the code, prompts, sample manifests, frozen configurations and the
scripts that generate every table, figure and number from the per-run outputs. Appendix~\ref{app:prompts} gives all
prompts, Appendix~\ref{app:algo} the algorithms and settings, Appendix~\ref{app:bench} the samples and metrics, and
Appendix~\ref{app:exp} every run. Code: \url{https://github.com/koanatakiyo/cogRSI}.

\bibliography{custom}

\clearpage  %
\appendix
\raggedbottom
\makeatletter
\let\apx@table\table\let\apx@figure\figure
\renewcommand{\table}[1][]{\apx@table[htbp]}\renewcommand{\figure}[1][]{\apx@figure[htbp]}
\setlength{\@fptop}{0pt}\setlength{\@fpsep}{14pt plus 2pt}\setlength{\@fpbot}{0pt plus 1fil}
\setlength{\@dblfptop}{0pt}\setlength{\@dblfpsep}{14pt plus 2pt}\setlength{\@dblfpbot}{0pt plus 1fil}
\makeatother

\section{Prompts}
\label{app:prompts}
All prompts of \reader{} (Boxes (\ref{box:solver})--(\ref{box:arbiter})), read from the frozen source: the prompt text is
verbatim, bold headers and mode tables are layout, and monospace braces mark placeholders. Every call uses the system
message of \boxref{box:solver}; samples are drawn at temperature 0.7, the perception and arbitration calls at temperature 0. The
perception call (one per story or conversation) uses PercepToM's perception rule with indexed outputs constrained by a
JSON schema of the unit IDs; in conversations, belief questions about named characters use a perspective view. The
principle slot holds the hand-written principle (\reader{}), a loop-written one (\readerL{}; \boxref{box:principles})
or nothing (Table~\ref{tab:ablation}).
\promptbox{Solver}{System}{%
\bh{Instructions}This is a theory-of-mind test. Answer the question about what the characters perceived, know, or believe, based only on the given story or conversation.
\bh{Answer format}\ph{answer format of the benchmark}
\bh{Reply}\par\smallskip\noindent\begin{tabular}{@{}p{1.95cm}@{\hspace{4pt}}p{\dimexpr\linewidth-1.95cm-4pt\relax}@{}}
\toprule
Style & Text \\
\midrule
default & Reply only as JSON: \texttt{\{"reasoning": "\ph{step-by-step reasoning}", "answer": \ph{answer}\}}. Think step by step in the reasoning field before giving the answer. \\
concise retry & Reply only as JSON: \texttt{\{"reasoning": "\ph{brief step-by-step reasoning, at most 120 words}", "answer": \ph{answer}\}}. Reason briefly before giving the answer. \\
\bottomrule
\end{tabular}}{Solver: the system message of every call; the concise reply is the retry style.}{box:solver}

\promptbox{Perception call}{User Prompt}{%
\bh{Input}Numbered source units in original order:
\codebox{[\{"id": "s0001", "text": "\ph{unit 1}"\}, \{"id": "s0002", "text": "\ph{unit 2}"\}]}
\bh{Perception rule}\par\smallskip\noindent\begin{tabular}{@{}p{1.95cm}@{\hspace{4pt}}p{\dimexpr\linewidth-1.95cm-4pt\relax}@{}}
\toprule
Format & Rule \\
\midrule
narratives & Assume that characters in the story can perceive every scene occurring in their location but not scenes occurring elsewhere. Also, include the actant of any action as a perceiver of that action. \\
conversations & Assume that characters in the story can hear every utterance that occurs while they are involved in the dialogue, but not those that occur when they are absent. Also, ensure that the speaker of each utterance is included in the audience. \\
\bottomrule
\end{tabular}
\bh{Output}Return only a JSON object mapping EVERY provided unit ID to its list of perceivers/audience. Do not output source text, explanations, or new IDs. Use consistent character names from the source. An individual list may be empty when no perceiver is supported; do not invent a perceiver just to fill a list. Never omit a unit ID. These encoding rules do not change the perception rules above.}{Perception call, one per story or conversation.}{box:perception}

\promptbox{Annotated view}{User Prompt}{%
\bh{Instructions}Each sentence or utterance below is followed by \texttt{[perceived by: ...]}, the characters who witnessed it. A character knows only what they perceived or were told; use these annotations to track what each character knows and believes.
\bh{Story}\codebox{\ph{sentence 1} [perceived by: \ph{names}]\\
\ph{sentence 2} [perceived by: \ph{names}]\\
...}}{Annotated view: each unit with its perceivers.}{box:annotated}

\promptbox{Perspective view}{User Prompt}{%
\bh{Header before the visible utterances}\par\smallskip\noindent\begin{tabular}{@{}p{1.95cm}@{\hspace{4pt}}p{\dimexpr\linewidth-1.95cm-4pt\relax}@{}}
\toprule
Question & Header \\
\midrule
first order & Here are the past utterances in sequence that \ph{char1} is aware of. \\
second order & Here are the past utterances in sequence that \ph{char1} thinks \ph{char2} is aware of. \\
\bottomrule
\end{tabular}}{Perspective view for belief questions in conversations.}{box:perspective}

\promptbox{Arbiter}{User Prompt}{%
\bh{Input}\ph{story and question as in the benchmark}
\bh{Disagreement}Two careful readings of this story disagree: one answers \ph{A}, the other answers \ph{B}.
\bh{Rule}\par\smallskip\noindent\begin{tabular}{@{}p{1.95cm}@{\hspace{4pt}}p{\dimexpr\linewidth-1.95cm-4pt\relax}@{}}
\toprule
Arbiter & Text \\
\midrule
evidential (\reader{}) & Decide which is correct. A character knows only what the story shows them witnessing, being told, or working out from what they themselves did. Do not credit a character with knowing about an event unless the story shows them present and noticing it; an event that happens while they are away, busy, or not shown noticing stays unknown to them, and their later actions can reveal what they did or did not notice. If the question does not depend on who knows what, choose the better answer on the story's merits. \\
neutral (control) & Decide which is correct. \\
\bottomrule
\end{tabular}}{Arbitration of view conflicts and close splits.}{box:arbiter}

\section{Proofs}
\label{app:proofs}
\begin{proposition}[Access determines location beliefs]
\label{prop:access}
If agents perceive exactly the events at their location and $\alpha$ is correct, an agent's first-order belief about
an object's location is the location set by the last move of the object in the agent's perspective view.
\end{proposition}
\begin{proof}
Let $o$ change location only through units that place it (an initial description or a move), and let agents perceive
exactly the units at their location, with no communication. Write $V_a$ for agent $a$'s perspective view. By induction
over $t$, after units $u_1,\dots,u_t$, $a$'s belief about $o$ is the location set by the last placing unit in
$V_a\cap\{u_1,\dots,u_t\}$ (undefined if there is none). For $t=0$ there is no belief. Step $t\to t{+}1$: if
$u_{t+1}\notin V_a$, $a$ did not perceive it ($\alpha$ is correct), so $a$'s belief is unchanged, and so is the last
placing unit in $a$'s view. If $u_{t+1}\in V_a$ places $o$, $a$ perceives the placement and believes it, which is the new
last placing unit in $V_a$; if it does not place $o$, neither side changes. Second-order questions use the chain view;
this matches second-order beliefs when co-present agents know of each other's presence, as ExploreToM's generator
assumes.
\end{proof}

\begin{proposition}[Registration only climbs]
\label{prop:reg}
Along a seed of the principle loop, every admitted principle beats the incumbent it replaces by more than $\delta$ in
mean accuracy on the $K$ trial cohorts of its round, and the reader changes only when a principle is admitted.
\end{proposition}
\begin{proof}
In each round every candidate $x$ and the incumbent $P$ answer the same $K$ trial cohorts, and $\Delta A_k(x)$ is the
accuracy of $x$ minus that of $P$ on cohort $k$. By Eq.~\ref{eq:reg} the candidate with the largest mean gain replaces $P$
only if that mean exceeds $\delta$. If none is admitted, $P$ is kept, and the reader, which depends on the loop only
through $P$, is unchanged. The bound holds on the trial cohorts, not on the population; the final cohort
(Appendix~\ref{app:rsip}) checks the last incumbent on new questions (from stories the loop had seen).
\end{proof}

\begin{proposition}[Arbitration is local]
\label{prop:local}
\reader{}'s answer differs from the vote over its six samples only on view conflicts and close splits; it never
overturns an answer held by five or six samples and always returns a sampled answer. Hence arbitration changes
accuracy by at most the share of arbitrated questions.
\end{proposition}
\begin{proof}
The answer equals the vote unless an arbitration call is made and its choice is accepted; a call is made only on a view
conflict or a close split, and a choice is accepted only if it equals one of the two answers put to the arbiter, both
sampled. Suppose $Y$ holds at least five of six votes. The runner-up has at most one vote, so there is no close split. A
view conflict needs both plain samples to agree and the annotated majority to differ: if they agree on $Y$, at least
three of the four annotated samples are $Y$, so the annotated majority is $Y$; if they agree on another answer, that
answer has two votes and $Y$ at most four, a contradiction. So no call is made, and accuracy differs from the vote's only
on arbitrated questions.
\end{proof}

\section{Algorithm details}
\label{app:algo}
\begin{algorithm}[t]
\caption{Artifact-level RSI (Stages 1 and 2)}
\label{alg:rsi}
\small
\begin{algorithmic}[1]
\REQUIRE development pool $\mathcal{D}$, builder $B$, system $S_0$ (empty library or $P=\emptyset$), rounds $R$, $K$, $\delta$
\FOR{$r=0,\dots,R-1$}
\STATE draw a discovery cohort and $K$ trial cohorts $T_1..T_K$ from unused stories of $\mathcal{D}$
\STATE $F\leftarrow$ failures of $S_r$ on the discovery cohort (Stage 2: grouped by question type $\times$ belief order)
\STATE $X\leftarrow B(F)$; drop candidates that break the interface (Stage 1) or name shown characters (Stage 2)
\FORALL{$x\in X$}
\STATE $g(x)\leftarrow\frac{1}{K}\sum_k \mathrm{acc}(S_r\oplus x,T_k)-\mathrm{acc}(S_r,T_k)$ \hfill (Eq.~\ref{eq:reg})
\ENDFOR
\STATE $x^*\leftarrow\arg\max_x g(x)$
\IF{$g(x^*)>\delta$}
\STATE $S_{r+1}\leftarrow S_r\oplus x^*$ \hfill (Stage 1 arm A: add the tool; Stage 2: replace $P$)
\ELSE
\STATE $S_{r+1}\leftarrow S_r$
\ENDIF
\ENDFOR
\RETURN $S_R$ (arm C: rebuild once from round-0 failures pooled over $\mathcal{D}$, same budget)
\end{algorithmic}
\end{algorithm}
\begin{algorithm}[t]
\caption{\reader{} (Stage 3), one question}
\label{alg:reader}
\small
\begin{algorithmic}[1]
\REQUIRE context $c$, question $q$, principle $P$
\STATE $\alpha\leftarrow$ \textsc{Perceive}$(c)$ \hfill (who witnessed each unit; one call per context)
\IF{$c$ is a conversation and $q$ asks a named chain's belief}
\STATE $v\leftarrow$ perspective view of the chain
\ELSE
\STATE $v\leftarrow$ $c$ annotated with $\alpha$
\ENDIF
\STATE $Y\leftarrow$ four samples on $(v,P)$ and two on $c$ \hfill (temperature 0.7; concise retry)
\STATE $\hat y\leftarrow$ majority of $Y$
\IF{view conflict, or close split (top $\le 4/6$, runner-up $\ge 2/6$; narratives)}
\STATE $y'\leftarrow$ \textsc{Arbitrate}$(c,\alpha,$ top two answers$,P)$; \textbf{if} $y'$ is one of them \textbf{then} $\hat y\leftarrow y'$
\ENDIF
\RETURN $\hat y$
\end{algorithmic}
\end{algorithm}

\subsection{RSI over procedures}
\label{app:algo-proc}
Algorithm~\ref{alg:rsi} runs on the MedRSI code base \citep{wu2026medrsi} at a pinned commit, with the backbone as
solver and builder. Each round the builder reads prioritised failures of the current solver on ExploreToM
physical-access development stories (trace, gold and wrong answer) and proposes up to four tools; tools that fail to
load or to export the interface are discarded. A candidate is admitted if it improves a discovery cohort and, by at least
one point on average, three trial cohorts of 40 stories. Table~\ref{tab:rsi-runs} gives every run round by round.
\begin{table*}[t]
\centering
\footnotesize
\setlength{\tabcolsep}{0pt}
\begin{tabular*}{\textwidth}{@{\extracolsep{\fill}}l*{8}{r}rrrrr@{}}
\toprule
Run & r0 & r1 & r2 & r3 & r4 & r5 & r6 & r7 & Adm. & Tools & Access & Nesting & Parsing \\
\midrule
A s0 & 50.5 & 46.6 & 51.5 & 45.8 & 49.7 & 50.3 & 47.6 & 52.5$^{\star}$ & 1 & 32 & 32 & 16 & 10 \\
A s1 & 41.6 & 44.5 & 46.0 & 53.1 & 46.9 & 50.1 & 43.2 & 51.3 & 0 & 30 & 30 & 16 & 12 \\
A s2 & 54.4 & 54.5 & 55.5 & 47.6$^{\star}$ & 52.4 & 52.7 & 59.2 & 50.8 & 1 & 31 & 29 & 17 & 23 \\
C s0 & 52.6 & 46.3 & 51.7 & 46.9$^{\star}$ & 50.5$^{\star}$ & 48.3 & 47.8 & 51.2 & 2 & 31 & 31 & 17 & 29 \\
C s1 & 42.8 & 46.3 & 45.8 & 50.9 & 46.0$^{\star}$ & 48.6$^{\star}$ & 44.4 & 53.8 & 3 & 31 & 31 & 13 & 28 \\
C s2 & 53.2 & 53.6$^{\star}$ & 52.9$^{\star}$ & 47.4 & 51.3 & 54.4$^{\star}$ & 58.4 & 44.3$^{\star}$ & 5 & 28 & 28 & 15 & 0 \\
\bottomrule
\end{tabular*}
\caption{Stage~1 (E1) runs round by round: development accuracy of the solver with the current library (A: inherited, C: rebuilt; $^\star$: tools admitted in that round) and the number of admitted tools (a round can admit more than one); right: the candidate tools each run wrote and how many encode perceptual access, nested belief or story parsing (a tool can encode several). \textbf{Admission is rare, so the inheriting arm rarely has a library to inherit.}}
\label{tab:rsi-runs}
\end{table*}

\subsection{RSI over principles}
\label{app:rsip}
Development data: ExploreToM \texttt{e1\_dev} (29{,}752 questions over 2{,}519 stories in 2{,}189 structural groups; the sampler draws by story), disjoint from every
test sample. Every cohort takes one question from each of a set of unused stories: per round a discovery cohort of 80 and three
trial cohorts of 80; after eight rounds a final cohort of 300 compares no principle, the hand-written principle of
\reader{} and the loop's principle. Eight rounds and the final cohort need more stories than the pool holds: once every
story has been used, a new pass reshuffles the stories and takes from each a question not used before. Seeds 0--2 ran
the first seven rounds before this rule existed and were resumed from their saved rounds with it, which reproduces their
earlier draws exactly; later seeds used it from the start. Builder: the backbone at temperature 0.7; one call per
failure group (the four largest groups, up to four failures each), replying with a one-sentence diagnosis and a
principle of at most 80 words that must not name the shown stories' characters or places (checked against the
perception call's names and capitalised words; violators and principles over 120 words are rejected). Test runs use the
frozen \reader{} runner with its two texts replaced by the seed's principle (\boxref{box:principles}).
\begin{figure*}[t]
\promptbox{Loop-written principles}{Model Output}{%
\noindent\textbf{Every seed's principle says how a character comes to believe or know a state (50--67 words).}
\codebox{seed 0: To predict where A believes B will search, identify the last location B moved the object that A also witnessed. Ignore any moves B made that A did not see, and ignore the object's true current location. A's prediction is based solely on the most recent move in the shared history.\\[3pt]
seed 1: A character believes an object has a state if they entered the location after that state was established and no intervening action altered it while they were absent. Characters retain beliefs about object states from their last direct observation or from the state existing in the location when they arrived.\\[3pt]
seed 2: When determining if a character believes X, first verify if X is an objective fact in the story. If X is true, assume the character believes X unless the story explicitly states they were misled, told a falsehood, or have direct evidence contradicting X. Do not assume ignorance solely because the character did not witness X.\\[3pt]
seed 3: To determine what A thinks B believes: If B's last perceived event occurred before the target event, and A knows B was present or unobserved leaving before the target event, A assumes B believes the target event occurred. If A knows B was absent or saw an earlier state, A assumes B believes the earlier state.\\[3pt]
seed 4: To predict where A thinks B will search: First, determine B's last observation of the object. If B never saw the object move to a new location, A predicts B searches that last observed spot. However, if A knows B entered the scene *after* the object was moved to a new location, A predicts B searches that new location, as B's knowledge starts from their entry point.\\[3pt]
seed 5: Assume a character believes another knows a fact if the first character knows the second was physically present when the event occurred, unless explicitly told the second was unaware. Do not require the second character to have consciously witnessed the event; mere presence suffices for the first character to attribute knowledge.}}
{The principles written and admitted by the principle-level loop (\S\ref{sec:loop2}), one per seed, verbatim.}{box:principles}
\end{figure*}

\subsection{\reader{}: reading and design history}
\label{app:reader}\label{app:variants}\label{app:v6dev}
Algorithm~\ref{alg:reader} gives the reader. Ties in the vote are broken by the first sampled answer; samples that hit
the 1{,}024-token cap are retried once with a concise instruction. \reader{} was built in three steps, each frozen and
then tested on every test set (Table~\ref{tab:variants}): a vote over the annotated and plain views (frozen at commit
6035943), the arbitration of view conflicts (c86b903) and the arbitration of close splits (e59dcbb). The first two were
frozen before any of their test results were seen. The third was developed afterwards on development data only
(held-out sample~1 and development samples of four further ToM benchmarks) after inspecting the second design's test
failures for diagnosis; six candidates were tried and five rejected on development data:
\begin{itemize}
\item \textbf{Access gate}: plain sampling when the perception call finds no access asymmetry. Development: held-out-1 narrative items: 79.7 / 78.9 (classifier) vs 80.0 for the base design.
\item \textbf{Construct router}: emotion and non-literal questions answered from plain samples. Development: held-out-1: 80.0, equal to the base design.
\item \textbf{Last-witnessed trace}: explicit trace of what each character last witnessed. Development: Hi-ToM 44.2 vs 66.4 for the base design.
\item \textbf{Extra voters}: zero-shot, RSI-library and SC answers added to the vote. Development: lower than the base design on most development datasets.
\item \textbf{Conflict-only gate}: plain reading on conflicts the classifier deems access-free. Development: held-out-1: 80.4 vs 81.2 for the base design.
\end{itemize}

The adopted change applies the existing arbitration call, unchanged, to every remaining close split
(Table~\ref{tab:v6dev}); it was frozen with byte-identical source checks, run once on all test data, and every variant
was re-run on the confirmatory sample (Table~\ref{tab:t1}B).
\begin{table}[t]
\centering
\small
\begin{tabular}{llrcc}
\toprule
Sample & Benchmark & $n$ & without & with \\
\midrule
Held-out 1 & SimpleToM & 501 & 91.8 & 93.2 \\
 & Hi-ToM & 500 & 66.4 & 66.8 \\
 & OpenToM & 500 & 77.4 & 77.0 \\
 & ToMBench & 500 & 76.4 & 77.2 \\
 & CogToM & 500 & 85.4 & 85.6 \\
 & BigToM & 504 & 89.9 & 90.3 \\
\bottomrule
\end{tabular}
\caption{Development results without and with the arbitration of close splits, on the same six samples per question on held-out sample 1, the development set of the design without it. Rejected candidates are listed in the text.}
\label{tab:v6dev}
\end{table}

\begin{table*}[t]
\centering
\small
\setlength{\tabcolsep}{4pt}
\resizebox{\textwidth}{!}{%
\begin{tabular}{lccccc}
\toprule
 & \multicolumn{2}{c}{Main backbone} & \multicolumn{3}{c}{Other backbones (FAN, Simple, Hi, Big)} \\
\cmidrule(lr){2-3}\cmidrule(lr){4-6}
Design & Main sample & FANToM All$^*$ & Llama-70B & GLM-4.6V & Qwen3.5-9B \\
\midrule
SC ($k{=}6$) & 68.3 & 17.6 & 61.9 & 48.0 & 51.4 \\
\midrule
Vote over both views (v3) & 72.2 & 28.7 & 65.5 & 54.7 & 62.3 \\
+ conflict arbitration (v5) & 72.6 & 27.3 & 67.2 & 53.6 & 60.3 \\
\rowcolor{ours}+ close-split arbitration = \sys{} (v6) & 73.1 & 28.0 & 67.1 & 54.9 & 60.9 \\
\bottomrule
\end{tabular}}
\caption{\sys{}'s design variants on every test condition (three runs each; per-benchmark results and every run in Table~\ref{tab:runs-h2}). Each design was frozen before its test runs; the close-split arbitration was added after the first test run of the design without it and was selected on development data only (\S\ref{app:v6dev}).}
\label{tab:variants}
\end{table*}

\section{Validity-aware RSI: details}
\label{app:vc}
\paragraph{Pre-registration.} Every experiment in this section was registered before its first output (files
\texttt{configs/prereg\_J099}--\texttt{J104} in our repository) with its arms, seeds, thresholds, hypotheses and failure
handling. For the principle arms the primary hypothesis H1 is \cogrsi{}'s library minus the accuracy-only loop on test
Main. The secondary hypotheses, Holm-corrected together, are the library against the hand-written principle (H2), the
library minus the accuracy-only library on held-out pair validity (H3), the library minus the single principle (H4) and
the library minus GEPA (H5), the last two on test Main. All outcomes are reported, and no arm was rerun or extended after
its outputs existed. Two test cells failed on a connection error and were rerun with identical inputs, as the
registration allows. No seed was missing at the cutoff.

\paragraph{Loop settings.} All principle arms use the frozen reader and backbone and ExploreToM development data only:
eight rounds, a discovery cohort of 80 new questions, three trial cohorts of 80, up to four candidates per round and a
final cohort of 300 unseen questions shared by every comparison of a seed. The builder answers in JSON with a diagnosis,
an operation (\textsc{add}, \textsc{revise}, \textsc{retire}), an entry number, a condition and a rule. The library is
composed as a numbered list, ``($i$) When condition: rule'', into both principle positions of the reader. Among
admissible candidates the largest mean accuracy gain wins, with ties broken by the larger gate gain. GEPA
\citep{agrawal2025gepa} optimises the same slot from an empty seed, with the reader as its metric, the reader's failures
with gold answers as its reflective data, and 10{,}240 reader evaluations; it selects on its own validation split of
development questions, so its development accuracy is not comparable and Table~\ref{tab:arms} omits it.

\paragraph{Counterfactual pairs.} Narratives follow ToMi's numbered-sentence format: two characters, an object in a
container, and a move. Conversations are group chats in which one speaker makes a claim. Each pair varies one access
fact. Narrative flip pairs are watch/leave, watch/asleep, told/leave, look/leave, told/lied and look/lied, and narrative
same pairs leave/asleep and watch/told. Dialogue flip pairs are hear/away, hear/muted, relayed/away and relayed/misled,
and dialogue same pairs away/muted and hear/relayed. Gold answers are computed by rule from the access facts. The gate
split has 48 narrative and 48 dialogue pairs and the evaluation split 160 of each, drawn with different seeds from
disjoint name, place, object and template lists. In the look variant the character who missed the move returns and
checks the new container, so the right answer is the object's true location. Table~\ref{tab:arms-detail} gives every
arm by benchmark and by variant.

\paragraph{Registration noise.} The margin was set from development logs of the accuracy-only loop only. A fixed
principle's accuracy varies by \RCohortSD{} points across trial cohorts of 80 questions. Simulating four candidates
with no real gain per round, a 1-point margin admits one in \RNullAdm{}\% of rounds and a 3-point margin in about 10\%.
The gate tolerances come from the reader's run-to-run variation in valid pairs on the counterfactual items.

\paragraph{Selection quality.} For each finished loop we scored every candidate that reached registration, admitted or
not, and its incumbent on the first 120 questions of the loop's final cohort, which no registration step saw. A
candidate helps if it beats its incumbent there. The accuracy-only loop proposed \RNCandLoop{} scored candidates in
\RNRoundsLoop{} rounds and admitted \RNAdmLoop{}. \cogrsi{}'s library proposed \RNCandVC{} and admitted \RNAdmVC{}, and
the accuracy-only library and the single principle admitted \RNAdmAO{} and \RNAdmVCs{} of \RNCandAO{} and \RNCandVCs{}. The
selection gap, the best candidate's held-out gain minus the admitted one's, averages \RGapLoop{}, \RGapAO{},
\RGapVCs{} and \RGapVC{} points in the order of Table~\ref{tab:selection}. With so few admissions per arm the precision
estimates are coarse.

\paragraph{Matched inheritance.} Arm M runs the procedure loop with A's inheritance (its builder sees and composes
registered tools, which persist) and C's evidence (failures of the round-0 solver), for seeds 0--2 on the same pool and
schedule. On the full sealed FANToM set, with the seed-paired bootstrap of the original comparison, M $-$ C is
\RInhMC{} points (95\% CI \RInhciMC{}; the 90\% interval lies within the pre-registered 3-point equivalence margin), A $-$
C \RInhAC{} (\RInhciAC{}) and M $-$ A \RInhMA{} (\RInhciMA{}). Inheritance makes no meaningful difference, whichever
evidence the builder sees.

\paragraph{Controlled reader.} To separate the view from the rest of the reader, we replaced the annotated story with
PercepToM's filtered story or SimToM's perspective-taking story and kept perception, sampling, voting, arbitration and
the hand-written principle fixed. On the narrative benchmarks of the main test sample the annotated view beats the
filtered view by \RViewPercep{} Main points (\RViewciPercep{}) and the perspective view by \RViewSimToM{}
(\RViewciSimToM{}). The largest gaps are on Hi-ToM (\RViewHiPercep{} and \RViewHiSimToM{}), where the filters drop
context that higher-order questions need.

\paragraph{Semantic audit (J105).} The oracle (\texttt{src/cogrsi/access\_oracle.py}) samples worlds event by event from
the legal moves: leave, enter, fall asleep, wake, move the object, whisper a location to one listener (true or false),
peek into the container that holds the object, or change a visible or hidden property. Moves and changes are perceived
by the agents present and awake and by the actor; a whisper by its speaker and listener; a peek by the peeker; a
visible property also by anyone who later enters or wakes while the object is in view. Second-order beliefs change
only through events both agents perceive, or when one tells the other. The audited population is every candidate that
reached registration in the accuracy-only loop, the accuracy-only library, the gated library and the gated single
principle (\QAudTexts{} texts; the entry an \textsc{add} or \textsc{revise} writes, or a whole single principle). The
compilers see only the rule text. Before the freeze, three hand-written rules were used to fix the sandbox, the compile
instructions, a deterministic scope guard for second-order questions, private telling and peeking, and the
shared-counterexample criterion. The pre-registered run gave \QAudPreInv{} invalid, \QAudPreVal{} valid,
\QAudPreAmb{} ambiguous and \QAudPreVac{} vacuous texts; \QAudTrunc{} of the ambiguous ones had no code because the
Qwen3.5-122B replies hit the 2{,}048-token output cap. A recorded deviation recompiled those texts with an 8{,}192-token
cap and nothing else changed (\QAudTruncLeft{} still hit it), giving \QAudInv{}, \QAudVal{}, \QAudAmb{} and \QAudVac{}.
Shadow registration replays each candidate under each rule against its own incumbent and logged trial gains; the loops
that ran without the gate were scored on the gate pairs afterwards. Admitted rules that were invalid, by arm:
accuracy-only loop \QAdmLoop{}, accuracy-only library \QAdmAO{}, gated library \QAdmVC{}, gated single principle
\QAdmVCs{}. Without the rule, the reader answers \QWrongWithoutCx{}\% of rendered counterexamples and \QWrongWithoutCt{}\%
of controls wrongly: these stories are hard in themselves.

\paragraph{Objective vs.\ data (J106) and transfer (J107).} Arm B (pairs as data) and seeds 3--5 of the accuracy-only
library were run with the J101 settings; their test-sample and own-pair evaluations were not run before this version.
BigToM-U uses BigToM's conditions that state the initial belief, with true/false-belief flip pairs and true/false
control same pairs (800 pairs from 200 scenarios); every scenario's other version had appeared in an earlier split, so
the set is new in its conditions and pairing, not its scenarios. On flip pairs: gated library \QBUFlipVC{},
accuracy-only library \QBUFlipAO{}, pairs as data \QBUFlipB{}, expert \QBUFlipHand{}, none \QBUFlipNone{}, SC ($k{=}9$)
\QBUFlipSCnine{}, GEPA \QBUFlipGEPA{}. Backward belief (400 pairs): gated library \QBWVC{}, accuracy-only library
\QBWAO{}, expert \QBWHand{}, none \QBWNone{}.

\paragraph{Rule activation (J109).} Intact gated libraries score \QActIntact{} on BigToM-U, \QActNoCond{} with every
condition removed and \QActSwap{} with conditions swapped between entries (flip pairs \QActFlipIntact{},
\QActFlipNoCond{}, \QActFlipSwap{}).
\begin{table*}[t]
\centering
\scriptsize
\setlength{\tabcolsep}{1pt}
\begin{tabular*}{\textwidth}{@{\extracolsep{\fill}}lcccccccccccccccc@{}}
\toprule
 & \multicolumn{9}{c}{Main test sample} & \multicolumn{7}{c}{Pair validity by access variant (eval split)} \\
\cmidrule(lr){2-10}\cmidrule(l){11-17}
Principle & FAN & Simple & Explore & Hi & Open & TBench & Cog & Big & Main & saw & left & present & told & lied & looks & All \\
\midrule
no principle & 23.2 & 73.3 & 72.6 & 69.1 & 74.6 & 74.2 & 84.1 & 90.1 & 70.1 & 100.0 & 100.0 & 100.0 & 100.0 & 99.5 & 34.2 & 91.7 \\
hand (expert) & 28.0 & 90.0 & 72.5 & 69.3 & 75.8 & 74.1 & 84.9 & 90.3 & 73.1 & 100.0 & 100.0 & 100.0 & 100.0 & 100.0 & 33.3 & 91.7 \\
accuracy-only loop & 25.3 & 81.7 & 74.0 & 70.0 & 76.5 & 72.8 & 84.7 & 89.6 & 71.8 & 100.0 & 99.1 & 98.7 & 99.6 & 98.8 & 41.2 & 91.5 \\
GEPA (same slot) & 23.2 & 71.7 & 84.7 & 62.7 & 71.5 & 71.8 & 77.8 & 72.8 & 67.0 & 100.0 & 73.1 & 54.1 & 99.5 & 96.5 & 95.0 & 75.3 \\
accuracy-only library & 23.7 & 81.3 & 76.0 & 70.3 & 75.5 & 72.6 & 84.0 & 88.7 & 71.5 & 100.0 & 95.6 & 99.6 & 99.5 & 98.5 & 51.7 & 91.2 \\
VC, single principle & 26.4 & 81.7 & 75.8 & 70.7 & 75.5 & 72.9 & 83.7 & 88.8 & 71.9 & 100.0 & 98.8 & 100.0 & 100.0 & 100.0 & 90.0 & 98.1 \\
\rowcolor{ours} VC (ours) & 24.5 & 83.6 & 75.6 & 69.2 & 75.8 & 72.9 & 84.4 & 88.6 & 71.8 & 100.0 & 99.6 & 99.6 & 99.8 & 100.0 & 74.2 & 96.4 \\
SC ($k{=}9$) & 18.0 & 71.2 & 72.1 & 64.9 & 74.4 & 73.3 & 85.9 & 89.4 & 68.7 & 100.0 & 100.0 & 100.0 & 98.8 & 100.0 & 10.8 & 88.3 \\
\bottomrule
\end{tabular*}
\caption{\textbf{Arms by benchmark and by access variant} (main backbone, means over seeds or runs). Variant columns give item accuracy on the evaluation split's items of that variant (narrative and dialogue pooled); All is pair validity.}
\label{tab:arms-detail}
\end{table*}

\begin{figure*}[t]
\promptbox{Validity-aware libraries}{Model Output}{%
\noindent\textbf{\cogrsi{}'s final libraries, one per seed, verbatim (the doubled ``When'' is in the logged text).}
\codebox{seed 0: (1) When When the question asks about a character's belief regarding another character's belief about an event, and the second character witnessed the event while the first did not: If Character A did not witness an event but Character B did, and Character A knows Character B is present or aware, assume Character A attributes Character B's knowledge of the event to them, even if Character A lacks direct perception.\\[3pt]
seed 1: (1) When When the question asks where a character will search for an object, and the character's last known perception of the object's location differs from the object's actual final location: Identify the last location the character personally witnessed the object being placed. If the object was subsequently moved by anyone (including the character themselves) after that perception without the character witnessing it, the character will search at their last witnessed location. If the character witnessed the final move, they search there. (2) When When the question asks if Character A thinks Character B believes X, and Character B witnessed an event establishing X, regardless of whether Character A witnessed it: If Character B witnessed the event that makes X true, Character A must attribute that belief to Character B. Character A's lack of perception of the event does not negate their ability to know that Character B knows X.\\[3pt]
seed 2: (1) When When a question asks if a character believes an object has a visible property, and that property was created by an event that occurred before the character entered the location: If the character enters the location after the event, they can perceive the resulting visible state of the object. Assume they believe the object has the property unless told otherwise. (2) When When a question asks if Character A believes Character B knows a past event, and Character B witnessed that event (even in secret) while Character A was unaware of the witnessing: If Character B witnessed the event, Character A believes Character B knows the event occurred if Character A assumes Character B was present or if no information suggests Character A knows Character B was absent during the event. (3) When When a question asks if Character A believes Character B holds a belief about a state, and Character B was present in the location during the event that created that state, regardless of whether B witnessed the event explicitly: If Character B was present in the location when the event occurred, assume Character A believes Character B knows the resulting state, unless A has specific reason to doubt B's perception.\\[3pt]
seed 3: (1) When The question asks where a character will search, and that character was absent during the final move of the object to a new location: The character will search in the location where the object was moved immediately before the question, provided the character who performed that final move is present or the character has no prior knowledge contradicting it. The character's belief is updated by the last known location, not their last witnessed location.\\[3pt]
seed 4: (1) When When a character enters a location after an event and the resulting physical state of an object is visible, they perceive the state: If an object's final visible state (e.g., covered, dyed, damaged) persists after the character arrives, the character believes the object is in that state, regardless of whether they witnessed the cause.\\[3pt]
seed 5: (1) When When the question asks where a character will search for an object, and that character did not witness the last movement of the object by another character: The character will search in the location where the object actually is, as the character's belief is overridden by the fact that the object was moved without their knowledge, but the system must correctly identify the current physical location based on the full narrative sequence. (2) When When the question asks about a character's belief regarding a state of an object that is currently visible or physically present to that character, even if the character did not witness the specific action that created it: If the object's current state is observable by the character upon their return or presence, the character believes the state exists based on visible evidence, regardless of whether they witnessed the event that created it. (3) When When the question asks if Character A thinks Character B believes an object has a state, and Character A knows the object has that state due to visible evidence or a witnessed event, while Character B was absent for the cause but present for the evidence: Character A attributes the belief to Character B if the physical state of the object is visible to Character B, regardless of whether Character B witnessed the event that created the state.}}
{Final libraries of \cogrsi{} (\S\ref{sec:vc}). Each numbered entry is one admitted ADD or REVISE.}{box:vclib}
\end{figure*}
\begin{figure*}[t]
\promptbox{Ablation libraries and principles}{Model Output}{%
\noindent\textbf{Accuracy-only library (gate off)}
\codebox{seed 0: (1) When When the question asks what Character A believes Character B will search for, and Character B did not witness Character A's most recent action: Identify the location of the object according to Character B's last witnessed event. Character A, knowing B's limited knowledge, predicts B will search in that location.\\[3pt]
seed 1: (1) When When the question asks about a character's belief regarding an object's location, and that character moved the object to a location after which no other character moved it while the first character was present: If a character moved an object to a location and was present for subsequent moves, they know the true location. If they were absent for subsequent moves, they believe the location held while they were absent, unless they moved it again later. (2) When When the question asks if Character A thinks Character B believes a state, and Character B witnessed the state directly: If Character B witnessed the state, assume Character A attributes this belief to B, even if A did not witness the state themselves, unless A knows B is mistaken. (3) When When a character enters a location after an object's state has changed, and the object remains in that state without further changes: If a character did not witness a state change but the object is currently in that changed state, assume they believe the object is in that state.\\[3pt]
seed 2: (1) When When a question asks if a character believes a state exists, and that state was created or altered by an event perceived by someone who was present before the question-asker entered: If a character enters a location after an event occurred there, assume they can perceive the resulting state of objects unless the story explicitly states they cannot. Infer belief in the state based on the physical reality established by the event.}
\noindent\textbf{Single principle under the full rule}
\codebox{seed 0: A character knows an object's state if they perceive the change, OR if they are in a location where the object is present and its state is visible to them. Characters do not need to witness the specific event causing a change if they enter after it and can see the result.\\[3pt]
seed 1: If Character A performs an action, they know the result. If Character B enters after the action but before any concealment, assume B perceives the result unless told otherwise. When inferring what A thinks B believes, if A knows B was present during the event, A assumes B knows, even if A did not see B's eyes.\\[3pt]
seed 2: 1. Characters know the current state of objects in a location if they entered after a change occurred and the state remains visible. 2. Characters know about an action if they witnessed it or if the action's result is visibly present when they enter. 3. Do not assume ignorance of visible states simply because the character did not witness the specific moment of change.}}
{Final artifacts of the two ablations (\S\ref{sec:vc}), verbatim.}{box:ablib}
\end{figure*}
\begin{figure*}[t]
\promptbox{GEPA principles (same slot)}{Model Output}{%
\noindent\textbf{The first 90 words of each seed's final GEPA principle; full texts are in our repository.}
\codebox{seed 0 (2{,}004 words; first 90): You are an expert in Theory of Mind reasoning, multi-agent belief tracking, and narrative logic. Your task is to answer questions based on provided stories involving characters performing actions in specific settings. You must rigorously track events, physical perception, and resulting nested beliefs (e.g., "What A thinks B knows"). The primary goal is to predict where a character (B) will search for an object based on what another character (A) *thinks* B knows, or to determine the truth value of nested belief statements. \#\#\# Critical Logic Rules (Domain Specific \& \dots\\[3pt]
seed 1 (1{,}838 words; first 90): You are an expert in Theory of Mind reasoning and logical deduction, specifically designed to solve complex nested belief problems involving object states, movements, and perception constraints. \#\#\# Task Description Your goal is to analyze narrative stories containing sequences of events, character actions, and explicit perception tags. You must answer multiple-choice questions regarding what a character thinks or believes about a specific state of an object, or where a character thinks another character will search for an object. \#\#\# Core Logic \& Rules for Perception \& Belief To solve these \dots\\[3pt]
seed 2 (2{,}383 words; first 90): \#\#\# Task Instructions: State Tracking, Belief Proxy, and Temporal Logic **Objective**: Your task is to answer questions regarding the state, location, or attributes of objects in a story. You must strictly distinguish between **Objective Reality** (what actually happened) and **Character Perspective** (what a character knows), applying specific domain logic based on the question type. **Crucial Domain Rules \& Definitions**: 1. **Physical Reality is Absolute (The "Truth" Layer)**: * Events (moving, breaking, bending) occur in the story's timeline regardless of witness. * The `[perceived by: ...]` tag indicates the actor or \dots}}
{Final principles of GEPA in the reader's principle slot (\S\ref{sec:loop2}), excerpts.}{box:gepa}
\end{figure*}

\section{Benchmark details}
\label{app:bench}
\subsection{Task design}
A system receives a context (story or conversation) and a question in the benchmark's native format and returns an
answer in that format; it never sees gold answers or test items during development. The RSI loops see only ExploreToM
development stories; \reader{}'s design used held-out sample~1.

\begin{figure*}[t]
\centering
\includegraphics[width=\textwidth]{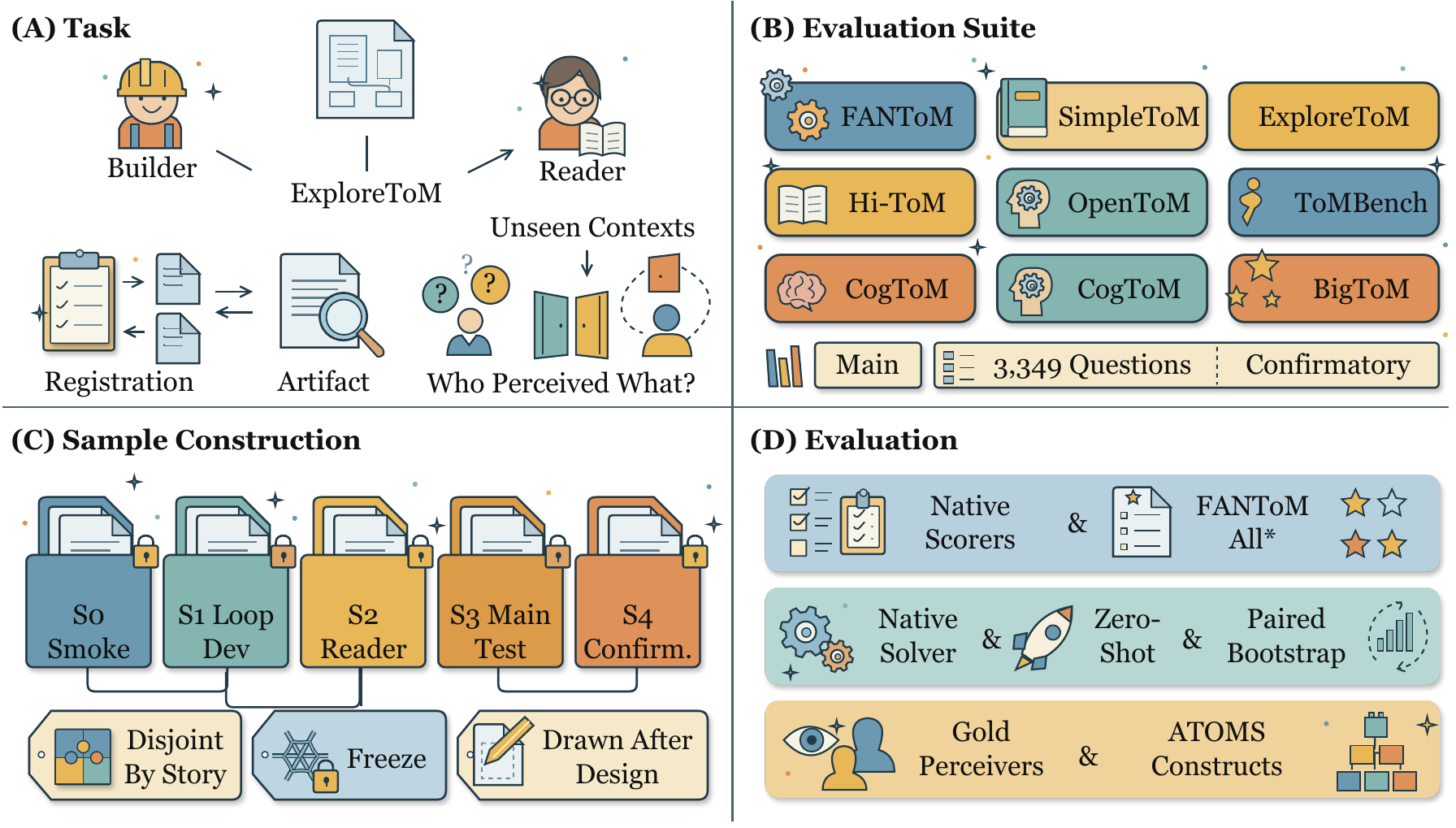}
\caption{\textbf{Task and suite.} (A) Task. (B) Eight benchmarks. (C) Sealed, disjoint samples. (D) Native scoring.}
\label{fig:suite}
\end{figure*}

\paragraph{Sample construction.} Samples are drawn by story with fixed seeds and are disjoint by construction
(Figure~\ref{fig:suite}C). \textbf{S0} holds the smoke subsets (never reported), \textbf{S1} the development data,
\textbf{S2} held-out sample~1 (\reader{}'s development set), \textbf{S3} held-out sample~2 (the main test sample) and
\textbf{S4} held-out sample~3, drawn after every design decision. Every design was frozen, with byte-identical source
checks, before its test runs (Appendix~\ref{app:reader}). Two qualifications apply. Close-split arbitration was developed
after inspecting the test failures of the design without it, so main-sample results for the final reader are exploratory
and S4 is its confirmatory test. For FANToM, SimpleToM and ExploreToM, every held-out sample, S4 included, is a subset of
the sealed test sets on which the baselines and Stage-1 libraries ran before \reader{} existed, and FANToM baseline
failures from those sets were read for diagnosis. For these three benchmarks S4 is independent of every \reader{} design
decision, not unseen by the project.

\subsection{Construction}
\label{app:bench-construction}
All inputs are text. MMToM-QA and MuMA-ToM are run in their text-only versions (the stories and dialogues released
with them; no video), and Social-IQ 2.0 as a written description of each clip followed by its transcript; the
descriptions were written once from the clip frames by the backbone with image input enabled for that pass only. No
system sees images or video, so these results say nothing about multimodal transfer.
Each benchmark's test pool was sealed before the formal runs. Samples are drawn by story, conversation or ExploreToM
structural group with fixed seeds (smoke and held-out~1: 0, held-out~2: 1, confirmatory: 3), and each excludes every unit
of the earlier ones; the confirmatory draw is disjoint from the development data, the smoke subsets and held-out
samples~1 and~2. For FANToM, SimpleToM and ExploreToM all three held-out samples are subsets of the sealed test sets on
which the baselines and Stage-1 libraries were evaluated on October 6--9, before \reader{} existed
(Tables~\ref{tab:sealed}--\ref{tab:sealed-runs}), and FANToM baseline failures from those sets were read for diagnosis.
No \reader{} design decision used held-out sample~3, but for these three benchmarks its independence is at the level of
design decisions, not of exposure to the project. Table~\ref{tab:suite} gives the sizes. ToMBench items with repeated IDs were disambiguated and options
with empty text removed before any draw.
\begin{table*}[t]
\centering
\footnotesize
\setlength{\tabcolsep}{6pt}
\resizebox{\textwidth}{!}{%
\begin{tabular}{llccc}
\toprule
Benchmark & Input & Held-out 1 & Main test & Confirm. \\
\midrule
FANToM & conversation & 84 conv. & 187 sets (74 conv.) & 75 conv. \\
SimpleToM & story & 334 (167) & 334 (167) & 334 (167) \\
ExploreToM E1 & story & 253 (51) & 254 (52) & 250 (52) \\
ExploreToM E2 & story & 254 (51) & 250 (50) & 254 (51) \\
Hi-ToM & story & 500 (102) & 502 (101) & 70 (14) \\
OpenToM & story & 500 (100) & 500 (100) & 500 (100) \\
ToMBench & story & 500 (243) & 505 (253) & 500 (246) \\
CogToM & story & 500 (309) & 500 (257) & 502 (292) \\
BigToM & story & 504 (84) & 504 (84) & 60 (10) \\
\bottomrule
\end{tabular}}
\caption{Samples per benchmark: questions (stories or groups in parentheses; FANToM: All$^*$ sets and conversations). Held-out sample~1 is \reader{}'s development set; the main test sample is held-out sample~2; the confirmatory sample is held-out sample~3. SimpleToM: behavior and judgment questions scored.}
\label{tab:suite}
\end{table*}

\subsection{Metrics}
\label{app:metrics}
Accuracy is each benchmark's own rule: option match for multiple choice, the benchmarks' scripts for free text and
lists; FANToM uses its official scorer on short contexts, and All$^*$ counts a target set as right only if every question
in it is right. PercepToM's FANToM runs are scored from their per-question results like every other system, because the
official scorer's printed summary for those runs cannot be reproduced from the per-question results it writes (for every
other system the two agree exactly). ATOMS constructs are mapped from each benchmark's question type by a fixed table; no
independent annotation validates these labels. Paired differences are bootstrapped over stories or conversations (2{,}000
resamples); within-benchmark construct effects centre each question's difference on its benchmark's mean.

\section{Baseline details}
\label{app:baselines}
All baselines use the backbone, decoding settings and output cap of \reader{} (Table~\ref{tab:cost}).
\textbf{Native solver}: the Stage-1 solver with an empty library (floor). \textbf{Zero-shot}: the question and context
with the answer-format instruction. \textbf{CoT} \citep{kojima2022large}: step-by-step reasoning before the answer.
\textbf{SC ($k{=}6$), SC ($k{=}9$)} \citep{wang2023selfconsistency}: majority over $k$ CoT samples at temperature 0.7,
$k{=}6$ being \reader{}'s number of samples and $k{=}9$ its tokens per question, run by the frozen \reader{} runner with $k$
fixed. \textbf{Retrieval}: the most similar development failures (sentence embeddings of question and story), shown as
solved examples with their correct answers. \textbf{GEPA} \citep{agrawal2025gepa}: prompt-only optimisation on ExploreToM
development data, three optimizer seeds. \textbf{SimToM} \citep{wilf2024simtom}: a first call asks which events or
utterances the target character knows about, a second answers from them with our CoT prompt (CoT when no character is
named). \textbf{PercepToM} \citep{jung2024perceptom}: its perception and perspective-filtering steps with a zero-shot
fallback when the target character is not found; \textbf{PercepToM + SC (belief)} on FANToM: six samples over its
filtered conversation for belief questions, six plain samples otherwise, majority vote. \textbf{PercepToM + SC} and
\textbf{SimToM + SC}: \reader{}'s six samples at temperature 0.7 drawn over PercepToM's filtered context (every
question) or after SimToM's perspective-taking call, with SimToM's answer prompt, majority vote. \textbf{\libA{}/C}: the Stage-1
libraries, inherited and rebuilt, three loop seeds. Not run: DEL-ToM \citep{wu2025deltom} needs a verifier trained on
its own simulator; AutoToM \citep{zhang2025autotom} and ThoughtTracing \citep{kim2025thoughttracing} answer through a
planning model with a different interface; Decompose-ToM \citep{sarangi2025decompose} and Shoes-of-Others
\citep{shinoda2026shoes} are perspective methods of the kind SimToM and PercepToM represent.
\begin{table}[t]
\centering
\footnotesize
\setlength{\tabcolsep}{3pt}
\resizebox{\columnwidth}{!}{%
\begin{tabular}{lcccc}
\toprule
 & \multicolumn{3}{c}{Per question} & Accuracy \\
\cmidrule(lr){2-4}\cmidrule(lr){5-5}
Method & Calls & In (k) & Out (k) & Main \\
\midrule
CoT & 1.0 & 0.35 & 0.26 & 64.8 \\
SC ($k{=}6$) & 6.0 & 2.11 & 1.67 & 68.3 \\
SC ($k{=}9$) & 9.0 & 3.17 & 2.50 & 68.7 \\
PercepToM & 1.4 & 0.47 & 0.08 & 63.8 \\
\rowcolor{ours}\sys{} & 7.3 & 3.33 & 2.36 & 73.1 \\
\rowcolor{ours}\readerL{} & 7.3 & 3.51 & 2.40 & 71.8 \\
\bottomrule
\end{tabular}}
\caption{Cost per question (multiple-choice test benchmarks, main backbone; all calls, including perception, retries and arbitration) and overall accuracy (Table~\ref{tab:t1}). SC ($k{=}9$) matches \sys{}'s tokens.}
\label{tab:cost}
\end{table}

\section{Experiment details}
\begin{table*}[t]
\centering
\footnotesize
\setlength{\tabcolsep}{8pt}
\resizebox{\textwidth}{!}{%
\begin{tabular}{lcccccc}
\toprule
Method & Hi & Open & TBench & Cog & Big & Mean \\
\midrule
PercepToM (filtered story) & 48.2 & 68.4 & 72.0 & 79.7 & 84.8 & 70.6 \\
SC ($k{=}6$, full story) & 62.2 & 75.0 & 73.9 & 85.5 & 89.2 & 77.2 \\
PercepToM + SC + arbitration & 67.5 & 74.1 & 74.0 & 85.7 & 89.0 & 78.1 \\
\rowcolor{ours}\sys{} (annotated + plain) & 69.3 & 75.8 & 74.1 & 84.9 & 90.3 & 78.9 \\
\bottomrule
\end{tabular}}
\caption{Annotate or filter? Accuracy on the five multiple-choice benchmarks of the main test sample, three runs. The third row gives PercepToM's filtered reading our arbitration: where its answer and self-consistency's differ, the evidential rule of \S\ref{sec:arbitration} decides.}
\label{tab:annfilter}
\end{table*}

\begin{table}[t]
\centering
\footnotesize
\setlength{\tabcolsep}{0pt}
\begin{tabular*}{\columnwidth}{@{\extracolsep{\fill}}lcccc@{}}
\toprule
Backbone & \reader{} & SC ($k{=}9$) & SC ($k{=}6$) & Tokens \\
\midrule
Qwen3.5-35B (main) & \textbf{69.4} & 60.9 & 59.8 & 1.00 \\
Qwen3.5-122B & \textbf{70.4} & 64.2 & 64.2 & 0.98 \\
Llama-3.3-70B & \textbf{67.1} & 62.0 & 61.9 & 1.15 \\
Qwen3.5-9B & \textbf{60.9} & 52.3 & 51.4 & 0.99 \\
GLM-4.6V-Flash & \textbf{54.9} & 47.9 & 48.0 & 1.09 \\
\bottomrule
\end{tabular*}
\caption{\textbf{Token-matched self-consistency on every backbone.} Mean of FANToM All$^*$, SimpleToM, Hi-ToM and BigToM (three runs). On FANToM, Llama-3.3, Qwen3.5-9B and GLM use \reader{} without close-split arbitration, as in Figure~\ref{fig:cost}b. Tokens: SC ($k{=}9$)'s tokens per question relative to \reader{}'s (SimpleToM, Hi-ToM, BigToM; main backbone: all main benchmarks).}
\label{tab:sc9bb}
\end{table}

\begin{table}[t]
\centering
\footnotesize
\setlength{\tabcolsep}{3pt}
\resizebox{\columnwidth}{!}{%
\begin{tabular}{lccccc}
\toprule
 & & \multicolumn{3}{c}{Development (final cohort)} & Test \\
\cmidrule(lr){3-5}\cmidrule(lr){6-6}
Seed & Admitted & None & Hand & Loop & Main \\
\midrule
0 & 4/8 & 61.7 & 61.3 & 63.0 & 70.9 \\
1 & 5/8 & 60.0 & 62.3 & 64.0 & 72.5 \\
2 & 2/8 & 52.3 & 51.7 & 60.0 & 72.1 \\
3 & 3/8 & 58.3 & 59.0 & 61.7 & 72.4 \\
4 & 6/8 & 62.7 & 63.0 & 62.7 & 71.4 \\
5 & 2/8 & 61.0 & 60.7 & 68.0 & 71.6 \\
\bottomrule
\end{tabular}}
\caption{Principle-level RSI, one row per loop seed. Admitted: rounds whose candidate passed the registration rule. Development: accuracy on a final cohort of 300 unseen ExploreToM development questions with no principle (the annotation format only), the hand-written principle of \sys{} and the loop's final principle. Test: one run of the reader with the seed's principle (Table~\ref{tab:t1}). Principles verbatim in Appendix~\ref{app:rsip}.}
\label{tab:rsip}
\end{table}

\begin{table*}[t]
\centering
\footnotesize
\setlength{\tabcolsep}{8pt}
\resizebox{\textwidth}{!}{%
\begin{tabular}{llllc}
\toprule
Method & What is kept & Learned from & Access & Rebuild \\
\midrule
Voyager & code skills & env.\ feedback & -- & \textcolor{neg}{$\times$} \\
Reflexion & verbal memory & task feedback & -- & \textcolor{neg}{$\times$} \\
ADAS, DGM & agent code & benchmark score & -- & \textcolor{neg}{$\times$} \\
GEPA & prompt & rollout reflection & -- & \textcolor{neg}{$\times$} \\
\midrule
SymbolicToM & -- (fixed) & hand-designed & belief graphs & \textcolor{neg}{$\times$} \\
SimToM & -- (fixed) & hand-designed & filtered story & \textcolor{neg}{$\times$} \\
PercepToM & -- (fixed) & hand-designed & filtered story & \textcolor{neg}{$\times$} \\
\midrule
\rowcolor{ours}\cogrsi{} & tools; principle & failures & annotated story & \textcolor{pos}{$\surd$} \\
\bottomrule
\end{tabular}}
\caption{\cogrsi{} against representative self-improving systems and ToM methods. \emph{What is kept}: the artifact
carried to later rounds; \emph{Access}: how perceptual access enters reasoning; \emph{Rebuild}: whether inheritance is
tested against a rebuild with the same budget. Only \cogrsi{} inherits across rounds and keeps the whole story while
marking who perceived each part.}
\label{tab:related}
\end{table*}

\label{app:exp}
\subsection{Protocol and implementation}
\label{app:impl}
Every system runs the same served model per backbone through vLLM 0.30.0 (PyTorch 2.13, Transformers 5.18), thinking
off, 32k context and a 1{,}024-token output cap (Table~\ref{tab:impl}). Each run is one execution over a sample; failed
runs are retried and reported (Table~\ref{tab:runs-h2}). Every frozen design is recorded with its
commit and source hash next to its outputs.
\begin{table*}[t]
\centering
\footnotesize
\setlength{\tabcolsep}{0pt}
\begin{tabular*}{\textwidth}{@{\extracolsep{\fill}}llll@{}}
\toprule
Backbone & Architecture & Weights & GPUs \\
\midrule
Qwen3.5-35B-A3B (main) & MoE, 256 experts, 40 layers & FP8 & 1$\times$H200 / A6000 \\
Qwen3.5-122B-A10B & MoE, 256 experts, 48 layers & FP8 & 1$\times$H200 \\
Llama-3.3-70B-Instruct & dense, 80 layers & AWQ 4-bit & 1$\times$H200 / 2$\times$A6000 \\
GLM-4.6V-Flash & dense, 40 layers, text-only serving & BF16 & 1$\times$A6000 \\
Qwen3.5-9B & dense, 32 layers & BF16 & 1$\times$H200 / A6000 \\
\bottomrule
\end{tabular*}
\caption{Backbones and hardware (NVIDIA H200 NVL 141\,GB, RTX A6000 48\,GB). All served with vLLM 0.30.0.}
\label{tab:impl}
\end{table*}

\subsection{Main results of the reader}
\label{app:rq1}\label{app:runs}
\begin{enumerate}
\item \textbf{\reader{} is first on Main.} On the main backbone it scores \OvVsix{}, against \OvSC{} for SC ($k{=}6$) and
\OvSCnine{} for SC ($k{=}9$). On Qwen3.5-122B and Llama-3.3-70B it scores
\TBQwenLVsix{} and \TBLlamaVsix{} against \TBQwenLSC{} and \TBLlamaSC{}, and it beats SC ($k{=}9$) on all five
backbones at equal tokens (Table~\ref{tab:sc9bb}).
\item \textbf{The margin is largest where access decides the answer.} On FANToM All$^*$ it beats
the strongest baseline by \PairFanVsix{} points (95\% CI \PairFanVsixCI{}), on Hi-ToM by \PairHiToMVsix{}
(\PairHiToMVsixCI{}), and SC ($k{=}6$) on SimpleToM by \PairSimpleVsixSC{}. FANToM and SimpleToM carry \ShareMainFS{}\% of the gain over SC ($k{=}9$) (\ShareConfFS{}\% on the confirmatory
sample), and the other six average \RestMainGain{} points, so the gain targets perception-dependent tasks.
\item \textbf{It holds on the confirmatory sample.} On it (Table~\ref{tab:t1}B), \reader{} scores \HOvVsix{}
(three runs, \HLoVsix{}--\HHiVsix{}) against \HOvSCnine{} for SC ($k{=}9$) (\HLoSCnine{}--\HHiSCnine{}), and \readerL{}
scores \HOvAllRsip{} over six seeds. The vote-only design, frozen before testing, already scores \HOvVthree{}.
\item \textbf{It does not hold on seven further benchmarks.} There \reader{} scores \NewMeanVsix{}, against
\NewMeanSCnine{} for SC ($k{=}9$) and \NewMeanPercepToMSCsix{} for PercepToM + SC, and \readerL{} scores \NewMeanRsip{}
(Table~\ref{tab:newsets}). SimToM + SC is best on ToMi (\NewToMiSimToMSC{} vs.\ \NewToMiVsix{}), and on the
confirmatory items SimToM leads.
\end{enumerate}

Table~\ref{tab:t1-h1} gives the development sample of \reader{}'s arbitration (held-out sample~1, not a test sample). Before
\reader{} existed, the RSI libraries and the baselines were evaluated on the full sealed test sets of FANToM, SimpleToM
and ExploreToM (the planned E1/E2 study, whose primary endpoint is reported in \S\ref{sec:rq3});
Tables~\ref{tab:sealed}--\ref{tab:sealed-runs} give those results, including appendix-only systems.
Table~\ref{tab:runs-h2} lists every run.
The seven test-only benchmarks are in Table~\ref{tab:newsets}. On them \reader{} is level with self-consistency
(\NewMeanVsix{} vs.\ \NewMeanSC{} and \NewMeanSCnine{}), PercepToM + SC is best on average (\NewMeanPercepToMSCsix{}),
and SimToM + SC leads on ToMi (\NewToMiSimToMSC{} vs.\ \NewToMiVsix{}). On the confirmatory sample's five test-only
benchmarks the picture is the same (\HNewAllVsix{} vs.\ \HNewAllSCnine{} for SC ($k{=}9$) and \HNewAllSimToM{} for SimToM).
\begin{table*}[t]
\centering
\footnotesize
\setlength{\tabcolsep}{0pt}
\begin{tabular}{@{}>{\raggedright\arraybackslash}p{4cm}*{8}{>{\centering\arraybackslash}p{\dimexpr(\textwidth-4cm)/8\relax}}@{}}
\toprule
Method & ToMi & Para & Nego & Omni & MMToM & MuMA & SIQ & Mean \\
\midrule
\multicolumn{9}{@{}l}{\textit{Prompting}} \\
Native solver (floor) & 69.3 & \underline{78.6} & 64.6 & 61.8 & 39.2 & 44.3 & 62.5 & 60.0 \\
Zero-shot & 66.2 & 74.4 & 66.3 & \textbf{64.5} & 41.4 & 44.9 & \underline{63.2} & 60.1 \\
CoT & 67.9 & 66.3 & 68.7 & 59.4 & 53.1 & 66.5 & 61.3 & 63.3 \\
SC ($k{=}6$) & 66.1 & 67.1 & 71.7 & 61.1 & 57.7 & 70.1 & \underline{63.2} & 65.3 \\
SC ($k{=}9$) & 66.2 & 66.9 & 72.5 & 61.7 & \textbf{60.1} & \textbf{72.8} & \underline{63.2} & \underline{66.2} \\
\midrule
\multicolumn{9}{@{}l}{\textit{Adapted on ExploreToM dev}} \\
Retrieval & 66.3 & 67.4 & 61.8 & 63.7 & 37.5 & 46.5 & \textbf{64.2} & 58.2 \\
GEPA & 69.5 & 72.4 & 59.9 & 62.8 & 28.3 & 49.1 & 62.1 & 57.7 \\
\midrule
\multicolumn{9}{@{}l}{\textit{ToM procedure}} \\
PercepToM & 70.7 & 73.6 & 66.5 & \underline{63.9} & 40.0 & 45.1 & \underline{63.2} & 60.4 \\
PercepToM + SC & 69.7 & 69.7 & 72.3 & 61.8 & \underline{59.8} & 71.1 & 62.7 & \textbf{66.7} \\
SimToM & \underline{84.6} & 78.4 & 67.7 & 53.4 & 38.8 & 55.5 & 59.7 & 62.6 \\
SimToM + SC & \textbf{85.6} & \textbf{82.2} & 71.3 & 56.4 & 41.8 & 59.6 & 63.1 & 65.7 \\
\midrule
\multicolumn{9}{@{}l}{\textit{RSI over procedures (Stage 1)}} \\
\libC{} (rebuilt library) & 69.5 & 72.4 & 61.2 & 57.8 & 36.1 & 48.1 & 59.3 & 57.8 \\
\libA{} (inherited library) & 76.0 & 72.7 & 58.4 & 58.0 & 36.6 & 45.3 & 60.5 & 58.2 \\
\midrule
\multicolumn{9}{@{}l}{\textit{Grounded reading (Stages 2--3)}} \\
\reader{} (no principle) & 65.6 & 67.3 & 72.0 & 60.7 & 56.4 & \underline{71.3} & 63.1 & 65.2 \\
\reader{} (vote only) & 66.2 & 67.1 & 72.3 & \underline{63.9} & 53.5 & 68.5 & 62.7 & 64.9 \\
\reader{} (no close-split) & 65.6 & 67.4 & 72.0 & 62.3 & 57.0 & 68.6 & 61.9 & 65.0 \\
\rowcolor{ours} \readerL{} & 67.5 & 69.0 & \underline{72.8} & 61.6 & 49.6 & 65.2 & 62.0 & 64.0 \\
\rowcolor{ours} \reader{} & 66.0 & 67.7 & \textbf{73.5} & 62.5 & 57.1 & 67.1 & 62.9 & 65.2 \\
\bottomrule
\end{tabular}
\caption{\textbf{Seven test-only benchmarks} (main backbone, main test sample, accuracy \%, mean of three runs; \readerL{}: six loop seeds). They were added after the first design was frozen and never used for development. Best in bold, second underlined. \textbf{Here \reader{} is level with self-consistency, and SimToM leads on the ToMi family.}}
\label{tab:newsets}
\end{table*}

\begin{table*}[t]
\centering
\footnotesize
\setlength{\tabcolsep}{6pt}
\resizebox{\textwidth}{!}{%
\begin{tabular}{lcccc}
\toprule
\textbf{Method} & \textbf{FAN} & \textbf{Simple} & \textbf{Explore} & \textbf{Overall} \\
 & {\footnotesize All$^*$} & {\footnotesize Beh.+Judg.} & {\footnotesize E1+E2} & {\footnotesize (3/3)} \\
\midrule
\multicolumn{5}{l}{\cellcolor{card}\textit{Qwen3.5-35B-A3B-FP8}\; {\footnotesize(MoE 35B / 3B active, FP8, thinking off)}} \\
Native solver (floor) & 4.3 & 51.9 & 61.3 & 39.2 \\
\midrule
\multicolumn{5}{l}{\textit{Prompting}} \\
Zero-shot & 7.8\up{3.5} & 60.9\up{9.0} & 60.1\dn{1.2} & 42.9\up{3.8} \\
CoT & \underline{14.4}\up{10.1} & \underline{68.4}\up{16.5} & 62.0\up{0.7} & \underline{48.2}\up{9.1} \\
SC ($k{=}6$) & \textbf{18.2}\up{14.0} & 68.4\up{16.5} & 64.8\up{3.5} & \textbf{50.5}\up{11.3} \\
\midrule
\multicolumn{5}{l}{\textit{Adapted on ExploreToM dev}} \\
Retrieval & 2.8\dn{1.5} & 49.7\dn{2.2} & 65.2\up{3.9} & 39.2\up{0.1} \\
GEPA & 0.6\dn{3.7} & 46.7\dn{5.2} & 68.7\up{7.4} & 38.6\dn{0.5} \\
\midrule
\multicolumn{5}{l}{\textit{ToM procedure}} \\
PercepToM & 5.0$^\ddagger$\up{0.7} & \textbf{74.6}\up{22.7} & 62.9\up{1.6} & 47.5\up{8.3} \\
\midrule
\multicolumn{5}{l}{\textit{RSI over procedures (Stage 1)}} \\
\libC{} (rebuilt library) & 2.2\dn{2.1} & 51.6\dn{0.3} & \underline{68.7}\up{7.5} & 40.8\up{1.7} \\
\libA{} (inherited library) & 2.9\dn{1.4} & 52.7\up{0.7} & \textbf{70.7}\up{9.4} & 42.1\up{2.9} \\
\bottomrule
\end{tabular}}
\caption{\textbf{Full sealed test sets (RSI study, design frozen before its runs).} Accuracy (\%) per dataset, full sealed test sets of FANToM, SimpleToM and ExploreToM (October 6--9; before \sys{} v3 existed); smoke subsets excluded. Mean over three runs (\sys{} and baselines; GEPA: three optimizer seeds; Stage-1 tools: three loop seeds; \readerL{}: one run per principle-loop seed). FANToM: official All$^*$ on short contexts. SimpleToM: mean of behavior and judgment. ExploreToM: mean of the E1 (physical access) and E2 (communication access) partitions. Subscripts: gap to the floor (the same native solver with an empty tool library). Best in bold, runner-up underlined. Overall averages the datasets every row has (count in the header). $^\ddagger$PercepToM on FANToM is scored from its per-question results like every other system; the official scorer's printed summary for these runs is higher and cannot be reproduced from the per-question results it writes (for every other system the two agree exactly). Per-metric detail in Table~\ref{tab:sealed-detail}; every run in Table~\ref{tab:sealed-runs}.}
\label{tab:sealed}
\end{table*}

\begin{table*}[t]
\centering
\footnotesize
\setlength{\tabcolsep}{2.6pt}
\resizebox{\textwidth}{!}{%
\begin{tabular}{lccccccccc}
\toprule
\textbf{Method} & \textbf{FAN} & \textbf{Simple} & \textbf{Explore} & \textbf{Hi} & \textbf{Open} & \textbf{TBench} & \textbf{Cog} & \textbf{Big} & \textbf{Overall} \\
 & {\footnotesize All$^*$} & {\footnotesize Beh.+Judg.} & {\footnotesize E1+E2} & {\footnotesize Acc.} & {\footnotesize Acc.} & {\footnotesize Acc.} & {\footnotesize Acc.} & {\footnotesize Acc.} & {\footnotesize (8/8)} \\
\midrule
\multicolumn{10}{l}{\cellcolor{card}\textit{Qwen3.5-35B-A3B-FP8}\; {\footnotesize(MoE 35B / 3B active, FP8, thinking off)}} \\
Native solver (floor) & 3.9 & 51.8 & 67.6 & 49.5 & 65.9 & 76.1 & 80.4 & 84.9 & 60.0 \\
\midrule
\multicolumn{10}{l}{\textit{Prompting}} \\
Zero-shot & 6.8\up{2.9} & 58.0\up{6.2} & 70.4\up{2.8} & 54.2\up{4.7} & 66.7\up{0.8} & 74.7\dn{1.4} & 80.4\eq & 86.0\up{1.2} & 62.2\up{2.1} \\
CoT & 15.0\up{11.1} & 68.6\up{16.8} & 67.8\up{0.2} & 48.5\dn{1.0} & 71.0\up{5.1} & 74.5\dn{1.7} & 81.9\up{1.5} & \underline{88.7}\up{3.8} & 64.5\up{4.5} \\
SC ($k{=}6$) & \underline{19.3}\up{15.4} & 69.5\up{17.7} & 72.3\up{4.7} & \underline{59.7}\up{10.2} & \underline{74.0}\up{8.1} & \underline{76.8}\up{0.7} & \textbf{84.7}\up{4.3} & \textbf{89.2}\up{4.3} & \underline{68.2}\up{8.2} \\
\midrule
\multicolumn{10}{l}{\textit{Adapted on ExploreToM dev}} \\
Retrieval & 2.8\dn{1.1} & 48.5\dn{3.3} & 69.8\up{2.2} & 51.2\up{1.7} & 65.9\dn{0.1} & 75.5\dn{0.7} & 80.3\dn{0.1} & 83.9\dn{0.9} & 59.7\dn{0.3} \\
GEPA & 0.5\dn{3.4} & 44.2\dn{7.6} & 72.8\up{5.2} & 47.5\dn{2.1} & 66.2\up{0.3} & 73.2\dn{2.9} & 76.8\dn{3.6} & 81.0\dn{3.8} & 57.8\dn{2.2} \\
\midrule
\multicolumn{10}{l}{\textit{ToM procedure}} \\
PercepToM & 4.2\up{0.3} & \underline{86.6}\up{34.8} & 69.6\up{2.0} & 49.1\dn{0.5} & 66.6\up{0.7} & 75.0\dn{1.1} & 80.0\dn{0.4} & 86.2\up{1.3} & 64.7\up{4.6} \\
\midrule
\multicolumn{10}{l}{\textit{CogRSI}} \\
\libC{} (rebuilt library) & 1.1\dn{2.8} & 50.8\dn{1.0} & 72.2\up{4.6} & 49.3\dn{0.2} & 67.2\up{1.3} & 71.2\dn{4.9} & 78.1\dn{2.3} & 82.7\dn{2.1} & 59.1\dn{0.9} \\
\libA{} (inherited library) & 2.3\dn{1.6} & 50.7\dn{1.1} & \textbf{75.3}\up{7.7} & 51.6\up{2.1} & 64.6\dn{1.3} & 71.3\dn{4.9} & 78.1\dn{2.3} & 80.6\dn{4.3} & 59.3\dn{0.7} \\
\midrule
\multicolumn{10}{l}{\textit{Ours}} \\
\rowcolor{ours} CogRSI v3 (ours) & \textbf{25.4}\up{21.5} & \textbf{87.2}\up{35.4} & \underline{72.8}\up{5.2} & \textbf{63.3}\up{13.8} & \textbf{78.5}\up{12.5} & \textbf{76.8}\up{0.7} & \underline{83.6}\up{3.2} & 85.4\up{0.6} & \textbf{71.6}\up{11.6} \\
\bottomrule
\end{tabular}}
\caption{\textbf{Development sample: held-out sample 1.} Accuracy (\%) per dataset, held-out sample 1, the development set of \sys{}'s arbitration (not a test sample); smoke subsets excluded. Mean over three runs (\sys{} and baselines; GEPA: three optimizer seeds; Stage-1 tools: three loop seeds; \readerL{}: one run per principle-loop seed). FANToM: official All$^*$ on short contexts. SimpleToM: mean of behavior and judgment. ExploreToM: mean of the E1 (physical access) and E2 (communication access) partitions. Subscripts: gap to the floor (the same native solver with an empty tool library). Best in bold, runner-up underlined. Overall averages the datasets every row has (count in the header). Test results: Table~\ref{tab:t1}.}
\label{tab:t1-h1}
\end{table*}

\begin{table*}[t]
\centering
\footnotesize
\setlength{\tabcolsep}{2.4pt}
\resizebox{\textwidth}{!}{%
\begin{tabular}{llcccccccccc}
\toprule
System & Run & FAN & Simple & Simple & Explore & Explore & Hi & Open & TBench & Cog & Big \\
 & & All$^*$ & Behav. & Judg. & E1 & E2 &  &  &  &  &  \\
\midrule
Native solver & 0 & 3.2 & 68.3 & 33.5 & 54.7 & 79.2 & 54.8 & 66.4 & 70.3 & 79.6 & 83.3 \\
 & 1 & 3.7 & 68.3 & 33.5 & 54.7 & 81.2 & 56.0 & 66.8 & 70.1 & 79.6 & 83.5 \\
 & 2 & 4.3 & 68.3 & 33.5 & 52.0 & 81.6 & 56.6 & 66.4 & 69.9 & 79.8 & 82.9 \\
\midrule
Zero-shot & 0 & 7.5 & 75.4 & 47.3 & 51.6 & 77.6 & 59.2 & 68.2 & 72.1 & 80.0 & 85.1 \\
 & 1 & 8.0 & 75.4 & 44.3 & 50.8 & 79.2 & 57.6 & 68.2 & 72.3 & 79.6 & 84.7 \\
 & 2 & 8.6 & 74.9 & 47.3 & 51.6 & 78.4 & 58.4 & 68.8 & 72.1 & 79.4 & 84.9 \\
\midrule
CoT & 0 & 13.9 & 82.6 & 54.5 & 56.3 & 83.6 & 51.8 & 72.0 & 71.7 & 81.4 & 87.9 \\
 & 1 & 13.9 & 82.0 & 55.1 & 54.3 & 85.6 & 53.2 & 72.2 & 71.3 & 83.0 & 88.5 \\
 & 2 & 11.8 & 83.8 & 56.3 & 53.5 & 84.0 & 50.4 & 73.2 & 72.3 & 83.2 & 88.7 \\
\midrule
SC ($k{=}6$) & 0 & 18.2 & 83.2 & 56.3 & 57.5 & 88.4 & 62.4 & 75.0 & 73.5 & 86.6 & 89.1 \\
 & 1 & 16.6 & 83.2 & 59.9 & 56.7 & 87.6 & 62.4 & 74.8 & 74.1 & 85.2 & 89.9 \\
 & 2 & 18.2 & 83.2 & 55.1 & 58.3 & 87.2 & 62.0 & 75.2 & 74.3 & 84.8 & 88.5 \\
\midrule
SC ($k{=}9$) & 0 & 17.6 & 83.8 & 61.1 & 55.5 & 88.4 & 64.3 & 74.4 & 73.1 & 86.4 & 90.3 \\
 & 1 & 19.8 & 83.2 & 55.1 & 56.3 & 87.2 & 65.1 & 74.8 & 73.1 & 86.0 & 88.7 \\
 & 2 & 16.6 & 84.4 & 59.3 & 57.5 & 87.6 & 65.3 & 74.0 & 73.9 & 85.4 & 89.1 \\
\midrule
Retrieval & 0 & 2.1 & 61.7 & 37.1 & 62.2 & 76.8 & 58.8 & 66.0 & 71.1 & 76.6 & 83.1 \\
 & 1 & 1.6 & 61.7 & 35.3 & 62.2 & 75.2 & 58.2 & 66.0 & 70.5 & 76.2 & 82.7 \\
 & 2 & 1.6 & 63.5 & 37.1 & 62.2 & 74.4 & 58.8 & 65.6 & 70.7 & 76.4 & 83.5 \\
\midrule
GEPA & 0 & 0.5 & 61.1 & 28.1 & 63.8 & 76.8 & 45.0 & 65.8 & 67.1 & 75.8 & 81.9 \\
 & 1 & 0.0 & 61.7 & 30.5 & 65.0 & 79.2 & 49.8 & 64.6 & 68.9 & 75.4 & 81.5 \\
 & 2 & 0.5 & 58.1 & 34.7 & 63.8 & 82.4 & 44.8 & 67.4 & 66.5 & 73.2 & 81.5 \\
\midrule
PercepToM & 0 & 1.6 & 94.0 & 81.4 & 65.4 & 69.6 & 50.4 & 67.8 & 71.9 & 79.8 & 84.5 \\
 & 1 & 2.7 & 93.4 & 80.8 & 65.7 & 70.0 & 48.4 & 69.0 & 71.9 & 79.8 & 85.1 \\
 & 2 & 1.6 & 92.8 & 82.6 & 65.0 & 70.8 & 45.8 & 68.4 & 72.3 & 79.6 & 84.7 \\
\midrule
SimToM & 0 & 0.5 & 82.0 & 71.9 & 66.1 & 86.4 & 36.7 & 73.2 & 68.1 & 76.8 & 85.1 \\
 & 1 & 1.1 & 83.2 & 71.3 & 63.4 & 84.4 & 38.2 & 74.8 & 66.5 & 75.4 & 85.5 \\
 & 2 & 1.6 & 85.6 & 74.3 & 64.2 & 86.8 & 37.8 & 74.2 & 68.9 & 74.8 & 84.5 \\
\midrule
\libC{} (rebuilt library) & 0 & 2.7 & 53.9 & 44.3 & 67.3 & 86.4 & 56.4 & 71.8 & 69.9 & 76.2 & 82.3 \\
 & 1 & 3.7 & 63.5 & 45.5 & 52.8 & 81.2 & 56.2 & 68.0 & 69.1 & 76.4 & 81.0 \\
 & 2 & 1.6 & 44.9 & 58.7 & 60.2 & 85.2 & 49.0 & 60.0 & 65.9 & 64.4 & 79.4 \\
\midrule
\libA{} (inherited library) & 0 & 1.6 & 67.1 & 57.5 & 74.4 & 87.2 & 55.2 & 68.4 & 68.3 & 76.2 & 82.7 \\
 & 1 & 1.6 & 49.1 & 43.7 & 64.2 & 83.2 & 56.4 & 64.4 & 67.9 & 74.6 & 79.4 \\
 & 2 & 3.7 & 58.1 & 52.1 & 57.5 & 77.2 & 47.6 & 64.8 & 70.3 & 76.0 & 79.0 \\
\midrule
\reader{} (vote only) & 0 & 25.1 & 92.8 & 88.0 & 59.1 & 83.6 & 67.1 & 77.4 & 71.9 & 84.8 & 86.7 \\
 & 1 & 29.4 & 92.2 & 85.6 & 59.8 & 84.8 & 66.3 & 77.4 & 71.7 & 84.8 & 86.3 \\
 & 2 & 31.6 & 92.2 & 88.0 & 57.1 & 84.0 & 68.1 & 77.6 & 72.3 & 84.2 & 86.9 \\
\midrule
\reader{} (no close-split) & 0 & 27.8 & 93.4 & 84.4 & 57.5 & 84.4 & 71.7 & 75.8 & 72.1 & 83.0 & 89.7 \\
 & 1 & 26.7 & 92.8 & 85.0 & 57.5 & 85.6 & 70.5 & 76.6 & 72.5 & 84.2 & 90.1 \\
 & 2 & 27.3 & 92.2 & 83.8 & 58.7 & 86.0 & 70.1 & 76.4 & 73.3 & 83.8 & 90.3 \\
\midrule
\reader{} & 0 & 31.0 & 95.8 & 83.8 & 59.1 & 85.6 & 69.7 & 76.2 & 74.3 & 85.2 & 90.3 \\
 & 1 & 24.6 & 95.2 & 85.0 & 58.3 & 88.4 & 71.5 & 75.4 & 74.1 & 85.0 & 89.9 \\
 & 2 & 28.3 & 95.2 & 85.0 & 57.9 & 86.0 & 66.7 & 75.8 & 74.1 & 84.4 & 90.9 \\
\midrule
\readerL{} (loop-written principle) & 0 & 24.1 & 86.8 & 66.5 & 57.1 & 89.2 & 72.5 & 76.0 & 70.7 & 84.6 & 89.9 \\
 & 1 & 25.1 & 90.4 & 81.4 & 59.8 & 86.4 & 71.1 & 76.0 & 72.7 & 85.6 & 90.3 \\
 & 2 & 26.2 & 91.6 & 78.4 & 64.6 & 86.8 & 67.9 & 75.2 & 73.9 & 85.0 & 88.3 \\
 & 3 & 24.6 & 90.4 & 76.6 & 58.7 & 88.0 & 70.1 & 79.6 & 73.3 & 85.2 & 89.5 \\
 & 4 & 25.7 & 89.8 & 68.9 & 59.1 & 86.8 & 70.1 & 75.6 & 73.5 & 84.4 & 89.7 \\
 & 5 & 26.2 & 83.8 & 76.0 & 61.4 & 90.4 & 68.3 & 76.4 & 72.9 & 83.4 & 89.7 \\
\bottomrule
\end{tabular}}
\caption{Every run behind the main test sample (Table~\ref{tab:t1}) (accuracy \%; FANToM: All$^*$ over 187 target sets). Runs of static baselines are independent executions; GEPA runs are optimizer seeds.}
\label{tab:runs-h2}
\end{table*}

\begin{table*}[t]
\centering
\small
\setlength{\tabcolsep}{5pt}
\begin{tabular}{lcccccc}
\toprule
& \multicolumn{2}{c}{\textbf{FANToM}} & \multicolumn{2}{c}{\textbf{SimpleToM}} & \multicolumn{2}{c}{\textbf{ExploreToM}} \\
\cmidrule(lr){2-3}\cmidrule(lr){4-5}\cmidrule(l){6-7}
\textbf{Method} & All$^*$ & Belief & Behav. & Judg. & E1 & E2 \\
\midrule
Native solver & 4.3 & 52.9 & 65.3 & 38.6 & 50.0 & 72.5 \\
Zero-shot & 7.8\up{3.5} & 63.5\up{10.5} & 72.0\up{6.7} & 49.8\up{11.2} & 45.8\dn{4.2} & 74.3\up{1.8} \\
CoT & \underline{14.4}\up{10.1} & 68.2\up{15.2} & 78.0\up{12.7} & \underline{58.9}\up{20.3} & 44.4\dn{5.6} & 79.5\up{7.0} \\
SC ($k{=}6$) & \textbf{18.2}\up{14.0} & \underline{71.2}\up{18.3} & \underline{78.5}\up{13.2} & 58.3\up{19.8} & 47.2\dn{2.9} & \textbf{82.3}\up{9.8} \\
Retrieval & 2.8\dn{1.5} & 41.7\dn{11.2} & 61.4\dn{3.9} & 38.1\dn{0.5} & 56.7\up{6.7} & 73.7\up{1.2} \\
GEPA & 0.6\dn{3.7} & 38.8\dn{14.2} & 58.8\dn{6.5} & 34.6\dn{4.0} & 60.2\up{10.2} & 77.1\up{4.6} \\
PercepToM & 5.0$^\ddagger$\up{0.7} & \textbf{81.0}\up{28.0} & \textbf{80.1}\up{14.9} & \textbf{69.0}\up{30.5} & \underline{61.1}\up{11.1} & 64.7\dn{7.8} \\
\libC{} (rebuilt library) & 2.2\dn{2.1} & 50.3\dn{2.6} & 54.8\dn{10.5} & 48.4\up{9.8} & 56.8\up{6.7} & \underline{80.7}\up{8.2} \\
\libA{} (inherited library) & 2.9\dn{1.4} & 48.3\dn{4.7} & 54.7\dn{10.6} & 50.6\up{12.1} & \textbf{61.8}\up{11.8} & 79.5\up{7.0} \\
\bottomrule
\end{tabular}
\caption{Per-metric detail for FANToM, SimpleToM and ExploreToM (same runs as Table~\ref{tab:sealed}). Subscripts: gap to the floor. $^\dagger$Fewer than three runs. $^\ddagger$PercepToM on FANToM is scored from its per-question results like every other system; the official scorer's printed summary for these runs is higher and cannot be reproduced from the per-question results it writes (for every other system the two agree exactly). }
\label{tab:sealed-detail}
\end{table*}

\begin{table*}[t]
\centering
\small
\setlength{\tabcolsep}{4pt}
\begin{tabular}{lccccccc}
\toprule
& \multicolumn{2}{c}{\textbf{FANToM}} & \multicolumn{3}{c}{\textbf{SimpleToM}} & \multicolumn{2}{c}{\textbf{ExploreToM}} \\
\cmidrule(lr){2-3}\cmidrule(lr){4-6}\cmidrule(l){7-8}
\textbf{System} & All$^*$ & Belief & Behav. & Judg. & Mental & E1 & E2 \\
\midrule
Native solver & 4.3 & 52.9 & 65.3 & 38.6 & 97.2 & 50.0 & 72.5 \\
Zero-shot & 7.8 & 63.5 & 72.0 & 49.8 & 95.0 & 45.8 & 74.3 \\
CoT & 14.4 & 68.2 & 78.0 & 58.9 & 98.0 & 44.4 & 79.5 \\
SC ($k{=}6$) & 18.2 & 71.2 & 78.5 & 58.3 & 98.1 & 47.2 & 82.3 \\
Retrieval & 2.8 & 41.7 & 61.4 & 38.1 & 89.4 & 56.7 & 73.7 \\
GEPA & 0.6 & 38.8 & 58.8 & 34.6 & 84.3 & 60.2 & 77.1 \\
PercepToM & 5.0$^\ddagger$ & 81.0 & 80.1 & 69.0 & 70.9 & 61.1 & 64.7 \\
PercepToM (v4 adapter, appendix) & 4.7 & 81.6 & 91.6 & 77.4 & 62.0 & 61.1 & 65.9 \\
E2 C+ (library visible to builder) & 1.7 & 44.9 & 55.2 & 47.8 & 78.2 & 60.6 & 81.8 \\
E2 C$-$ (library hidden from builder) & 1.9 & 40.5 & 56.0 & 50.5 & 79.8 & 60.5 & 83.6 \\
\libC{} (rebuilt library) & 2.2 & 50.3 & 54.8 & 48.4 & 88.2 & 56.8 & 80.7 \\
\libA{} (inherited library) & 2.9 & 48.3 & 54.7 & 50.6 & 80.3 & 61.8 & 79.5 \\
\bottomrule
\end{tabular}
\caption{All systems, including rows kept out of Table~\ref{tab:sealed}: the original PercepToM adapter (v4) and the E2 branches (parent library plus new tools). Mental-state questions all have gold answer ``No'', so a constant ``No'' scores 100\%; this column is reported separately. $^\dagger$Fewer than three runs. $^\ddagger$PercepToM on FANToM is scored from its per-question results like every other system; the official scorer's printed summary for these runs is higher and cannot be reproduced from the per-question results it writes (for every other system the two agree exactly). }
\label{tab:sealed-all}
\end{table*}

\begin{table*}[t]
\centering
\small
\setlength{\tabcolsep}{5pt}
\begin{tabular}{lccccc}
\toprule
Method & FANToM All$^*$ & SimpleToM & Hi-ToM & BigToM & Overall \\
\midrule
\rowcolor{card}\multicolumn{6}{l}{\textit{Qwen3.5-122B-A10B (FP8), larger-model spot check}} \\
Zero-shot & 3.0 & 70.0 & 68.8 & 85.9 & 56.9 \\
CoT & 11.6 & 78.5 & 55.6 & 89.0 & 58.7 \\
SC ($k{=}6$) & 18.2 & 80.8 & 68.0 & \textbf{89.9} & 64.2 \\
PercepToM & 6.2 & 88.2 & 48.0 & 86.2 & 57.2 \\
\rowcolor{ours}\reader{} & \textbf{25.3} & \textbf{94.0} & \textbf{72.4} & 89.7 & \textbf{70.4} \\
\midrule
\rowcolor{card}\multicolumn{6}{l}{\textit{Llama-3.3-70B-Instruct (AWQ 4-bit)}} \\
Zero-shot & 0.0 & 72.5 & 65.9 & 87.7 & 56.5 \\
CoT & 1.6 & 86.0 & 65.5 & 88.0 & 60.3 \\
SC ($k{=}6$) & 3.7 & 87.1 & 68.9 & 88.0 & 61.9 \\
PercepToM & 1.1 & 85.0 & 59.1 & 87.1 & 58.1 \\
\rowcolor{ours}\reader{} & \textbf{12.8} & \textbf{96.6} & \textbf{69.7} & \textbf{89.4} & \textbf{67.1} \\
\midrule
\rowcolor{card}\multicolumn{6}{l}{\textit{GLM-4.6V-Flash}} \\
Zero-shot & 0.9 & 45.8 & 39.5 & 75.3 & 40.4 \\
CoT & 0.0 & 50.5 & 57.4 & 80.0 & 47.0 \\
SC ($k{=}6$) & 0.9 & 49.8 & 61.0 & 80.2 & 48.0 \\
PercepToM & 1.1 & \textbf{72.5} & 29.2 & 74.3 & 44.3 \\
\rowcolor{ours}\reader{} & \textbf{2.9} & 71.1 & \textbf{61.5} & \textbf{84.1} & \textbf{54.9} \\
\midrule
\rowcolor{card}\multicolumn{6}{l}{\textit{Qwen3.5-9B}} \\
Zero-shot & 2.5 & 37.2 & 48.6 & 80.9 & 42.3 \\
CoT & 4.6 & 50.0 & 53.4 & 88.0 & 49.0 \\
SC ($k{=}6$) & 6.1 & 48.3 & 61.9 & 89.2 & 51.4 \\
PercepToM & 0.2 & 65.3 & 38.8 & 81.2 & 46.4 \\
\rowcolor{ours}\reader{} & \textbf{11.4} & \textbf{76.6} & \textbf{65.1} & \textbf{90.3} & \textbf{60.9} \\
\bottomrule
\end{tabular}
\caption{Other backbones. Same systems, prompts and test samples as Table~\ref{tab:t1}; mean of three runs. SimpleToM: mean of behavior and judgment. Best per backbone in bold. $^\dagger$Fewer than three runs finished. On FANToM, \sys{}'s close-split arbitration does not apply (dialogue questions), so its FANToM cells reuse the runs of the variant without it (Appendix~\ref{app:variants}).}
\label{tab:backbones}
\end{table*}

\begin{table*}[t]
\centering
\scriptsize
\setlength{\tabcolsep}{4pt}
\begin{tabular}{llccccccc}
\toprule
\textbf{System} & \textbf{Run} & All$^*$ & Belief & Behav. & Judg. & Mental & E1 & E2 \\
\midrule
Native solver & rep. 0 & 4.1 & 53.5 & 65.3 & 38.6 & 97.2 & 50.1 & 72.6 \\
 & rep. 1 & 4.5 & 52.5 & 65.3 & 38.6 & 97.2 & 50.3 & 72.3 \\
 & rep. 2 & 4.3 & 52.7 & 65.3 & 38.6 & 97.2 & 49.6 & 72.7 \\
\midrule
Zero-shot & rep. 0 & 7.5 & 63.1 & 71.7 & 49.3 & 95.0 & 46.0 & 74.3 \\
 & rep. 1 & 8.1 & 63.8 & 72.3 & 49.9 & 95.1 & 45.7 & 74.4 \\
 & rep. 2 & 7.9 & 63.5 & 71.8 & 50.2 & 94.8 & 45.8 & 74.3 \\
\midrule
CoT & rep. 0 & 14.6 & 67.8 & 78.9 & 59.4 & 98.1 & 44.1 & 79.7 \\
 & rep. 1 & 14.9 & 68.6 & 77.7 & 58.5 & 98.1 & 44.4 & 79.5 \\
 & rep. 2 & 13.6 & 68.1 & 77.4 & 58.7 & 97.8 & 44.7 & 79.3 \\
\midrule
SC ($k{=}6$) & rep. 0 & 17.2 & 70.7 & 78.6 & 58.0 & 98.3 & 47.0 & 82.5 \\
 & rep. 1 & 18.4 & 71.4 & 78.6 & 59.9 & 97.9 & 47.5 & 82.5 \\
 & rep. 2 & 19.2 & 71.6 & 78.2 & 57.1 & 98.1 & 46.9 & 82.0 \\
\midrule
Retrieval & rep. 0 & 2.7 & 42.2 & 62.0 & 38.5 & 89.9 & 56.8 & 73.5 \\
 & rep. 1 & 2.6 & 40.8 & 60.5 & 37.8 & 88.9 & 55.9 & 73.8 \\
 & rep. 2 & 3.1 & 42.1 & 61.6 & 37.9 & 89.5 & 57.4 & 73.9 \\
\midrule
GEPA & opt. seed 0 & 0.5 & 36.6 & 63.4 & 34.3 & 88.7 & 60.6 & 76.2 \\
 & opt. seed 1 & 0.0 & 38.0 & 57.6 & 32.4 & 86.8 & 63.1 & 76.9 \\
 & opt. seed 2 & 1.2 & 41.6 & 55.3 & 37.2 & 77.3 & 57.0 & 78.3 \\
\midrule
PercepToM & rep. 0 & 4.8 & 80.4 & 80.4 & 69.2 & 70.8 & 61.0 & 64.6 \\
 & rep. 1 & 5.3 & 81.5 & 80.1 & 68.9 & 70.9 & 61.3 & 64.7 \\
 & rep. 2 & 4.8 & 80.9 & 79.9 & 69.0 & 70.9 & 61.1 & 64.8 \\
\midrule
PercepToM (v4 adapter, appendix) & rep. 0 & 4.6 & 81.9 & 92.0 & 77.5 & 62.1 & 61.7 & 66.2 \\
 & rep. 1 & 5.7 & 80.9 & 91.5 & 77.6 & 61.8 & 61.1 & 65.8 \\
 & rep. 2 & 3.9 & 82.2 & 91.4 & 77.2 & 62.1 & 60.6 & 65.7 \\
\midrule
E2 C+ (library visible to builder) & seed 0 & 1.9 & 30.8 & 63.3 & 53.4 & 75.6 & 74.4 & 86.2 \\
 & seed 1 & 1.9 & 41.2 & 46.1 & 45.1 & 80.6 & 56.3 & 82.3 \\
 & seed 2 & 1.4 & 62.9 & 56.2 & 44.9 & 78.5 & 51.0 & 77.0 \\
\midrule
E2 C$-$ (library hidden from builder) & seed 0 & 1.7 & 31.6 & 62.6 & 53.8 & 75.5 & 75.3 & 86.4 \\
 & seed 1 & 1.4 & 40.0 & 46.9 & 49.5 & 79.2 & 57.6 & 83.5 \\
 & seed 2 & 2.7 & 50.0 & 58.6 & 48.2 & 84.6 & 48.6 & 81.0 \\
\midrule
\libC{} (rebuilt library) & seed 0 & 2.2 & 51.1 & 54.1 & 44.4 & 93.3 & 64.2 & 84.0 \\
 & seed 1 & 3.8 & 58.5 & 60.6 & 44.7 & 84.0 & 51.6 & 77.0 \\
 & seed 2 & 0.7 & 41.3 & 49.6 & 56.1 & 87.4 & 54.5 & 81.2 \\
\midrule
\libA{} (inherited library) & seed 0 & 1.5 & 32.1 & 63.0 & 55.9 & 73.7 & 75.1 & 84.9 \\
 & seed 1 & 2.1 & 43.8 & 46.6 & 43.7 & 79.4 & 58.0 & 78.3 \\
 & seed 2 & 5.0 & 68.9 & 54.5 & 52.3 & 87.8 & 52.5 & 75.2 \\
\bottomrule
\end{tabular}
\caption{Every run behind Tables~\ref{tab:sealed}--\ref{tab:sealed-all} (accuracy \%). Repeats of static baselines are independent executions, not controlled sampling seeds. -- = not run.}
\label{tab:sealed-runs}
\end{table*}

\subsection{Ablation study of the reader}
\label{app:rq2}
\paragraph{What each part adds.} The rows of Table~\ref{tab:ablation} compare whole systems. No Stage-1 output enters
Stage~2 or~3: Stage~1 only motivated the access annotation. \libA{} scores \AblRSIAMain{} Main because it lacks the whole reader. Inside the same reader,
the hand-written principle adds \MacroHandEmpty{} Main points over no principle (95\% CI \MacroHandEmptyCI{}), the loop's
principle adds \MacroLoopEmpty{} (\MacroLoopEmptyCI{}), and the reader with no principle already beats SC ($k{=}9$) by
\MacroEmptySCnine{} (\MacroEmptySCnineCI{}). In Stage~3 each reading and each arbitration step lifts belief accuracy,
within the bound of Proposition~\ref{prop:local}. On FANToM All$^*$ the differences between Stage-3 variants are within
the spread of three runs (Table~\ref{tab:runs-h2}).

\paragraph{Annotating and filtering are close.} With the same arbitration, PercepToM's filtered reading reaches
\AFMainPArb{} on the main multiple-choice benchmarks and \reader{}, which keeps the whole story, reaches \AFMainVsix{}.
The two also differ in sample allocation, so this small gap does not isolate annotation from filtering. With our six
samples PercepToM + SC and SimToM + SC reach \OvPercepToMSCsix{} and \OvSimToMSC{} Main, and on FANToM All$^*$
\FanPercepToMSCsix{} and \FanSimToMSC{}, against \FanVsix{}.

Table~\ref{tab:annfilter} gives PercepToM's filtered reading the same arbitration as \reader{}: where its answer and
self-consistency's differ, the evidential rule decides. On \ArbDevConf{} development conflicts the evidential rule is
right \ArbDevEvid{} times, the plain view \ArbDevPlain{}, the annotated view \ArbDevAnn{} and a neutral arbiter
\ArbDevNeutral{}; without the rule the model assumes characters noticed hidden facts (SimpleToM: \ArbDevSimpleNeutral{}
vs.\ \ArbDevSimpleEvid{} of \ArbDevSimpleConf{}; Appendix~\ref{app:rq8}).

\subsection{RSI over procedures}
\begin{table*}[t]
\centering
\small
\setlength{\tabcolsep}{4pt}
\resizebox{\textwidth}{!}{%
\begin{tabular}{ll rcccccc}
\toprule
 & Construct & $n$ & SC & PercepToM & \libA{} & \sys{} & \sys{}$-$SC [95\% CI] & within benchmark \\
\midrule
\textit{Epistemic} & Belief & 3438 & 73.8 & 71.8 & 63.6 & \textbf{77.7} & \textbf{+3.9 [+2.5, +5.4]} & \textbf{+1.2 [+0.3, +2.1]} \\
 & Knowledge & 3095 & 84.6 & 65.4 & 77.3 & \textbf{85.7} & \textbf{+1.1 [+0.2, +2.0]} & $-$0.8 [$-$1.8, +0.1] \\
\textit{Motivational} & Desire & 72 & \textbf{81.5} & 79.2 & 81.0 & 80.1 & $-$1.4 [$-$3.7, +0.5] & $-$1.1 [$-$3.5, +0.9] \\
 & Intention & 118 & 88.1 & \textbf{89.0} & 85.9 & 88.1 & 0.0 [$-$1.7, +1.8] & +0.2 [$-$1.4, +2.0] \\
\textit{Affective} & Emotion & 151 & \textbf{78.1} & 76.2 & 75.3 & 75.7 & $-$2.4 [$-$5.4, +0.2] & $-$2.4 [$-$5.4, +0.2] \\
\textit{Pragmatic} & Non-literal & 177 & 79.8 & \textbf{82.7} & 76.3 & 78.2 & $-$1.7 [$-$3.7, +0.2] & \textbf{$-$1.7 [$-$3.5, $-$0.1]} \\
\textit{Mixed} & Mixed & 97 & \textbf{79.7} & 70.8 & 68.7 & 74.9 & \textbf{$-$4.8 [$-$9.5, $-$0.4]} & \textbf{$-$4.1 [$-$8.4, $-$0.6]} \\
\bottomrule
\end{tabular}}
\caption{Accuracy by ATOMS construct \citep{beaudoin2020atoms} on all test questions of the main backbone (main test sample, three runs each). The difference to self-consistency is bootstrapped over stories/conversations (2{,}000 resamples); \emph{within benchmark} centres each question's difference on its benchmark's mean difference and resamples stories within benchmarks. Bold intervals exclude zero. Constructs with fewer than 50 questions (perception, other) are omitted.}
\label{tab:constructs}
\end{table*}

\begin{table*}[t]
\centering
\small
\setlength{\tabcolsep}{4pt}
\resizebox{\textwidth}{!}{%
\begin{tabular}{llccc}
\toprule
Benchmark & Strongest baseline & \sys{} $-$ best & Holm & \sys{} $-$ SC \\
\midrule
FANToM All$^*$ & PercepToM + SC (belief) (23.7) & \textbf{+4.3 [+0.2, +8.6]} & -- & \textbf{+10.3 [+5.2, +15.6]} \\
SimpleToM & PercepToM (87.5) & +2.5 [$-$0.8, +6.0] & -- & \textbf{+19.9 [+16.1, +23.7]} \\
ExploreToM & SimToM + SC (76.9) & \textbf{$-$4.3 [$-$8.3, $-$0.6]} & -- & $-$0.1 [$-$2.1, +1.9] \\
Hi-ToM & SC ($k{=}9$) (64.9) & \textbf{+4.4 [+1.2, +7.6]} & -- & \textbf{+7.1 [+4.2, +10.0]} \\
OpenToM & SimToM + SC (76.5) & $-$0.7 [$-$3.0, +1.5] & -- & +0.8 [$-$0.9, +2.7] \\
ToMBench & PercepToM + SC (74.3) & $-$0.1 [$-$1.6, +1.3] & -- & +0.2 [$-$1.1, +1.5] \\
CogToM & SC ($k{=}9$) (85.9) & $-$1.1 [$-$2.3, +0.1] & -- & $-$0.7 [$-$1.9, +0.5] \\
BigToM & PercepToM + SC (89.5) & +0.9 [$-$0.4, +2.2] & -- & +1.2 [$-$0.1, +2.5] \\
\bottomrule
\end{tabular}}
\caption{Paired differences (points) between \sys{} and the strongest baseline of each benchmark (its score in brackets) and self-consistency, main backbone, three runs per system. 95\% intervals from 2{,}000 bootstrap resamples of stories or conversations; bold intervals exclude zero. Strongest baseline: the highest Table~\ref{tab:t1} score among all baselines, including SimToM + SC and PercepToM + SC. Holm: significant after Holm correction over the eight benchmarks (bootstrap $p$-values, familywise 0.05). An interval that includes zero is not evidence of equivalence.}
\label{tab:paired}
\end{table*}

\label{app:rq3}
On Percept-ToMi's gold perceivers \citep{jung2024perceptom}, \libA{}'s tools export at most \InheritCovMax{}\% of the gold
access pairs; rebuilt tools omit perceivers (precision \ToolOmitP{}, recall \ToolOmitR{}) or credit characters in other
rooms (\ToolOverP{} / \ToolOverR{}). Inheriting the library minus rebuilding it is \EOneDiff{} points of FANToM All$^*$
(95\% CI \EOneCI{} under the pre-registered hierarchical bootstrap over runs and conversations, and \RInhciAC{} under the
seed-paired bootstrap used for the matched arm in Appendix~\ref{app:vc}); acquiring communication with the inherited library in reach rather than without it gives \ETwoDiff{}
(\ETwoCI{}). The inheriting runs admitted \AdmitA{} tools and had a non-empty library in \ALibRounds{} of
\ARoundsTotal{} rounds; the rebuilding runs admitted \AdmitC{} (Table~\ref{tab:rsi-runs}).

\subsection{RSI over principles}
\label{app:rq4}
\begin{figure*}[t]
\centering
\input{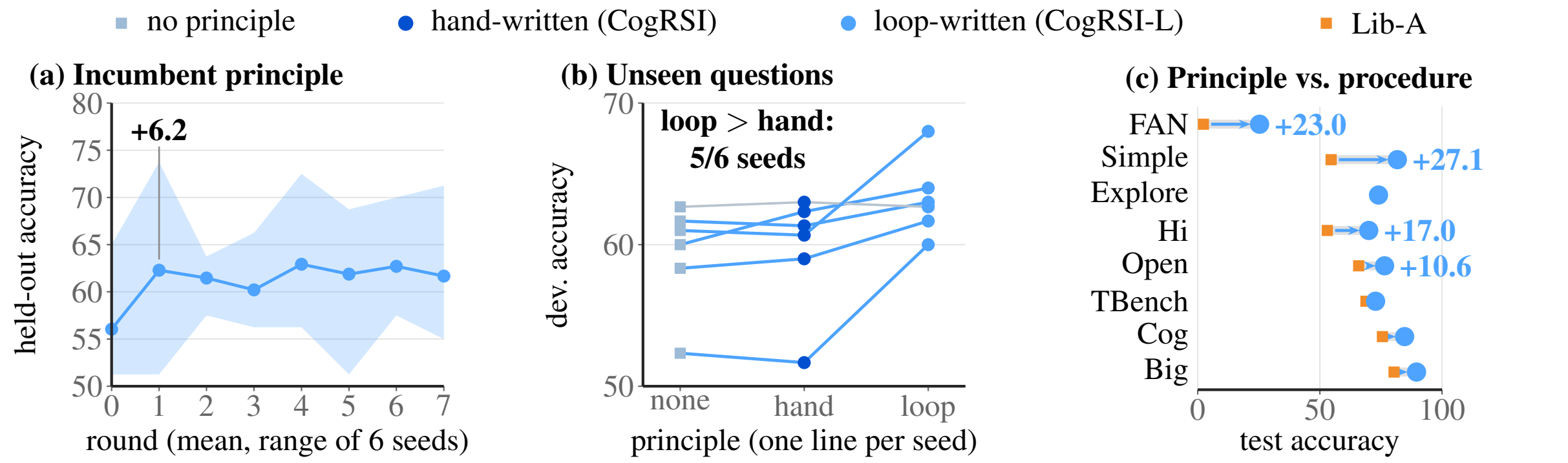}
\caption{\textbf{RSI over principles.} (a) Incumbent's accuracy on new development questions by round (mean and range over \RsipNumSeeds{} seeds; round 0 admits the first principle). (b) \RsipFinalN{} new development questions on seen stories, per seed (no, hand-written, loop-written principle). (c) Test accuracy per benchmark, \libA{} $\to$ \readerL{}, main test sample. \textbf{The loop's rule beats ours on development questions in most seeds, not on test.}}
\label{fig:rsip}
\end{figure*}
\paragraph{Code and text differ in more than the artifact.} \readerL{} scores \OvRsip{} Main and \FanRsip{} FANToM
All$^*$, against \OvRSIA{} and \RFanRSIA{} for \libA{} (Figure~\ref{fig:rsip}c), but they also differ in
perception, readings, voting and arbitration.
It beats \reader{} on ExploreToM (\RExploreRsip{} vs.\ \RExploreVsix{}), Hi-ToM (\RHiToMRsip{} vs.\ \RHiToMVsix{}) and
OpenToM (\ROpenToMRsip{} vs.\ \ROpenToMVsix{}), and trails it on FANToM All$^*$ and SimpleToM
(Table~\ref{tab:t1}).

Table~\ref{tab:rsip} gives every seed: admitted rounds, the final development cohort and the main test sample; near its
source \readerL{} leads \reader{} on ExploreToM (\RExploreRsip{} vs.\ \RExploreVsix{}) and trails it on FANToM All$^*$
(\FanRsip{} vs.\ \FanVsix{}) and SimpleToM (\RSimpleRsip{} vs.\ \RSimpleVsix{}). Trial accuracy rises by
construction, since a round is admitted only if it does. On a development cohort of new questions, scored after every round, the
incumbent goes from \CurveZero{} (no principle) to \CurveOne{} after round~0 and stays between \CurveLaterLo{} and
\CurveLaterHi{} after that. The final principles score \CurveFinalLoop{}, against \CurveFinalHand{} for ours and
\CurveFinalEmpty{} for none. On the main test sample the reader with no principle scores \OvEmpty{} Main (three runs,
highest \EmptyMaxMain{}), below the lowest \readerL{} seed (\RsipMinMain{}).
\subsection{Grounded reading}
\label{app:rq5}
\begin{figure*}[t]
\centering
\centering \begin{tikzpicture}[font=\figfont, x=1cm, y=1cm]
\fill[cSCnine] (-7.435,-0.100) rectangle (-7.155,0.100);\fill[pattern=north east lines, pattern color=white] (-7.435,-0.100) rectangle (-7.155,0.100);\draw[black!40, line width=0.2pt] (-7.43,-0.10) rectangle (-7.15,0.10);
\node[anchor=west, inner sep=1pt] at (-7.08,-0.00) {SC ($k{=}9$)};
\fill[cPercep] (-5.245,-0.100) rectangle (-4.965,0.100);\fill[pattern=dots, pattern color=white] (-5.245,-0.100) rectangle (-4.965,0.100);\draw[black!40, line width=0.2pt] (-5.25,-0.10) rectangle (-4.96,0.10);
\node[anchor=west, inner sep=1pt] at (-4.89,-0.00) {PercepToM};
\fill[cProc] (-2.875,-0.100) rectangle (-2.595,0.100);\draw[black!40, line width=0.2pt] (-2.88,-0.10) rectangle (-2.60,0.10);
\node[anchor=west, inner sep=1pt] at (-2.53,-0.00) {\libA{}};
\fill[cAnn] (-1.225,-0.100) rectangle (-0.945,0.100);\fill[pattern=dots, pattern color=white] (-1.225,-0.100) rectangle (-0.945,0.100);\draw[black!40, line width=0.2pt] (-1.23,-0.10) rectangle (-0.95,0.10);
\node[anchor=west, inner sep=1pt] at (-0.88,-0.00) {annotated};
\fill[cPlain] (1.145,-0.100) rectangle (1.425,0.100);\fill[pattern=north east lines, pattern color=white] (1.145,-0.100) rectangle (1.425,0.100);\draw[black!40, line width=0.2pt] (1.14,-0.10) rectangle (1.42,0.10);
\node[anchor=west, inner sep=1pt] at (1.49,-0.00) {plain};
\fill[cSCnine] (2.795,-0.100) rectangle (3.075,0.100);\fill[pattern=north east lines, pattern color=white] (2.795,-0.100) rectangle (3.075,0.100);\draw[black!40, line width=0.2pt] (2.79,-0.10) rectangle (3.07,0.10);
\node[anchor=west, inner sep=1pt] at (3.14,-0.00) {neutral arb.};
\fill[cOurs] (5.705,-0.100) rectangle (5.985,0.100);\draw[black!40, line width=0.2pt] (5.70,-0.10) rectangle (5.98,0.10);
\node[anchor=west, inner sep=1pt] at (6.05,-0.00) {\reader{}};
\end{tikzpicture}\\[3pt]
\begin{minipage}[b]{0.320\linewidth}\centering\fitbox{\begin{tikzpicture}[font=\figfont, x=1cm, y=1cm]
\draw[black!10] (0,0.000) -- (3.900,0.000);
\draw[black!60] (0,0.000) -- (-0.05,0.000) node[left, inner sep=1pt] {50};
\draw[black!10] (0,1.100) -- (3.900,1.100);
\draw[black!60] (0,1.100) -- (-0.05,1.100) node[left, inner sep=1pt] {70};
\draw[black!10] (0,2.200) -- (3.900,2.200);
\draw[black!60] (0,2.200) -- (-0.05,2.200) node[left, inner sep=1pt] {90};
\node[rotate=90, anchor=south] at (-0.6,1.100) {accuracy};
\node[anchor=south west, font=\figfont\bfseries, inner sep=0pt] at (-0.85,2.320) {(a) Epistemic questions};
\fill[cSCnine] (0.146,0.000) rectangle (0.549,1.328);\fill[pattern=north east lines, pattern color=white] (0.146,0.000) rectangle (0.549,1.328);
\fill[cPercep] (0.561,0.000) rectangle (0.963,1.200);\fill[pattern=dots, pattern color=white] (0.561,0.000) rectangle (0.963,1.200);
\fill[cProc] (0.975,0.000) rectangle (1.377,0.748);
\fill[cOurs] (1.389,0.000) rectangle (1.792,1.526);
\node[anchor=north, inner sep=1.5pt, align=center] at (0.975,-0.02) {Belief};
\fill[cSCnine] (2.096,0.000) rectangle (2.499,1.910);\fill[pattern=north east lines, pattern color=white] (2.096,0.000) rectangle (2.499,1.910);
\fill[cPercep] (2.511,0.000) rectangle (2.913,0.845);\fill[pattern=dots, pattern color=white] (2.511,0.000) rectangle (2.913,0.845);
\fill[cProc] (2.925,0.000) rectangle (3.327,1.499);
\fill[cOurs] (3.339,0.000) rectangle (3.742,1.962);
\node[anchor=north, inner sep=1.5pt, align=center] at (2.925,-0.02) {Knowledge};
\node[anchor=south, font=\figfont\bfseries, text=black, inner sep=1pt, align=center] at (1.591,1.546) {+3.6};
\node[anchor=south, font=\figfont\bfseries, text=black, inner sep=1pt, align=center] at (3.541,1.982) {+0.9};
\draw[black!85, line width=0.8pt] (0,0) -- (0,2.200); \draw[black!85, line width=0.8pt] (0,0) -- (3.900,0);
\path (0,-0.45) (0,2.70);
\end{tikzpicture}}\end{minipage}\hfill
\begin{minipage}[b]{0.320\linewidth}\centering\fitbox{\begin{tikzpicture}[font=\figfont, x=1cm, y=1cm]
\fill[cOurs!20] (0.000,1.650) rectangle (0.967,2.200);
\draw[white, line width=0.8pt] (0.000,1.650) rectangle (0.967,2.200);
\node[text=black] at (0.483,1.925) {78.4};
\fill[cOurs!8] (0.000,1.100) rectangle (0.967,1.650);
\draw[white, line width=0.8pt] (0.000,1.100) rectangle (0.967,1.650);
\node[text=black] at (0.483,1.375) {76.9};
\fill[cOurs!40] (0.000,0.550) rectangle (0.967,1.100);
\draw[white, line width=0.8pt] (0.000,0.550) rectangle (0.967,1.100);
\node[text=black] at (0.483,0.825) {80.8};
\fill[cOurs!51] (0.000,0.000) rectangle (0.967,0.550);
\draw[white, line width=0.8pt] (0.000,0.000) rectangle (0.967,0.550);
\node[text=white] at (0.483,0.275) {\textbf{82.1}};
\fill[cOurs!34] (0.967,1.650) rectangle (1.933,2.200);
\draw[white, line width=0.8pt] (0.967,1.650) rectangle (1.933,2.200);
\node[text=black] at (1.450,1.925) {80.1};
\fill[cOurs!57] (0.967,1.100) rectangle (1.933,1.650);
\draw[white, line width=0.8pt] (0.967,1.100) rectangle (1.933,1.650);
\node[text=white] at (1.450,1.375) {82.8};
\fill[cOurs!60] (0.967,0.550) rectangle (1.933,1.100);
\draw[white, line width=0.8pt] (0.967,0.550) rectangle (1.933,1.100);
\node[text=white] at (1.450,0.825) {83.2};
\fill[cOurs!80] (0.967,0.000) rectangle (1.933,0.550);
\draw[white, line width=0.8pt] (0.967,0.000) rectangle (1.933,0.550);
\node[text=white] at (1.450,0.275) {\textbf{85.5}};
\fill[cOurs!18] (1.933,1.650) rectangle (2.900,2.200);
\draw[white, line width=0.8pt] (1.933,1.650) rectangle (2.900,2.200);
\node[text=black] at (2.417,1.925) {78.1};
\fill[cOurs!22] (1.933,1.100) rectangle (2.900,1.650);
\draw[white, line width=0.8pt] (1.933,1.100) rectangle (2.900,1.650);
\node[text=black] at (2.417,1.375) {78.6};
\fill[cOurs!39] (1.933,0.550) rectangle (2.900,1.100);
\draw[white, line width=0.8pt] (1.933,0.550) rectangle (2.900,1.100);
\node[text=black] at (2.417,0.825) {80.7};
\fill[cOurs!51] (1.933,0.000) rectangle (2.900,0.550);
\draw[white, line width=0.8pt] (1.933,0.000) rectangle (2.900,0.550);
\node[text=white] at (2.417,0.275) {\textbf{82.1}};
\node[anchor=east, inner sep=1pt] at (-0.05,1.925) {annotated};
\node[anchor=east, inner sep=1pt] at (-0.05,1.375) {plain};
\node[anchor=east, inner sep=1pt] at (-0.05,0.825) {vote of both};
\node[anchor=east, inner sep=1pt] at (-0.05,0.275) {\reader{}};
\node[anchor=north, inner sep=1.5pt, text height=1.6ex, text depth=0.4ex] at (0.483,-0.02) {Belief};
\node[anchor=north, inner sep=1.5pt, text height=1.6ex, text depth=0.4ex] at (1.450,-0.02) {Knowl.};
\node[anchor=north, inner sep=1.5pt, text height=1.6ex, text depth=0.4ex] at (2.417,-0.02) {All};
\node[anchor=south west, font=\figfont\bfseries, inner sep=0pt] at (-1.300,2.320) {(b) Two readings beat one};
\path (0,-0.45) (0,2.70);
\end{tikzpicture}}\end{minipage}\hfill
\begin{minipage}[b]{0.330\linewidth}\centering\fitbox{\begin{tikzpicture}[font=\figfont, x=1cm, y=1cm]
\draw[black!10] (0,0.000) -- (3.900,0.000);
\draw[black!60] (0,0.000) -- (-0.05,0.000) node[left, inner sep=1pt] {0};
\draw[black!10] (0,1.100) -- (3.900,1.100);
\draw[black!60] (0,1.100) -- (-0.05,1.100) node[left, inner sep=1pt] {40};
\draw[black!10] (0,2.200) -- (3.900,2.200);
\draw[black!60] (0,2.200) -- (-0.05,2.200) node[left, inner sep=1pt] {80};
\node[rotate=90, anchor=south] at (-0.6,1.100) {\% right};
\node[anchor=south west, font=\figfont\bfseries, inner sep=0pt] at (-0.85,2.320) {(c) The rule settles conflicts};
\draw[cPlain, line width=1.4pt] (0.288,0) -- (0.288,1.309);
\filldraw[draw=cPlain, fill=cPlain] (0.288,1.309) circle (2.4pt);
\node[anchor=south, inner sep=1.6pt] at (0.288,1.349) {48};
\draw[cAnn, line width=1.4pt] (0.746,0) -- (0.746,1.245);
\filldraw[draw=cAnn, fill=cAnn] (0.746,1.245) circle (2.4pt);
\node[anchor=south, inner sep=1.6pt] at (0.746,1.285) {45};
\draw[cSCnine, line width=1.4pt] (1.204,0) -- (1.204,1.069);
\filldraw[draw=cSCnine, fill=cSCnine] (1.204,1.134) -- (1.269,1.069) -- (1.204,1.004) -- (1.139,1.069) -- cycle;
\node[anchor=south, inner sep=1.6pt] at (1.204,1.109) {39};
\draw[cOurs, line width=1.4pt] (1.662,0) -- (1.662,1.539);
\filldraw[draw=cOurs, fill=cOurs] (1.662,1.539) circle (2.4pt);
\node[anchor=south, inner sep=1.6pt] at (1.662,1.579) {56};
\node[anchor=north, inner sep=1.5pt] at (0.975,-0.02) {dev.};
\draw[cPlain, line width=1.4pt] (2.238,0) -- (2.238,1.165);
\filldraw[draw=cPlain, fill=cPlain] (2.238,1.165) circle (2.4pt);
\node[anchor=south, inner sep=1.6pt] at (2.238,1.205) {42};
\draw[cAnn, line width=1.4pt] (2.696,0) -- (2.696,1.466);
\filldraw[draw=cAnn, fill=cAnn] (2.696,1.466) circle (2.4pt);
\node[anchor=south, inner sep=1.6pt] at (2.696,1.506) {53};
\draw[cOurs, line width=1.4pt] (3.612,0) -- (3.612,1.613);
\filldraw[draw=cOurs, fill=cOurs] (3.612,1.613) circle (2.4pt);
\node[anchor=south, inner sep=1.6pt] at (3.612,1.653) {59};
\node[anchor=north, inner sep=1.5pt] at (2.925,-0.02) {test (belief)};
\draw[black!85, line width=0.8pt] (0,0) -- (0,2.200); \draw[black!85, line width=0.8pt] (0,0) -- (3.900,0);
\path (0,-0.45) (0,2.70);
\end{tikzpicture}}\end{minipage}
\caption{\textbf{Grounded reading.} (a) Belief and knowledge accuracy. (b) Narrative accuracy by reading. (c) Conflicts settled correctly. \textbf{Grounding helps epistemic questions, and two readings with the evidential rule beat one.}}
\label{fig:stage3}
\end{figure*}
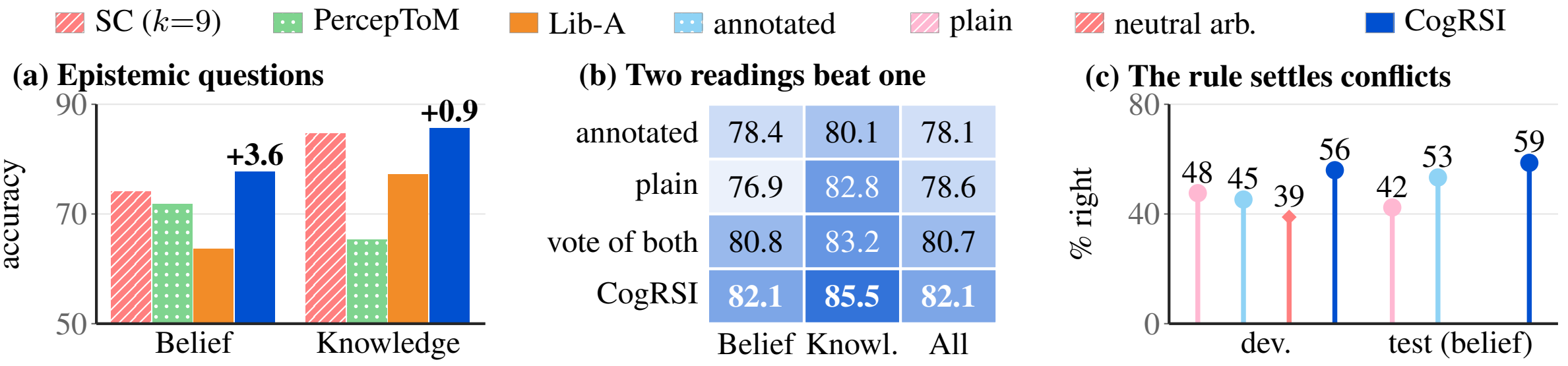
\paragraph{Grounding helps epistemic questions.} \reader{} is the most accurate system on belief and knowledge questions
(Figure~\ref{fig:stage3}a). Pooled over benchmarks it beats SC ($k{=}6$) on beliefs by \ConBeliefVsix{} points (CI
\ConBeliefVsixCI{}). Within benchmarks the belief gain remains (\ConWBelief{}, \ConWBeliefCI{}), while non-literal and
mixed questions lose (\ConWNonlit{} and \ConWMixed{}, Table~\ref{tab:constructs}).
Grounding helps the cognitive side of ToM, which in humans dissociates from the affective side
\citep{shamay2007dissociable}.

\paragraph{Perception limits true belief.} With the gold perceivers of Percept-ToMi in place of the perception call
(Table~\ref{tab:c3}), \reader{} rises from \PTVsixModel{} to \PTVsixGold{} and \readerL{} from \PTRsipModel{} to
\PTRsipGold{}, most on second-order true belief (\PTTBtwoVsixModel{} to \PTTBtwoVsixGold{}), while SC, which never
calls perception, stays near \PTSCModel{}. On FANToM gold perceivers lift belief accuracy (\FsixtyTwoBelVsixModel{} to
\FsixtyTwoBelVsixGold{}) but not All$^*$ (\FsixtyTwoAllVsixModel{} to \FsixtyTwoAllVsixGold{}).

\paragraph{Two readings, settled by the rule.} The vote over both readings beats either reading alone on belief,
knowledge and all narrative questions (Figure~\ref{fig:stage3}b). On \ArbDevConf{} development conflicts the evidential
rule is right \ArbDevEvid{} times, and an arbiter told only to choose the better answer is right \ArbDevNeutral{} times.
On belief test conflicts the rule beats both readings (\RouteBeliefAdj{} of \RouteBeliefConf{},
Figure~\ref{fig:stage3}c), and close-split arbitration fixes \FlipHelped{} answers while breaking \FlipHurt{}.

Table~\ref{tab:constructs} gives every ATOMS construct. Pooled over benchmarks \reader{} improves on SC ($k{=}6$) for
knowledge (\ConKnowledgeVsix{}, \ConKnowledgeVsixCI{}) as for beliefs; within benchmarks only the belief gain remains, and
non-literal (\ConWNonlit{}, \ConWNonlitCI{}) and mixed questions (\ConWMixed{}, \ConWMixedCI{}) lose. Where access is
irrelevant the annotations cost accuracy (non-literal: \AblNonlitAnn{} annotated vs.\ \AblNonlitPlain{} plain; emotions:
\AblEmotionAnn{} vs.\ \AblEmotionPlain{}); on non-literal conflicts the plain reading is right \RouteNonlitPlain{} of
\RouteNonlitConf{} times, arbitration \RouteNonlitAdj{}. The perception call is precise (\PercToMiP{}, \PercFanP{}) but incomplete (recall
\PercToMiR{}, \PercFanR{}).

\textbf{Gold perceivers.} Table~\ref{tab:c3} replaces each system's perception call with the gold perceivers of
Percept-ToMi (576 questions) and Percept-FANToM \citep{jung2024perceptom} (the 62 conversations of the main FANToM sample
that it annotates); everything else is unchanged. On Percept-ToMi \reader{} rises from \PTVsixModel{} to \PTVsixGold{},
\readerL{} from \PTRsipModel{} to \PTRsipGold{} and PercepToM from \PTPercepToMModel{} to \PTPercepToMGold{}. The largest
gains are on true-belief questions, where a character who saw the move must be counted: second order \PTTBtwoVsixModel{}
to \PTTBtwoVsixGold{} for \reader{} and \PTTBtwoRsipModel{} to \PTTBtwoRsipGold{} for \readerL{}. PercepToM gains instead on second-order false belief (\PTFBtwoPercepToMModel{} to
\PTFBtwoPercepToMGold{}), where its filter must keep the right perceivers, and loses on second-order true belief
(\PTTBtwoPercepToMModel{} to \PTTBtwoPercepToMGold{}). SC does not call
perception and stays at \PTSCModel{}/\PTSCGold{}. On FANToM gold perceivers raise belief accuracy to about
\FsixtyTwoBelVsixGold{} for every system but leave All$^*$ unchanged (\reader{} \FsixtyTwoAllVsixModel{} to
\FsixtyTwoAllVsixGold{}), since All$^*$ also needs every answerability and info-accessibility question of a set right.
\begin{table*}[t]
\centering
\footnotesize
\setlength{\tabcolsep}{0pt}
\begin{tabular}{@{}>{\raggedright\arraybackslash}p{2.1cm}>{\raggedright\arraybackslash}p{1.7cm}*{7}{>{\centering\arraybackslash}p{\dimexpr(\textwidth-3.8cm)/7\relax}}@{}}
\toprule
 & & \multicolumn{5}{c}{Percept-ToMi (576 questions)} & \multicolumn{2}{c}{FANToM (62 dialogues)} \\
\cmidrule(lr){3-7}\cmidrule(l){8-9}
Method & Perceivers & FB1 & FB2 & TB1 & TB2 & All & All$^*$ & Belief \\
\midrule
SC ($k{=}6$) & model & 84.8 & 92.1 & 56.6 & 5.6 & 59.8 & --- & --- \\
 & gold & 85.7 & 92.1 & 56.2 & 7.2 & 60.4 & --- & --- \\
\midrule
PercepToM & model & 89.0 & 1.9 & 56.9 & 47.2 & 48.8 & 2.2 & 82.3 \\
 & gold & 87.8 & 72.7 & 57.3 & 19.0 & 59.3 & 8.2 & 90.6 \\
\midrule
\rowcolor{ours} \reader{} & model & 89.9 & 92.6 & 56.6 & 2.8 & 60.5 & 26.0 & 79.3 \\
\rowcolor{ours}  & gold & 89.2 & 90.5 & 66.0 & 35.4 & 70.3 & 25.2 & 91.2 \\
\midrule
\rowcolor{ours} \readerL{} & model & 91.3 & 93.2 & 57.3 & 11.2 & 63.3 & 23.6 & 79.7 \\
\rowcolor{ours}  & gold & 93.4 & 94.7 & 76.5 & 59.1 & 81.0 & 22.0 & 90.5 \\
\bottomrule
\end{tabular}
\caption{\textbf{Perception or reasoning?} Each system with its own perception call (model) and with the gold perceivers of Percept-ToMi and Percept-FANToM in place of that call (gold; nothing else changes). FB/TB: false/true belief, first/second order. SC never calls perception, so its gold rows are a control. Mean of three runs (\readerL{}: six loop seeds). \textbf{Gold perceivers lift both readers most on true-belief questions. On FANToM, belief accuracy rises and All$^*$ does not.}}
\label{tab:c3}
\end{table*}

\subsection{Cost and robustness}
\label{app:rq67}
\paragraph{On the cost--accuracy frontier, with a significant margin.} \reader{} makes \CallsVsix{} calls per question against \CallsSC{} for SC
($k{=}6$) and uses \TokRatioVsix{}$\times$ its tokens. SC ($k{=}9$) uses as many and scores lower on Main
(\OvSCnine{} vs.\ \OvVsix{}, Figure~\ref{fig:cost}a). A Stage-1 run takes \RSIRunHours{} hours on one H200. Against SC ($k{=}9$) the paired 95\% intervals exclude zero on FANToM,
SimpleToM and Hi-ToM only, and Hi-ToM does not survive Holm correction (Table~\ref{tab:paired-small}). Against the
strongest baseline of each benchmark, \reader{} wins \PairWinsVsix{}, ties
\PairTiesVsix{} and loses \PairLossesVsix{}, and none of these differences survives Holm correction
(Table~\ref{tab:paired}). On each of the five
backbones its four-benchmark mean beats SC ($k{=}9$) by \BBMarginMin{} to \BBMarginMax{} points at \BBRatioMin{} to
\BBRatioMax{} times its tokens (Figure~\ref{fig:cost}b, Table~\ref{tab:sc9bb}).

\begin{figure}[t]
\centering
{\figfont\bfseries (a) Sample mix of the vote}\par\smallskip
\fitbox{\begin{tikzpicture}[font=\figfont, x=1cm, y=1cm]
\draw[black!20] (0.00,1.68) rectangle (1.35,2.10);
\fill[cOurs!29] (1.35,1.68) rectangle (2.70,2.10);
\node[text=black] at (2.03,1.89) {78.0};
\fill[cOurs!39] (2.70,1.68) rectangle (4.05,2.10);
\node[text=black] at (3.38,1.89) {78.4};
\fill[cOurs!10] (0.00,1.26) rectangle (1.35,1.68);
\node[text=black] at (0.68,1.47) {77.3};
\fill[cOurs!47] (1.35,1.26) rectangle (2.70,1.68);
\node[text=black] at (2.03,1.47) {78.7};
\fill[cOurs!59] (2.70,1.26) rectangle (4.05,1.68);
\node[text=white] at (3.38,1.47) {79.1};
\fill[cOurs!27] (0.00,0.84) rectangle (1.35,1.26);
\node[text=black] at (0.68,1.05) {77.9};
\fill[cOurs!66] (1.35,0.84) rectangle (2.70,1.26);
\node[text=white] at (2.03,1.05) {79.3};
\fill[cOurs!73] (2.70,0.84) rectangle (4.05,1.26);
\node[text=white] at (3.38,1.05) {79.6};
\fill[cOurs!52] (0.00,0.42) rectangle (1.35,0.84);
\node[text=white] at (0.68,0.63) {78.8};
\fill[cOurs!56] (1.35,0.42) rectangle (2.70,0.84);
\node[text=white] at (2.03,0.63) {79.0};
\fill[cOurs!79] (2.70,0.42) rectangle (4.05,0.84);
\node[text=white] at (3.38,0.63) {79.8};
\fill[cOurs!54] (0.00,0.00) rectangle (1.35,0.42);
\node[text=white] at (0.68,0.21) {78.9};
\fill[cOurs!73] (1.35,0.00) rectangle (2.70,0.42);
\node[text=white] at (2.03,0.21) {79.6};
\fill[cOurs!80] (2.70,0.00) rectangle (4.05,0.42);
\node[text=white] at (3.38,0.21) {79.9};
\draw[cNeg, line width=1pt] (2.70,0.00) rectangle (4.05,0.42);
\node at (0.68,2.28) {0};
\node at (2.03,2.28) {1};
\node at (3.38,2.28) {2};
\node[anchor=east] at (-0.05,1.89) {0};
\node[anchor=east] at (-0.05,1.47) {1};
\node[anchor=east] at (-0.05,1.05) {2};
\node[anchor=east] at (-0.05,0.63) {3};
\node[anchor=east] at (-0.05,0.21) {4};
\node at (2.03,2.62) {plain samples};
\node[rotate=90, anchor=south] at (-0.32,1.05) {annotated samples};
\end{tikzpicture}}\par\medskip
{\figfont\bfseries (b) Close-split thresholds}\par\smallskip
\begin{tabular}{@{}lc@{}}
\toprule
Close-split rule & Accuracy \\
\midrule
top $\le$ 3, runner-up $\ge$ 2 & 81.0 \\
top $\le$ 3, runner-up $\ge$ 3 & 81.0 \\
\rowcolor{ours} top $\le$ 4, runner-up $\ge$ 2 (ours) & \textbf{81.3} \\
top $\le$ 4, runner-up $\ge$ 3 & 81.0 \\
\bottomrule
\end{tabular}

\caption{\textbf{Sensitivity of \reader{}'s fixed choices} (narrative test benchmarks, macro mean, all runs). (a) Accuracy
of a majority over the first $a$ annotated and $p$ plain samples of the vote-only runs; ours (4+2) outlined. (b) Close-split
arbitration restricted to stricter thresholds (looser ones would need new arbiter calls). Robustness checks on test data,
not selections. \textbf{Mixes with both readings are best; ours scores \SensMixOurs{} (best mix: \SensMixBest{}), and our
thresholds score highest.}}
\label{fig:sens}
\end{figure}
Table~\ref{tab:paired} gives paired differences for every benchmark: against each benchmark's strongest baseline
\reader{} wins \PairWinsVsix{}, ties \PairTiesVsix{} and loses \PairLossesVsix{} of \PairCols{}, and after Holm correction
it wins \PairHolmWinsVsix{} and loses \PairHolmLossesVsix{}. A single \reader{} run
(\EnsVsixSingle{}) is above three pooled runs of SC ($k{=}6$) (\EnsSCPooled{}) on the main benchmarks other than FANToM, where All$^*$ sets cannot be pooled by vote (ExploreToM's two partitions counted separately). Each loop runs once, offline: one Stage-1 run takes \RSIRunHours{}
elapsed hours on one H200.
Table~\ref{tab:backbones} gives every backbone on the benchmarks run for all five, Table~\ref{tab:sc9bb} SC ($k{=}9$) on
each of them with its token ratio, and Figure~\ref{fig:sens}
the fixed choices. The sample mix is checked offline from the saved labels of the vote-only
design: for every $a\le4$ annotated and $p\le2$ plain samples, the accuracy of a majority over the first $a$ and $p$ of
them. The close-split thresholds can be checked only in the stricter direction, by restricting arbitration to splits
that also meet the stricter rule; looser thresholds would need new arbiter calls. Both checks use test data and are
robustness checks, not selections.

\subsection{Case study}
\label{app:rq8}\label{app:case}\label{app:example}
\paragraph{Arbitration breaks the tie.} On an egocentric lie from ToMBench (Figure~\ref{fig:case},
Table~\ref{tab:case}), every run of every other system answers that Miss Zhou's claim is true. In \reader{}'s first run
the annotated reading says false, the plain one true, and the vote ties. Arbitration chooses false in all three runs.

\begin{table}[t]
\centering
\footnotesize
\setlength{\tabcolsep}{3pt}
\begin{tabular}{p{0.21\columnwidth}p{0.08\columnwidth}p{0.61\columnwidth}}
\toprule
\multicolumn{3}{p{0.92\columnwidth}}{\textbf{Story.} Miss Zhou is a professional in the competitive advertising industry. At an industry gathering, someone asks her about the details of an innovative advertising strategy her company is about to launch. Miss Zhou replies that her team develops a brand new marketing method that changes the industry landscape, but currently cannot reveal more information. She also emphasizes that this is a major breakthrough in the industry. After the industry gathering ends, Miss Zhou lets her team start planning the innovative strategy within this week.\newline \textbf{Question.} Is what Miss Zhou says true? (A) Yes (B) No. \textbf{Gold:} B.} \\
\midrule
Method & Ans. & What decides it (from the run logs) \\
\midrule
SC ($k{=}6$) & A \textcolor{neg}{$\times$} & majority of six samples is A in all three runs \\
PercepToM & A \textcolor{neg}{$\times$} & ``target character not found'' $\to$ zero-shot fallback, A (all runs) \\
\libA{} & A \textcolor{neg}{$\times$} & inherited tools; A in all three seeds \\
\rowcolor{ours}Annotated & B \textcolor{pos}{$\surd$} & run 0: three of four annotated samples answer B \\
\rowcolor{ours}Plain & A \textcolor{neg}{$\times$} & run 0: both plain samples A; the 3--3 tie keeps the first sampled answer, A \\
\rowcolor{ours}\sys{} & B \textcolor{pos}{$\surd$} & view conflict $\to$ arbitration chooses B (runs 0, 1); close split $\to$ arbitration chooses B (run 2) \\
\rowcolor{ours}\readerL{} & A \textcolor{neg}{$\times$} & each seed's principle (object locations and states): all six samples A, no arbitration call \\
\bottomrule
\end{tabular}
\caption{\textbf{Case study}: a ToMBench item every baseline misses and \sys{} gets right by arbitration.}
\label{tab:case}
\end{table}

The main case (Table~\ref{tab:case}) was chosen as a test item that every baseline run answers wrongly and every \reader{}
run answers correctly; the decisive step of each run is read from its saved records. With the loop's principles
(\readerL{}), which are about object locations and states, all six samples agree on the wrong answer, so no arbitration
call is made. A second case, a SimpleToM development question on which the readings disagree (held-out sample~1): the
plain samples credit Isla with knowing about the malware, which the story never shows her learning; the annotation and
the evidential rule do not, while the neutral arbiter sides with the plain reading.

\begin{figure}[t]
\promptbox{Arbitration example}{Model Output}{%
\bh{Story}The E-Reader Isla is considering buying was previously owned by a notorious hacker and still contains malicious software. Isla examines the E-Reader, flipping through its pages and admiring its lightweight design.
\bh{Question}What will Isla likely do after examining the E-Reader?\\(A) Decide to purchase the E-Reader based on its design and functionality.\\(B) Decide not to purchase the E-Reader due to the risk of malicious software.\\Gold: (A).
\par\smallskip\noindent\textbf{The evidential rule keeps the gold answer; the neutral arbiter follows the plain samples.}
\codebox{\begin{tabular}{@{}lll@{}}annotated majority & (A) & correct\\plain samples & (B) & wrong\\evidential arbiter & (A) & correct\\neutral arbiter & (B) & wrong\end{tabular}}}{Arbitration on a SimpleToM development item (held-out sample 1).}{box:example}
\end{figure}

\section{Extended related work}
\label{app:related}
\paragraph{Self-improving systems.} Skill and tool libraries that grow with use go back to program induction
\citep{ellis2021dreamcoder} and embodied agents \citep{wang2023voyager}; agents distil experience into insights
\citep{zhao2024expel} or workflows \citep{wang2025awm}, learn from verbal reflection \citep{shinn2023reflexion}, improve
their own scaffolding \citep{zelikman2024stop,yin2025godel,robeyns2025sica}, design agents \citep{hu2025adas}, keep an
open-ended archive of self-modified agents \citep{zhang2025dgm}, and evolve prompts \citep{fernando2024promptbreeder,
agrawal2025gepa} or programs against evaluators \citep{novikov2025alphaevolve}. Table~\ref{tab:related} compares what each
keeps and how it admits a change; \cogrsi{} alone compares inheritance with a rebuild at the same budget and lets the
same loop write a principle instead of a procedure.

\paragraph{Theory of mind in language models.} SymbolicToM \citep{sclar2023symbolictom} tracks beliefs in graphs, SimToM
\citep{wilf2024simtom} answers from the events a character knows, TimeToM \citep{hou2024timetom} builds temporal belief
chains, PercepToM \citep{jung2024perceptom} filters the story by perception, Decompose-ToM \citep{sarangi2025decompose}
recursively simulates perspectives and Shoes-of-Others \citep{shinoda2026shoes} prefills the target's perspective; all
apply a fixed procedure. DEL-ToM \citep{wu2025deltom} scores belief traces with a trained verifier, AutoToM
\citep{zhang2025autotom} and ThoughtTracing \citep{kim2025thoughttracing} infer mental states by inverting an agent model,
and \citet{lu2025tomrl} train small models with reinforcement learning, whose ToM does not generalise beyond the training datasets
\citep{sarangi2025rltom}. Stress tests show that LLM ToM often relies on templates \citep{ullman2023large,shapira2024clever};
our construct analysis (Appendix~\ref{app:rq5}) adds to this.

\paragraph{Cognitive science.} Children learn that seeing leads to knowing \citep{wimmer1988access,pratt1990looking};
adults track who saw what alongside a flexible inferential system \citep{apperly2009two} and infer beliefs from actions
\citep{baker2017rational}; cognitive and affective ToM dissociate \citep{shamay2007dissociable}. We organise the
construct analysis by the ATOMS taxonomy \citep{beaudoin2020atoms}.

\end{document}